%% file: main.tex
\documentclass[manuscript,screen,nonacm]{format/acmart}
\AtBeginDocument{%
  }

\setcopyright{none}
\copyrightyear{2018}
\acmYear{2018}
\acmDOI{XXXXXXX.XXXXXXX}
\acmConference[Conference acronym 'XX]{Make sure to enter the correct
  conference title from your rights confirmation email}{June 03--05,
  2018}{Woodstock, NY}
\acmISBN{978-1-4503-XXXX-X/2018/06}

\usepackage{enumitem}

\usepackage[dvipsnames]{xcolor}
\usepackage{colortbl}
\usepackage{multirow}
\usepackage{array}
\usepackage{siunitx}
\usepackage{graphicx}
\usepackage{ragged2e}
\usepackage{soul}
\usepackage{xspace}
\usepackage{wrapfig}
\usepackage{balance} % balance cols
\usepackage{hyperref} % enable href in footnotes, load hyperref last
\usepackage{tabularx} % If tabularx, load it after hyperref
\usepackage{adjustbox} % resize table to fit page length
\usepackage{longtable} % handle long table
\usepackage{booktabs}
\usepackage{subcaption}
\usepackage[nameinlink]{cleveref}
\usepackage{ulem}

\usepackage{longtable}

\usepackage{booktabs}
\usepackage{array}
\newcolumntype{L}[1]{>{\raggedright\arraybackslash}p{#1}}

\usepackage[most]{tcolorbox} % create boxes in text
\tcbset{before skip=10pt, after skip=10pt}

\usepackage{dblfloatfix}

\definecolor{yhcolor}{RGB}{140, 70, 180}

\definecolor{navy}{RGB}{0,0,180}
\definecolor{maroon}{RGB}{128,0,0}
\definecolor{forest}{RGB}{34,139,34}

\renewcommand{\emph}[1]{\textit{\underline{#1}}}

\begin{document}

\title{\textsc{CounselReflect}: Opportunities and Challenges for Designing Tools to Support Self-Reflection on Mental Health and Well-Being Conversations with AI}


\author{Yahan Li}
\authornote{Equal contribution.}
\affiliation{%
  \institution{University of Southern California}
  \city{Los Angeles}
  \state{California}
  \country{United States}}
\email{yahanli@usc.edu}

\author{Chaohao Du}
\authornotemark[1]
\affiliation{%
  \institution{University of Southern California}
  \city{Los Angeles}
  \state{California}
  \country{United States}}
\email{chaohaodu@gmail.com}

\author{Christopher Chun Kuizon}
\authornotemark[1]
\affiliation{%
  \institution{University of Southern California}
  \city{Los Angeles}
  \state{California}
  \country{United States}}
\email{ckuizon@usc.edu}

\author{Zeyang Li}
\authornotemark[1]
\affiliation{%
  \institution{University of Southern California}
  \city{Los Angeles}
  \state{California}
  \country{United States}}
\email{lizeyang@usc.edu}

\author{Nimra Ishfaq}
\authornotemark[1]
\affiliation{%
  \institution{The University of Texas at Austin}
  \city{Austin}
  \state{Texas}
  \country{United States}}
\email{nimraishfaq@utexas.edu}

\author{Shupeng Cheng}
\affiliation{%
  \institution{University of Southern California}
  \city{Los Angeles}
  \state{California}
  \country{United States}}
\email{shupengc@usc.edu}

\author{Angelica Yinling Sun}
\affiliation{%
  \department{USC Rossier School of Education}
  \institution{University of Southern California}
  \city{Los Angeles}
  \state{California}
  \country{United States}}
\email{yinlings@usc.edu}

\author{Adam C. Frank}
\affiliation{%
  \department{Department of Psychiatry and Behavioral Sciences}
  \institution{University of Southern California}
  \city{Los Angeles}
  \state{California}
  \country{United States}}
\email{adamfran@usc.edu}

\author{Angel Hsing-Chi Hwang}
\authornote{Co-corresponding authors.}
\affiliation{%
  \institution{University of Southern California}
  \city{Los Angeles}
  \state{California}
  \country{United States}}
\email{angel.hwang@usc.edu}

\author{Ruishan Liu}
\authornotemark[2]
\affiliation{%
  \institution{University of Southern California}
  \city{Los Angeles}
  \state{California}
  \country{United States}}
\email{ruishanl@usc.edu}

% \renewcommand{\shortauthors}{Last Name et al.}
\renewcommand{\shortauthors}{Li et al.}

\begin{abstract}
  \input{paper_abstract}
\end{abstract}



% \includegraphics[width=\textwidth]{sampleteaser}
%  \Description{Enjoying the baseball game from the third-base seats. Ichiro Suzuki preparing to bat.}


\maketitle

\input{paper_body}

\begin{acks}
We thank all participants for their participation and valuable input, and the mental health professionals for their insights and feedback. We also thank John Bosco S. Bunyi for facilitating connections with mental health professionals who provided valuable feedback on \textsc{CounselReflect}, Konghao Zhao for insightful feedback, and Yuehan Qin for helpful discussions.
\end{acks}
\bibliographystyle{ACM-Reference-Format}
\bibliography{ref}

\appendix

\input{paper_appendix}

\end{document}

%% file: paper_abstract.tex
AI is increasingly used for mental health and well-being support, creating an urgent need for safer engagement, while design, evaluation, and governance take time to develop. We explore a complementary approach: helping users critically reflect on their own AI conversations. We introduce \textsc{CounselReflect}\footnote{Website: \url{https://www.counselreflect.com}; \\
Chrome Extension: \url{https://chromewebstore.google.com/detail/counselreflect/kkplffeneacdfjaonjhinlhconmdiibb}; \\
Github: \url{https://github.com/counselreflectteam/CounselReflect}\\
Demo Video: \url{https://youtu.be/7B7dHhs1cvg}
}, a tool that translates literature-grounded counseling quality metrics into a user-facing reflection framework. Using \textsc{CounselReflect} as a study probe, we interviewed 21 users of AI for mental health and well-being support. Although most participants did not routinely reflect on their conversations, they articulated concrete questions they would want reflection to address. Tool-assisted reflection also revealed challenges: participants selectively sought evidence confirming existing perceptions of AI and prioritized dimensions they already valued. We argue that reflection tools should surface blind spots and scaffold more holistic examination of AI interactions. Finally, overcoming emotional barriers to revisiting tense conversations remains a major design challenge and warrants input from future work.

%% file: paper_body.tex
\section{Introduction}
% 1--1.5 pages

Approximately 27\% of current AI users regularly use AI for mental health and well-being support, with an additional 40\% reporting occasional use or openness to using AI for these purposes~\cite{bodner2026barriers}. 
Many of these interactions involve highly sensitive topics. For example, OpenAI recently reported that more than one million users in the United States express suicidal thoughts to ChatGPT each week~\cite{openai2025sensitive}. 
As AI becomes increasingly embedded in emotional support seeking, researchers, practitioners, and policymakers have emphasized the urgent need for safer system design and governance.

However, the path toward safer AI systems is necessarily long. Developing clinically robust systems, rigorously evaluating their safety, and generating evidence to inform regulation are all multi-year endeavors~\cite{Cooper-responsible-AI-CHI26}.
Meanwhile, there is little indication that people will stop using AI for mental health support despite repeated warnings about its risks {\citep{Stade_Tait_Campione_Stirman_Eichstaedt_2026,De_Freitas_Cohen_2024_the_health_risks}}.
Indeed, existing safety interventions, such as encouraging users to take breaks or reminding them that AI is not human, have also shown limited evidence of substantially changing user behavior {\citep{Laestadius_Campos_Castillo_2026_reminders_that_chatbos_are_not_human_can_be_risky, Golden_Aboujaoude_2026_a_transdiagnostic_model_for_how_general_purpose_AI_chatbots_can_perpetuate_OCD_and_anxiety_disorders}}.

In this work, we explore a complementary approach: rather than focusing on designing AI systems that deliver mental health interventions, we investigate how to help users reflect on and engage more mindfully with their own AI conversations. 
This mindset shift is motivated by two considerations.
First, many people who seek support from AI do not identify as needing formal mental health treatment or are reluctant to pursue clinical services, making reflective support applicable to a broader population {\citep{Stade_Tait_Campione_Stirman_Eichstaedt_2026, Kosyluk_Baeder_Greene_Tran_Bolton_Loecher_DiEva_Galea_2024_label_avoidance_and_use_of_a_mental_health_chatbot}}.
Second, even conversations that users perceive as relatively low stakes can gradually develop patterns associated with emotional reliance, distorted social expectations, or other safety concerns {\citep{shi2026stumblingaiemotionaldependence}}. 
Consequently, we propose designing tools that are not tied to specific diagnoses or treatment goals, but instead support users in monitoring and reflecting on their AI interactions, empowering them to recognize emerging patterns and make more informed decisions about their engagement.

To this end, we design and evaluate \textsc{CounselReflect}, a reflection tool that helps users examine their mental health conversations with AI. 
\textsc{CounselReflect} adapts literature-grounded counseling quality metrics into a user-facing reflection framework.
Rather than assessing whether AI can replace human therapists, \textsc{CounselReflect} prompts users to consider the extent to which their conversations resemble supportive counseling interactions and highlights conversational patterns that may warrant attention. 
To iteratively develop the system, we conducted two rounds of design and evaluation with mental health professionals, individuals with experience receiving mental health care, and users seeking mental health support from LLMs. 

Our study shows that, although most users do not currently reflect on their mental health and well-being conversations with AI, there are opportunities to support such practices. Participants identified concrete questions they would like to revisit. However, designing reflection tools requires caution: users may selectively seek evidence that confirms their existing perceptions of AI or focus primarily on dimensions they already value. Reflection tools should therefore not merely facilitate users' existing inquiries, but also proactively surface blind spots and scaffold more holistic examination of AI interactions.

The contributions of this work are threefold. 
(1) We propose reflection as a complementary safety paradigm for AI-mediated mental health support, shifting the focus from making AI act more like a therapist to helping users critically evaluate their own interactions with AI. 
(2) We present \textsc{CounselReflect}, a literature-grounded reflection tool that operationalizes counseling quality metrics into actionable feedback for end users. 
(3) Through iterative design and evaluation with mental health professionals, mental health care clients, and LLM users, we provide empirical insights into the opportunities and challenges of supporting reflective AI use, informing the design of future safety interventions for high-stakes AI conversations.

\section{Related Work}

\subsection{Tools that Support In-Situ Reflection in Conversations}
There is a dearth of reflection tools designed for personal contexts, especially for people who are cognizant of their reliance on AI for mental health support \cite{namvarpour2026understanding}. Previous HCI tools focus on supporting professionals with reflection to help them improve their performance, communication, and collaboration skills. These systems use AI to facilitate in-the-moment or retroactive reflection, using nudges, roleplay, and data visualizations. Studies find that even lightweight and interactive feedback can trigger deep reflection, learning, and significant behavioral change \cite{10.1145/3710993, 10.1145/3816046.3816219}. Accordingly, \textsc{CounselReflect} is designed to fill the gap in our understanding of best practices for developing and evaluating reflection tools for mental health conversations with AI.

Prior work discusses design features that scaffold reflection: systems rely on contextualized feedback, interactive learning, and retrospective memory that enhance the impact of interpretive knowledge. As a whole, it is important to focus on the trajectory of a user's conversation, rather than individual turns, to surface insights that may go unnoticed otherwise \cite{10.1145/3816046.3816219}. The designers' choice of modality also plays a role in this process, as voice-based input and roleplay are especially helpful for users in processing higher-level issues, refining support-seeking behaviors, and decreasing cognitive burden. These affordances, however, are only helpful if the user retains autonomy by overriding undesirable outputs, editing responses, and postponing reflection \cite{10.1145/3743735}. The linking of evidence from scholarly literature further improves the credibility of AI-generated advice. These design principles can be applied to \textsc{CounselReflect}, which is grounded in psychological research and post-assessment reflection, traces how a conversation unfolds, and allows users to ask questions.

Many prior studies evaluate reflection tools using cross-sectional design, which may leave the risk of producing rumination in users undetected. Studies find this risk to be prevalent in tools that give excessive, in-the-moment criticism and create visual overload \cite{chaszczewicz-etal-2024-multi, 10.1145/3379503.3405662, 10.1145/3772318.3791546}. While legitimizing emotional discourse can be a strength of such AI tools, they might create anxiety in individuals with mental health issues \cite{10.1145/3810204,10.1145/3772318.3791821}.

\subsection{Users' Conversations with AI on Mental Health and Well-being}
Individuals are increasingly turning to AI for support with emotional coping, companionship, psychoeducation, and self-diagnoses \cite{luo2025seeking}. Most support-seeking concentrates around topics related to stress, anxiety, and interpersonal difficulties \cite{diaz2026spontaneous}. Many users craft fictional personas and use them to gain advice, build direct relationships, and articulate personal values \cite{laestadius2024too, zhang2025dark}. Overall, these tools are received quite positively as coping partners, especially if there is a shortage of offline support. 

AI chatbots fill genuine gaps in therapeutic and clinical services. Their low costs, constant availability, and negligible barrier to access appear to address poor health coverage and misalignment with clinician values. Individuals report that the facility to have a listening ear in itself is valuable \cite{song2025typing}. Additionally, they serve underserved communities who might experience barriers to formal care \cite{ma2024evaluating}. In terms of design features, AI tools create agency by allowing users to steer or exit conversations and AI-generated responses are perceived as more empathic than those of humans \cite{ayers2023comparing}. However, as these systems are founded on entrepreneurial motives, they forgo clinical validation and prioritize engagement, which can compromise user well-being.

While AI chatbots enhance access, there are significant gaps in the quality of care provided to users. Clinicians are absent from development processes, and training data is rife with biases and stereotypes. Chatbots have been documented to fail users in high-risk situations, encourage self-harm behaviors, and inappropriately respond to minors \cite{moore2025expressing}. Researchers emphasize that these biases cannot be simply retrofitted or fine-tuned for accuracy but require active community involvement in development workflows \cite{ma2024evaluating}. Furthermore, users report being overreliant on chatbots and experiencing close bonds that can displace human care \cite{namvarpour2026understanding}. In particular, young individuals describe chatbots as superior due to their unconditional support, ceaseless availability, and empathy that researchers deem surface-level \cite{namvarpour2026understanding, bae2021social}. 

\subsection{Reflecting on Mental Health and Well-being Conversations}
%- What has already been covered
 While early reflection tools supported record-keeping, the identification of patterns in stress, behaviors, and health status, recent systems also help users contextualize these cues and identify their underlying causes \cite{neupane2024momentary, ren2025emovis}. Previous tools created emotion records, emotion typologies, and facilitated sense-making for data from sensors, photos, and reflection questions. Systems such as EmoVis generated data visualizations of information collected from wearables, while Neupane et al. used a combination of location and PPG data and user feedback \cite{ren2025emovis,neupane2024momentary}. 
 Subsequently, researchers found that users needed more than emotion records, highlighting the need for interpretive signs, causes behind emotions, and coping mechanisms \cite{schueller2021understanding}. This resulted in systems that let users lead the reflection activities and participate actively. Kim et al. suggested a mechanism for expressing moods through symbolic and colorful drawings on a mobile application \cite{kim2025lino}. Systems also allow individuals to create tailored personas, utilize long-term memory, engage in personal writing, engage with multiple perspectives, and create records for clinical consultations \cite{zheng2025customizing, kim2025lino,nepal2024mindscape, song2025exploreself,kim2024diarymate, kim2024mindfuldiary}.

%- What didn't work/ limitations
It is noteworthy that AI-driven reflection tools can be disruptive as AI is designed to create engagement \cite{wester2026chaplains}. AI may interrupt the user's flow of thought and make the users adopt its emotional interpretations \cite{kim2024diarymate}. Moreover, continuous questioning and reflection can create rumination, which has been linked to depression, anxiety, and psychosis \cite{eikey2021beyond, patel2023associations, watkins2020reflecting}.

\subsection{Guiding Principles for Evaluating Mental Health and Well-being Conversations}
Evaluative frameworks broadly include three categories: approaches focused on user outcomes, clinical and conceptual assessments, and automated evaluations \cite{inkster2018empathy, chen2024framework, moore2025expressing, bentley2026benchmarking,byun2026cradle, liu2026companionbench}. Inkster et al. explored whether chatbots can improve outcomes for depressive moods by comparing low and high engagement users \cite{inkster2018empathy}. Their in-app based feedback captured users' thoughts, feelings, and experiences in real time and personalized the application design accordingly. Moore et al. conducted a mapping review of clinical and multicultural guidelines and synthesized principles of good therapy \cite{moore2025expressing}. Their review found significant safety issues and barriers in forming therapeutic alliance with LLMs. Jin et al. proposed the first suite of mental health tasks and evaluated LLMs for clinical knowledge, diagnostic acumen, fluency, and empathy in responses  \cite{jin2023psyeval}. Xu et al. leveraged online text data to evaluate prediction capabilities and found that few-shot prompting and instruction-finetuning improved model performance on mental health tasks \cite{xu2024mental}. Chen et al. proposed a roleplay-based evaluation method that leveraged red-teaming to assess model responses for trustworthiness, safety, and appropriateness \cite{chen2024framework}. More recent studies automated these assessments at scale. Byun et al. created a clinician-annotated benchmark spanning seven types of crisis events to support detection \cite{byun2026cradle}. Bentley et al. verified an LLM judge against clinicians' safety scores for conversations that had varying risk levels  \cite{bentley2026benchmarking}. Liu et al. created scenarios and user simulations that were grounded in real world data and discovered that warmth and substance differed sharply in AI-generated responses \cite{liu2026companionbench}.

The diverse and complex nature of mental health contexts creates several limitations for these frameworks. Evaluation frameworks are designed for shorter conversations, which may not apply to chronic conditions and constrain understanding of safety concerns that develop over long multiturn conversations \cite{liu2026companionbench, bentley2026benchmarking}. Similarly, manual red-teaming approaches lack diversity in test cases and scalable application \cite{chen2024framework}.

\section{\textsc{CounselReflect}: A Reflection Probe}
To support reflection on AI-mediated mental health and well-being conversations, we developed \textsc{CounselReflect}, a user-facing system that turns conversational evaluations into structured feedback on users’ AI interactions. Rather than providing diagnosis or treatment, we use \textsc{CounselReflect} as a study probe to examine how users interpret and engage with this feedback.

\subsection{System Overview and Interaction Workflow}
\begin{figure}[h!]
    \centering
    \includegraphics[clip, trim=20pt 60pt 40pt 85pt, width=\textwidth]{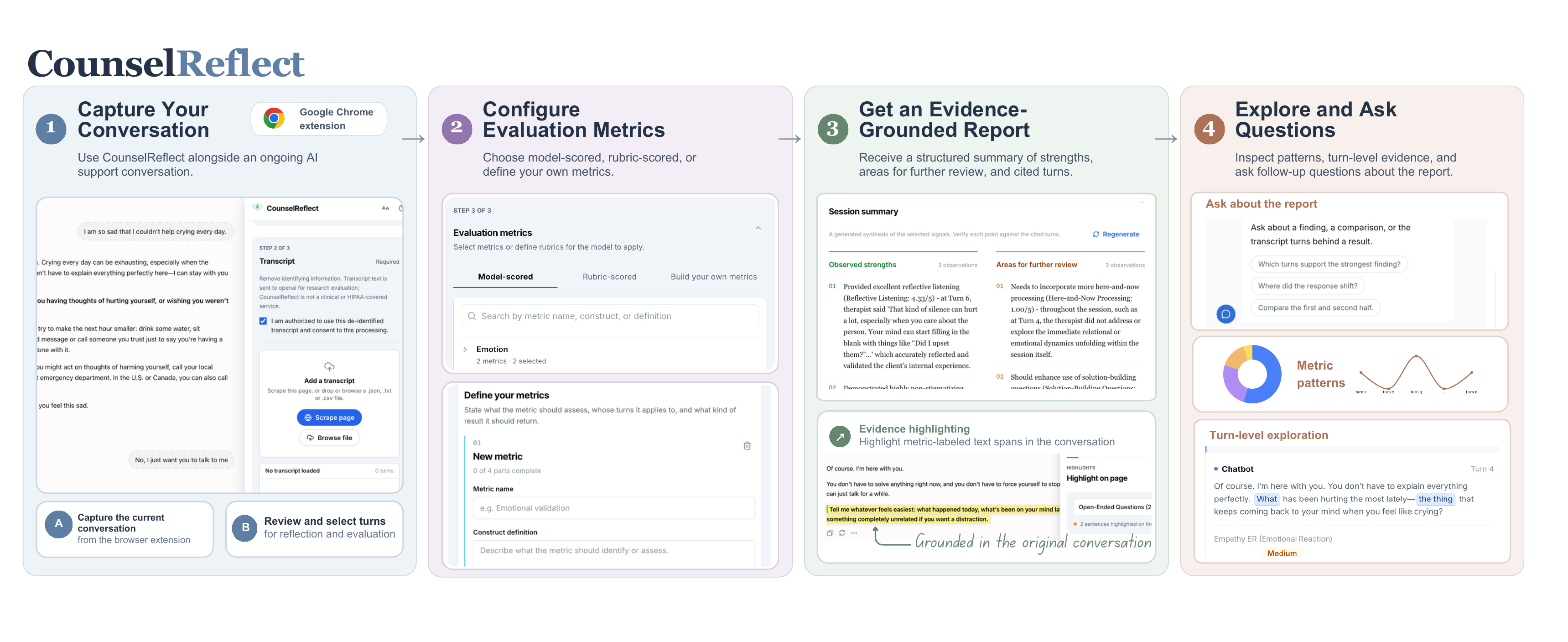}
\caption{Overview of the \textsc{CounselReflect} browser-extension workflow: conversation capture, metric configuration, evidence-grounded reporting, and interactive exploration.}
    \label{fig:extension_cli_screenshots}
\end{figure}

Figure~\ref{fig:extension_cli_screenshots} summarizes the \textsc{CounselReflect} browser-extension workflow. Users first (1) capture an AI support conversation and select the turns to evaluate. They then (2) configure the evaluation by selecting model-based or rubric-based metrics, with the option to define custom metrics. \textsc{CounselReflect} (3) generates a report that summarizes strengths and areas for review; for metrics that provide span-level labels, the corresponding text spans are highlighted in the conversation. Users can then (4) inspect metric trajectories and turn-level results and ask follow-up questions about individual findings.

\paragraph{Deployment Modes.}
\textsc{CounselReflect} supports three deployment modes. The \textbf{browser extension} supports in-context evaluation of conversations on web-based chat platforms. The \textbf{web interface} supports interactive analysis of individual conversations, while the \textbf{command-line interface (CLI)} supports batch evaluation for research workflows.

\subsection{Evaluation Framework}

{\textsc{CounselReflect} organizes evaluation metrics into two families: (i) 12 model-based metrics produced by task-specific predictors (Sec. \ref{sec:model-based}), and (ii) rubric-based metrics that extend coverage beyond trained predictors, including literature-derived metrics and user-defined custom metrics scored with configurable LLM judges (Sec. \ref{sec:llm-judge}). We then present the reporting interface that connects session-level summaries with turn-level scores and supporting excerpts. 

% \ruishan{I didn't find the subsection for reporting interface?} \yahan{I deleted that since we mentioned what we have in 3.1}

\subsubsection{Model-Based Metrics} \label{sec:model-based}
\textsc{CounselReflect} includes 12 model-based metrics spanning eight evaluation tasks (Table~\ref{tab:model_metric_table}; Appendix~\ref{appendix_model_based_metrics}). We use off-the-shelf models when available and train or fine-tune models when needed to fill coverage gaps. We report performance on the original benchmark dataset when available, or on a closely related external dataset otherwise\footnote{For datasets originally formulated as multi-label classification tasks, we use only single-label instances due to the predictor's single-label setting.}. The framework is modular, allowing users to add local evaluators through the released codebase.

\input{table/tab-metric}

\subsubsection{Rubric-Based Metrics} \label{sec:llm-judge}

To extend coverage beyond task-specific predictors, \textsc{CounselReflect} supports literature-derived and user-defined rubrics scored by an LLM-as-judge \footnote{Users may select hosted or locally deployed models based on cost, latency, and privacy requirements. }\cite{liu2023gevalnlgevaluationusing, gu2025surveyllmasajudge, li2024crowdsourceddatahighqualitybenchmarks}. 
For literature-derived metrics, the authors identified 91 candidates with assistance from GPT-5.1. A psychologist then reviewed the candidates and selected 69 for the final set. Each metric includes a definition, references, and behavioral anchors for low (1), medium (3), and high (5) ratings.
Beyond these literature-derived metrics, users can define custom metrics by describing the metric they want to evaluate. \textsc{CounselReflect} then generates a rubric that users can iteratively refine using synthetic examples \citep{shi2025humanintheloopframeworkreliablepatch, bai2026irulerintelligiblerubricbaseduserdefined}.

\subsection{AI-Assisted Implementation.}

We used generative AI tools as coding assistants during the implementation of \textsc{CounselReflect}, including for generating, revising, and debugging portions of the software code. All AI-assisted code was reviewed and modified as needed by the research team. 

\subsection{Validating Usability of \textsc{CounselReflect}}

Before using \textsc{CounselReflect} as a study probe, we conducted a formative evaluation of its web interface with end users and mental-health professionals to assess usability and interpretability. All procedures were approved by the Institutional Review Board (IRB) at University of Southern California.

\subsubsection{User Study} \label{sec:user_study}

We evaluated \textsc{CounselReflect} with 20 participants: 10 LLM mental-health support users and 10 individuals with prior professional counseling experience. Participants evaluated \textsc{CounselReflect} through the web interface using synthetic counseling transcripts based on CounselBench topics \citep{li2025counselbench} to avoid privacy and confidentiality concerns; generation details are provided in Appendix~\ref{human_study_design}. The study comprised a 30-minute semi-structured interview \cite{dejonckheere_vaughn_2019_semistructured_interview} and a 10-minute post-study survey. Participants identified important counseling dimensions, interpreted sampled metric outputs, and provided system feedback. The survey included the System Usability Scale (SUS) \cite{lewis_2018_sus}, Trust in Automation (TiA) scale \cite{jian_bisantz_drury_2000_tia}, and measures of satisfaction and transcript realism. Participants rated both transcript formats as reasonably authentic and representative of counseling interactions, reported high satisfaction (5.80 and 6.10) and overall trust (5.73 and 5.10) on 7-point scales, where 4 was the neutral point, and achieved SUS scores above the reference threshold of 68 in both groups (78.8 and 75.2) \cite{hyzy_bond_mulvenna_bai_dix_leigh_hunt_2022_sus_for_digital_health_apps}. Full study details and additional results are reported in Appendix~\ref{human_study_design}, Figure~\ref{fig:human_eval_results}, and Tables~\ref{tab:key_stats_overall}--\ref{tab:role_specific_scores}.

\subsubsection{Expert Review} \label{sec:expert}

We additionally evaluated \textsc{CounselReflect} with six mental-health professionals: one psychologist, two licensed therapists, two therapists under supervision, and one clinical psychology doctoral trainee. Using the same transcripts and protocol, clinicians highlighted supervision and training as promising use cases and valued both session-level summaries and turn-level feedback, reporting above-neutral satisfaction ($M=4.67$, $SD=1.03$) and trust ($M=5.01$, $SD=0.67$).

Together, these results indicated that \textsc{CounselReflect} was sufficiently usable and interpretable to support our subsequent study of reflection on AI-mediated mental-health and well-being conversations.

\section{Interview with Users of AI for Mental Health \& Wellbeing Support}

We used the browser-extension version of \textsc{CounselReflect} as a study probe to examine how people reflect on AI-mediated mental health and well-being conversations. Following a pilot study with \(N=4\) participants to refine the protocol, we conducted semi-structured interviews with \(N=21\) participants who use AI for mental health and well-being support. Participant recruitment and characteristics are reported in Appendix~\ref{appendix_participants}. The full interview protocol is enclosed in Appendix~\ref{sec:interview-protocol}.

All interviews were conducted remotely over Zoom and lasted approximately 60--75 minutes. Each session comprised three parts: (1) participants' current AI use and reflection practices (\S\ref{method_section1}); (2) a mock support conversation with a commonly used LLM chatbot (e.g., ChatGPT, Gemini, Claude), followed by self-reflection (\S\ref{method_section2}); and (3) reflection on the same conversation using \textsc{CounselReflect} (\S\ref{method_section3}). All participants provided informed consent, and the study protocol was approved by the Institutional Review Board (IRB) at University of Southern California.

\input{table/tab-participants}

\subsection{Part 1: Understanding Current AI Use and Reflection Practices}
\label{method_section1}
We first examined participants' current use of AI for mental health and well-being support, including motivations, perceived benefits and drawbacks, and concrete experiences. We then asked whether and how they reflected on past AI conversations; those without existing reflection practices described how they might evaluate the quality of AI support.

\subsection{Part 2: Mock AI Support Conversation and Self-Reflection}
\label{method_section2}
Participants then had a five-minute conversation with an LLM chatbot they commonly used about a topic they felt comfortable sharing. Using a think-aloud protocol, they discussed whether responses were helpful and how they might respond. We also asked how the interaction compared with seeking support from friends, family, or therapists. Afterwards, participants reflected on the conversation's overall quality, including whether AI addressed their needs, understood them, and provided trustworthy support.

\subsection{Part 3: Tool-Assisted Reflection with \textsc{CounselReflect}}
\label{method_section3}
Finally, participants reflected on the same conversation using \textsc{CounselReflect}. Before viewing its evaluation, they articulated questions they wanted to explore and selected any number of literature-grounded metrics (e.g., emotion, therapeutic alliance, motivational interviewing), with the option to define custom metrics. Participants explained their selections and could ask about unfamiliar concepts. They then reviewed the report's ``strengths'' and ``areas to review,'' assessed whether these insights aligned with their own impressions, and discussed whether the exercise changed their views of AI and whether they would use \textsc{CounselReflect} in practice.

\section{Results}

\subsection{Users' Own Approaches to Evaluating "Good Support"}
Participants varied in their preferred forms of support, ranging from \textit{emotion-oriented} support that helps them process difficult feelings to \textit{functional or solution-oriented} support that helps address problems affecting their mental health and well-being. % \todo{add examples}
These preferences also differed in temporal orientation. Participants favoring emotional support often valued being ``\textit{ground[ed] in the present moment}'' (P02) and helped to unpack difficult feelings, whereas those favoring solution-oriented support wanted guidance that could help them ``\textit{move forward and move on with [their] lives}'' (P18).

Across these preferences, however, participants consistently characterized good support as empathetic, emotionally validating, and attentive. A good supporter listens patiently and non-judgmentally and attempts to understand the support seeker's experience, such as asking relevant follow-up questions. 
In contrast, poor support centered the supporter who ``\textit{always go back to talk about [themselves]}'' (P04) rather than focusing on the support seeker, rushed to conclusions or solutions that felt ``\textit{overwhelming}'' or even ``\textit{irritating}'' (P16), normalized concerns through generic advice, or offered premature optimism such as ``\textit{everything will just be fine}'' (P02). Participants also cautioned against indiscriminate agreement. As P11 explained, ``\textit{being supportive is different from being indulgent. Sometimes, I might be worrying about the wrong thing. So, in those cases, I want to be proved wrong as well.}''

\subsection{Users' Current Reflection Practice}

The majority of participants (18 of 21) did not regularly reflect on their mental health and well-being conversations with AI. Even among those who occasionally revisited past conversations (P04, P05, P11, P16, P17), doing so was primarily for \textit{information retrieval} rather than reflection. Namely, they returned to retrieve useful information from past conversations; particularly, all five of these participants mentioned revisiting conversations where AI suggested CBT techniques that they were unfamiliar with but wanted to reference later.

Participants described two primary reasons for not reflecting. 
First, revisiting conversations could resurface stressful, painful, or traumatic experiences that they preferred not to re-experience. 
Second, participants often felt that the AI interaction had already helped them work through the problem; once they had ``moved on,'' they saw little reason to revisit it (P03). 
Additionally, P21 represented a distinct case: as they primarily used AI as a pointer to human support, they saw little value in revisiting past AI conversations once human support became available. They would ``\textit{confidently say that I graduated from AI help once I get in touch with my people.}''

As exceptions, three participants (P06, P08, P10) routinely reflected on their conversations; P06 even organized chats into folders to facilitate later review. 
P06 and P10 used past conversations to see ``\textit{how far [they] have gone since those difficult moments}'' and track progressions in their mental health symptoms over time (e.g., occurrence of depressive episodes). 
P08, who preferred solution-oriented support, revisited conversations to examine the outcomes of AI suggestions, specifically, whether AI-generated advice they followed ultimately helped resolve the situation.

\subsection{Inquiries for Self-Reflection}
Participants' verbal elaborations during the interviews surfaced several core questions they would want to explore when reflecting on their mental health and well-being conversations with AI. Rather than simply revisiting what was said, participants were interested in interrogating particular qualities of AI support after the immediate emotional burden of the situation had subsided. We present these inquiries in order of commonality:

\emph{Does AI actually ``understand'' and ``empathize'' with me?}
Because participants consistently valued empathy, emotional validation, and non-judgmental listening as core qualities of good support, 12 of 21 participants expressed interest in retrospectively examining whether AI had actually understood their struggles and emotions. Participants noted that revisiting conversations after difficult moments, when their thinking was less clouded by immediate emotional distress, could offer a different perspective on how well AI responded.
Importantly, participants were interested not only in making this judgment themselves, but also in using external tools to provide another perspective. P12, for example, wanted to see ``\textit{some numbers for emotional intelligence,}'' while others were broadly interested in whether AI genuinely appeared to ``\textit{understand}'' or ``\textit{empathize}'' with them (P05, P08, P14, P15, P18, and P19). 
Participants also connected emotional understanding to the appropriateness of AI's subsequent suggestions. For instance, P18 was motivated to perform self-reflection with tools, ``\textit{because AI keeps recommending me all these CBT techniques, I want to know whether these [suggestions] are indeed grounded in something ... like does it actually understand what I was going through?}''

\emph{How good is the quality of AI support?}
Most participants were interested in retrospectively assessing the overall quality of AI support, although they proposed varying criteria and little consensus for reflecting on AI support quality. These range from comparing AI responses against what human professionals might provide (P02, P11), assessing the safety, risks, and feasibility of AI suggestions (P04, P07, P19), fact-checking information provided by AI (P07, P10), evaluating whether responses were contextually relevant, reviewing the types of support AI provided (P14), and identifying important perspectives or information that AI might have missed (P21). 
Notably, many of the criteria participants spontaneously proposed overlapped with the dimensions incorporated into \textsc{CounselReflect}. 

\emph{Whether and how can I put AI suggestions into practice?}
Participants who found AI suggestions insightful were also interested in reflecting on whether and how to translate them into real-world practice. Rather than treating AI suggestions as one-off solutions, participants considered whether useful strategies could be adapted to other situations. P01, for example, found AI's suggestions for rethinking and reframing a stressful situation helpful for coping with anxiety. In reflecting on the conversation, they wanted to consider how these reframing exercises might be applied to other situations in their life and other forms of mental health symptoms that they experience (e.g., depression, phobia). Similarly, P10 learned about practicing breathing exercises during the onset of their panic disorder, and P12 learned about trauma processing from AI support; they likewise wanted to learn whether and how to practice these techniques in other applicable scenarios. 

\emph{Why did AI make me feel this way in a conversation?}
All participants acknowledged that receiving support from AI introduced unique experiences compared with receiving human support. For this reason, they wanted to reflect on what made AI support different. When AI support felt preferable, participants were curious about the ``\textit{realness} of the empathy'' they experienced (P01, P02, P05). 
Conversely, when AI made them feel worse or uncomfortable, participants wanted to understand what about the response produced that reaction and whether a human professional would respond similarly. P02, for example, recalled an unpleasant interaction with Claude in which its questions felt abrupt and uncomfortable. Through retrospective reflection and subsequent research, they came to understand these questions as related to suicide-risk assessment, which a therapist or counselor would not typically jump directly to such questions. 

Besides these primary inquiries, a few participants (P06, P17, P19) were interested in \textit{\ul{how they could interact with AI more effectively}}. Their reflection focused on how they might adjust future interactions to elicit more personalized support, such as P06's interest in phrasing prompts that could promote more culturally relevant responses.

Notably, only one participant (P08) spontaneously raised the broader question of \textit{\ul{How might seeking support from AI have an impact on me?}} While participants frequently wanted to evaluate the quality of AI responses or improve future interactions, they rarely considered the implications of repeated interactions and reliance on AI support. 

\subsection{How Users Use \textsc{CounselReflect}}

While most participants reacted positively to \textsc{CounselReflect}, we observed tensions in how they used it. Because systematic, critical reflection on AI conversations was new to many, participants often interpreted the report and selected metrics through their existing expectations of what AI support should provide, rather than as a neutral opportunity for critical reflection. 

\subsubsection{Reflection Can Reinforce Existing Beliefs About AI Support}

A common pattern was using \textsc{CounselReflect} to confirm why existing AI interactions felt helpful. Participants often focused on positive evaluations, particularly scores for emotional intelligence and empathy. P05, P15, and P16 explicitly described these results as validating qualities they already valued in AI support. P01 even used \textsc{CounselReflect}'s Q\&A feature to request examples of a ``\textit{bad response}'' after the report identified shortcomings in ChatGPT's responses to their struggles with phobia, concluding that ``\textit{if you consider how bad a bad response can be, ChatGPT is already doing a pretty good job.}'' 
P21 was a notable exception, using turn-by-turn analysis to examine both what AI did well and what was insufficient or missing.

\subsubsection{Reflection Predominantly Focuses on Emotional and Empathetic Support}

\input{table/tab-counselreflect-use}

A second pattern was participants' strong preference for reflecting on the emotional and empathetic dimensions of AI support. As shown in \Cref{tab:metric_selection_distribution}, emotion- and empathy-related metrics were selected substantially more often than other evaluation dimensions. This preference again reflected participants' primary expectations of AI support. 
In contrast, participants often considered factuality and alignment with clinical best practices ``\textit{not relevant to what [they] use AI for}'' as long as they were not explicitly seeking clinical advice (P11). Thus, participants' reflection tended to prioritize whether AI felt supportive over whether its responses met broader professional or clinical standards.
For example, P04 argued that evaluation should adapt to the conversational context: ``\textit{sometimes, I just want to vent [to AI]. I'm not sure if I need a textbook-like answer to what I'm venting for.}''

However, the boundary between clinical and non-clinical topics was less clear-cut in participants' actual conversations. P01 moved from discussing self-consciousness to asking about clinical definitions and diagnostic criteria for paranoia; P06 shifted from discussing a difficult romantic relationship to practicing behavioral therapeutic techniques; and P12 moved from discussing recent memories to trauma processing, chain analysis, and psychoeducation. These examples show how seemingly informal emotional support can gradually incorporate clinical concepts and therapeutic practices without users necessarily recognizing the interaction as clinical advice.

\subsubsection{Unclear Boundaries Between AI Support and Human Therapy}

Relatedly, a third challenge emerged around participants' understanding of the boundary between informal AI support and professional therapy. Although 17 of 21 participants described ``\textit{using AI as a therapist,}'' they often resisted evaluating AI against professional therapeutic standards. When \textsc{CounselReflect} gave relatively low scores for Talk Type, therapeutic techniques, or adherence to therapeutic best practices, participants commonly perceived the evaluation as overly stringent. As P20 responded, ``\textit{I think the tool is being a bit too harsh. I think AI actually did a pretty good job.}''

Participants' expectations of AI could also directly conflict with therapeutic best practices. P11, who frequently used AI for in-depth psychoanalysis, was disappointed by low scores for session management and the AI's failure to move the conversation forward. Yet this was precisely what they valued about AI: ``\textit{This is exactly why I like to do psychoanalyses with AI; I can just go on and on to talk about topics that I'm interested in.}'' Thus, even when participants positioned AI as a therapist, their conception of a ``\textit{good AI therapist}'' did not necessarily align with professional therapeutic practice.

\subsection{How Might Reflection Help?}

Most participants (15 of 21) were uncertain or considered it unlikely that reflection would substantially change how they use AI for mental health and well-being support. Nevertheless, participants identified benefits beyond changes in their AI use. To begin with, \textsc{CounselReflect} provided psychoeducational knowledge that helped participants better understand the support strategies and therapeutic practices embedded in AI responses. These insights sometimes introduced practices participants could carry into their everyday lives. P17, for example, found \textsc{CounselReflect}'s evaluation of communication skills and talk strategies useful for navigating tense conversations when family members experience emotional breakdowns. P05 similarly saw feedback about mindfulness techniques as potentially useful for coping with everyday stress.

Reflection could also help participants reinterpret AI responses that had initially felt unsupportive or even distressing. P21, who often tried to match others' emotional valence and found doing so stressful, was sometimes disappointed ``\textit{when AI itself seems so neutral.}'' Learning that managing emotional tone can be an intentional therapeutic practice helped them reconsider this behavior. Similarly, P02 described sometimes wondering ``\textit{if even AI gives up on me}'' when its responses felt inadequate. \textsc{CounselReflect} offered an alternative interpretation: ``\textit{looking at the results here, it reminds me that AI is not perfect either.}''

\section{Discussion}
% 3--4 pages

Our study shows that, although most users do not currently reflect on their mental health and well-being conversations with AI, there are opportunities to support such practices. Participants identified concrete questions they would like to revisit. However, designing reflection tools requires caution: users may selectively seek evidence that confirms their existing perceptions of AI or focus primarily on dimensions they already value. Reflection tools should therefore not merely facilitate users' existing inquiries, but also proactively surface blind spots and scaffold more holistic examination of AI interactions.

\subsection{Inquiries for Self-Reflection as Design Opportunities}

The self-reflection inquiries raised by participants offer concrete opportunities to design future reflection tools. Rather than organizing reflection solely around predefined evaluation metrics, tools could scaffold questions users themselves want to explore---such as whether AI understood and empathized with them, how good the support was, why an interaction made them feel a certain way, and whether and how AI suggestions should be carried into practice. The latter is particularly important, as participants saw reflection as most consequential when insights could inform how they cope, communicate, or seek support in everyday life. Future designs could therefore help users translate reflection outcomes into concrete practices beyond the immediate AI interaction.

At the same time, reflection tools should not be limited to questions users spontaneously raise. Only one participant explicitly considered the broader question of how repeatedly seeking support from AI might affect them. Tools could proactively scaffold reflection on such less salient questions, including changes in reliance on AI, support-seeking practices, or relationships with human supporters.

Participants' inquiries also reveal opportunities for \textit{AI literacy}. One of their most common questions---``\textit{Does AI understand and empathize with me?}''---points to uncertainty about how seemingly empathetic AI responses are generated. Rather than simply assigning an empathy score, reflection tools could use users' own questions as teachable moments to explain what such evaluations can and cannot establish about AI's underlying understanding. In this way, reflection can serve not only to evaluate individual conversations, but also to develop users' capacity to critically interpret AI-mediated support.

Along the same vein, participants' feedback, unclear boundaries between AI support and professional therapy, and shifting standards for evaluating AI point to opportunities to strengthen \textit{mental health literacy}. Reflection tools could help users better understand different forms of mental health support, what therapeutic practices are intended to accomplish, and when professional standards become relevant to an AI interaction. Rather than simply reminding users that ``AI is not a therapist,'' such tools could help them recognize when seemingly informal emotional support (such as through highlighting certain text in the interface) begins to involve clinical concepts or therapeutic practices, and understand what additional caution or evaluation these shifts warrant.

\subsection{Problematic Uses of \textsc{CounselReflect} Surface Challenges for Future Tool Design}

Our findings surface several challenges for future work to consider when designing tools that support self-reflection on AI-mediated mental health and well-being conversations. Participants sometimes used \textsc{CounselReflect} to confirm that AI had provided good support, selectively focused on metrics that aligned with their existing interests, and dismissed clinical or therapeutic metrics as ``\textit{not relevant.}'' Even participants who described using AI ``\textit{as a therapist}'' did not necessarily hold AI responses to the same standards they expected from human professionals. Together, these behaviors suggest that, without effective interventions, users may reproduce existing expectations and blind spots rather than broaden critical reflection.

Future reflection tools could therefore combine \textit{user-selected} and \textit{system-recommended} criteria. Alongside metrics users choose based on their goals, tools could consistently surface a small set of critical dimensions---such as factuality, potential risks, and appropriateness of support---when relevant. Rather than forcing users to inspect every available metric, such designs could explain why a particular dimension warrants attention, thereby exposing users to considerations they might otherwise overlook.

Importantly, which criteria are relevant may also change \textit{within} a conversation. We observed participants commonly drifting between support functions. 
Reflection criteria selected at the beginning may therefore become insufficient as the conversation evolves. Future tools should detect these shifts and nudge users to reconsider what should be evaluated; for example, recommending factuality or therapeutic-practice metrics when a conversation transitions from emotional validation to clinical guidance.

Finally, reflection tools could make the \textit{role users assign to AI} itself an object of reflection. Systems might periodically remind users of the role they initially expected AI to play and surface evaluation criteria appropriate to that role. When AI begins performing functions beyond that role, the tool could make the shift visible and invite users to reconsider both their expectations and the standards against which the interaction should be evaluated. In this way, reflection tools can move beyond evaluating individual responses toward helping users recognize how the role and stakes of AI support evolve throughout an interaction.

\subsection{Safeguards in Tool Design: Navigating the Fine Line Between Supporting Reflection and Re-Triggering Painful Experiences}

A major challenge is the emotional barrier to self-reflection. As participants described, revisiting past mental health and well-being conversations can mean returning to stressful, painful, or potentially traumatic moments that they have intentionally moved beyond. 
Thus, while reflection tools aim to help users unpack and make sense of their AI interactions, revisiting these conversations can arguably be one of the biggest obstacles to address.

Future designs should therefore treat reflection as an emotionally sensitive activity rather than simply an analytical one. Tools could give users greater control over when, how deeply, and which parts of a conversation they revisit, allow them to pause or skip potentially distressing content, and summarize patterns without requiring direct re-exposure to the original conversation. Reflection activities could also incorporate trauma-informed design principles and appropriate safeguards for engaging with distressing material. More broadly, reflection should remain optional and user-paced: choosing \textit{not} to revisit a difficult interaction may itself be an appropriate form of self-care rather than a failure to reflect.

%% file: table/tab-metric.tex
\begin{table*}[ht]
    \centering
        
    \caption{
    We summarize the 12 model-based metrics used in \textsc{CounselReflect}, including the underlying predictors, their provenance (Off-the-shelf, O: adopted from prior work; In-house, I: trained or fine-tuned by us) 
    and evaluation results where applicable. FactScore and MedScore are not evaluated because counseling dialogues lack suitable reference-based factuality annotations. Black entries are evaluated by us; gray entries are reported from the original works. $^*$ indicates LCS F1, $^{**}$ indicates SQuAD F1, and $^{***}$ indicates AUC.
    }
    \label{tab:model_metric_table}
    \begin{adjustbox}{max width=\linewidth}
    \begin{tabular}{lllc l ccc}
    \toprule
      Category & Metric & Model & \begin{tabular}[c]{@{}c@{}} Provenance \\ (O/I) \end{tabular} & Test Dataset & Accuracy & Macro F1 & Weighted F1 \\
      \midrule
      \multirow{3}{*}{Empathy} & Emotional Reactions (ER) & \multirow{3}{*}{EPITOME \citep{sharma-etal-2020-computational-empathy}} & I & \multirow{3}{*}{Mental Health Subreddits \cite{sharma-etal-2020-computational-empathy}} & 80.84\% & 0.77 & - \\
      & Interpretations (IP) & & I & & 92.21\% & 0.56 & - \\
      & Explorations (EX) & & I & & 85.55\% & 0.58 & - \\
      \midrule
      \multirow{2}{*}{Emotion} & Emotion Classification & Emotion English Roberta Large \cite{hartmann2022emotionenglish} & O & \multirow{2}{*}{DailyDialog \cite{li2017dailydialogmanuallylabelledmultiturn}} & 57.92\% & 0.28 & 0.66 \\
      & Emotion Triggers & Reccon \cite{poria2021recognizingemotioncauseconversations} & I & & - & 0.75$^{*}$ & 0.75$^{**}$ \\ 
      \midrule
      Talk Type & Client Talk Type & roberta-base \cite{liu2019robertarobustlyoptimizedbert} & I & AnnoMI \cite{anno_mi_dataset} & 70.12\% & 0.58 & - \\ 
      \midrule
      Support Strategy & Support Strategy & roberta-base \cite{liu2019robertarobustlyoptimizedbert} & I & AnnoMI \cite{anno_mi_dataset} & 57.41\% & 0.50 & 0.56 \\ 
      \midrule
      Reflection & Reflection & PAIR \cite{min-etal-2022-pair} & O & AnnoMI \cite{anno_mi_dataset} & 70.8\% & 0.36 & 0.61 \\ 
      \midrule
      \multirow{2}{*}{Toxicity} & Toxicity (Perspective) & PerspectiveAPI \cite{lees2022newgenerationperspectiveapi} & O & \textcolor{gray}{TweetEval hate \cite{barbieri_camacho_collados_espinosa_anke_neves_2020_tweeteval}} & \textcolor{gray}{-} & \textcolor{gray}{0.55} & \textcolor{gray}{-} \\ 
      & Toxicity (DeToxify) & DeToxify \cite{Detoxify} & O & \textcolor{gray}{Civil Comments \cite{civil_comments_dataset}} & \textcolor{gray}{$93.74^{***}$} & \textcolor{gray}{-} & \textcolor{gray}{-} \\
      \midrule
      \multirow{2}{*}{Factuality} & FactScore & Factscore \cite{factscore} & O & - & - & - & - \\
      & Medscore & Medscore \cite{huang2025medscoregeneralizablefactualityevaluation} & O & - & - & - & - \\
    \bottomrule
    \end{tabular}
    \end{adjustbox} 
\end{table*}

\begin{table*}
    \centering
    \caption{
    We summarize the 69 literature-derived metrics used in \textsc{CounselReflect}.
    }
    \label{tab:rubric_metric_table}
    \begin{adjustbox}{max width=\linewidth}
    % \tiny
    \begin{tabular}{l c p{15cm}}
    \toprule
    Category & N & Metrics \\
    \midrule
    \textbf{Core Conditions} & 10 & 
(1) Empathy \cite{empathy_1, empathy_2, empathy_3, empathy_4_empathic_responding_4, empathy_5}; 
(2) Therapeutic Alliance \cite{therapeutic_alliance_1, therapeutic_alliance_2_open_ended_questions_1_engagement_participation_facilitation_5, therapeutic_alliance_3, therapeutic_alliance_4, therapeutic_alliance_5}; \newline
(3) Therapist Genuineness \cite{wampold_2015_how_important_are_the_common_factors_in_psychotherapy, therapist_genuineness_2, therapist_genuineness_3, therapist_genuineness_4, therapist_genuineness_5}; 
(4) Unconditional Positive Regard \cite{unconditional_positive_regard_1, unconditional_positive_regard_2, unconditional_positive_regard_3, unconditional_positive_regard_4, unconditional_positive_regard_5}; \newline
(5) Therapist Validation \cite{therapist_validation_1_therapeutic_concreteness_5, therapist_validation_2, therapist_validation_3, therapist_validation_4, therapist_validation_5}; 
(6) Affective Attunement \cite{affective_attunement_1, affective_attunement_2, affective_attunement_3, affective_attunement_4, affective_attunement_5}; \newline
(7) Cultural Humility \cite{cultural_humility_1, cultural_humility_2, cultural_humility_3_dialectical_strategies_2, cultural_humility_4, cultural_humility_5};  
(8) Empathic Responding \cite{empathic_responding_1, empathic_responding_2, empathic_responding_3, empathy_4_empathic_responding_4, empathic_responding_5}; \newline
(9) Emotional Support Provision \cite{emotional_support_provision_1, emotional_support_provision_2, emotional_support_provision_3, emotional_support_provision_4, emotional_support_provision_5};  
(10) Nonjudgmental Stance \cite{nonjudgmental_stance_1, nonjudgmental_stance_2, nonjudgmental_stance_3, boundary_setting_2_nonjudgmental_stance_4, nonjudgmental_stance_5}. \\
    \midrule
    \textbf{Communication Skills} & 10 & 
(1) Reflective Listening \cite{reflective_listening_1, reflective_listening_2, reflective_listening_3, reflective_listening_4, reflective_listening_5}; 
(2) Open-Ended Questions \cite{therapeutic_alliance_2_open_ended_questions_1_engagement_participation_facilitation_5, open_ended_questions_2, open_ended_questions_3, open_ended_questions_4, socratic_questioning_1_open_ended_questions_5}; \newline
(3) Affirmation \cite{affirmation_1_strength_identification_5, affirmation_2_normalizing_1, affirmation_3, affirmation_4, affirmation_5_autonomy_support_2_change_talk_1}; 
(4) Summarization \cite{summarization_1, summarization_2, summarization_3, summarization_4_patientcentered_personcentered_language_3, summarization_5}; \newline
(5) Normalizing \cite{affirmation_2_normalizing_1, normalizing_2, normalizing_3, normalizing_4, normalizing_5}; 
(6) Therapeutic Concreteness \cite{therapeutic_concreteness_1, therapeutic_concreteness_2, therapeutic_concreteness_3, therapeutic_concreteness_4, therapist_validation_1_therapeutic_concreteness_5}; \newline
(7) Clarification \cite{clarification_1, clarification_2, clarification_3_values_clarification_2_autonomy_support_1, clarification_4, clarification_5}; 
(8) Patient-Centered / Person-Centered Language \cite{patientcentered_personcentered_language_1, patientcentered_personcentered_language_2, summarization_4_patientcentered_personcentered_language_3, patientcentered_personcentered_language_4, patientcentered_personcentered_language_5}; \newline
(9) Non-Stigmatizing / Person-First Language \cite{nonstigmatizing_personfirst_language_1, nonstigmatizing_personfirst_language_2, nonstigmatizing_personfirst_language_3, nonstigmatizing_personfirst_language_4, nonstigmatizing_personfirst_language_5}; \newline
(10) Personalization to Client Context \cite{personalization_to_client_context_1, personalization_to_client_context_2, personalization_to_client_context_3, personalization_to_client_context_4, personalization_to_client_context_5}. \\
    \midrule
    \textbf{CBT Techniques} & 11 & 
(1) Socratic Questioning \cite{socratic_questioning_1_open_ended_questions_5, socratic_questioning_2, socratic_questioning_3, socratic_questioning_4, socratic_questioning_5}; 
(2) Cognitive Reframing \cite{cognitive_reframing_1, cognitive_reframing_2, cognitive_reframing_3, cognitive_reframing_4, cognitive_reframing_5}; \newline
(3) Psychoeducation \cite{psychoeducation_1, psychoeducation_2, psychoeducation_3, psychoeducation_4, psychoeducation_5};  
(4) Homework Assignment \cite{homework_assignment_1, homework_assignment_2, homework_assignment_3, homework_assignment_4, homework_assignment_5}; \newline
(5) Cognitive Defusion \cite{cognitive_defusion_1, cognitive_defusion_2, cognitive_defusion_3, cognitive_defusion_4, cognitive_defusion_5};  
(6) Behavioral Activation \cite{behavioral_activation_1, behavioral_activation_2, behavioral_activation_3, behavioral_activation_4, behavioral_activation_5}; \newline
(7) Chain Analysis \cite{chain_analysis_1, chain_analysis_2, chain_analysis_3, chain_analysis_4, chain_analysis_5}; 
(8) Sleep Hygiene Education \cite{sleep_hygiene_education_1, sleep_hygiene_education_2, sleep_hygiene_education_3, sleep_hygiene_education_4, sleep_hygiene_education_5}; \newline
(9) Social Skills Training \cite{assertiveness_training_1_social_skills_training_1, assertiveness_training_2_social_skills_training_2, assertiveness_training_3_social_skills_training_3, social_skills_training_4_assertiveness_training_4, social_skills_training_5_assertiveness_training_5}; 
(10) Assertiveness Training \cite{assertiveness_training_1, assertiveness_training_2, assertiveness_training_3, assertiveness_training_4, assertiveness_training_5}; \newline
(11) Identifying Core Beliefs \cite{identifying_core_beliefs_1, identifying_core_beliefs_2, identifying_core_beliefs_3, identifying_core_beliefs_4, identifying_core_beliefs_5}. \\
    \midrule
    \textbf{Relationship Repair} & 2 & 
(1) Alliance Rupture Repair \cite{alliance_rupture_repair_1, alliance_rupture_repair_2, alliance_rupture_repair_3, alliance_rupture_repair_4, alliance_rupture_repair_5}; 
(2) Therapist Self-Disclosure \cite{therapist_self_disclosure_1, therapist_self_disclosure_2, therapist_self_disclosure_3, therapist_self_disclosure_4, therapist_self_disclosure_5}. \\
    \midrule
    \textbf{Session Management} & 9 & 
(1) Goal Consensus \cite{goal_consensus_1, goal_consensus_2, goal_consensus_3, wampold_2015_how_important_are_the_common_factors_in_psychotherapy, goal_consensus_5};
(2) Agenda Setting \cite{agenda_setting_1, agenda_setting_2, agenda_setting_3, agenda_setting_4, agenda_setting_5}; \newline
(3) Feedback Solicitation \cite{feedback_solicitation_1, feedback_solicitation_2, feedback_solicitation_3, feedback_solicitation_4, feedback_solicitation_5};
(4) Client Monitoring / Progress Tracking \cite{client_monitoring_progress_tracking_1, client_monitoring_progress_tracking_2, client_monitoring_progress_tracking_3, client_monitoring_progress_tracking_4, client_monitoring_progress_tracking_5}; \newline
(5) Boundary Setting \cite{boundary_setting_1, boundary_setting_2_nonjudgmental_stance_4, boundary_setting_3, boundary_setting_4, boundary_setting_5};
(6) Termination Preparation \cite{termination_preparation_1, termination_preparation_2, termination_preparation_3, termination_preparation_4, termination_preparation_5}; \newline
(7) Managing Resistance \cite{managing_resistance_1, managing_resistance_2, managing_resistance_3, countertransference_management_1_managing_resistance_4, managing_resistance_5};
(8) Shared Decision-Making Support \cite{shared_decisionmaking_support_1, shared_decisionmaking_support_2, shared_decisionmaking_support_3, shared_decisionmaking_support_4, shared_decisionmaking_support_5}; \newline
(9) Engagement \& Participation Facilitation \cite{engagement_participation_facilitation_1, engagement_participation_facilitation_2, engagement_participation_facilitation_3, engagement_participation_facilitation_4, therapeutic_alliance_2_open_ended_questions_1_engagement_participation_facilitation_5}. \\
    \midrule
    \textbf{MI Techniques} & 3 & 
(1) Autonomy Support \cite{clarification_3_values_clarification_2_autonomy_support_1, affirmation_5_autonomy_support_2_change_talk_1, autonomy_support_3, autonomy_support_4, autonomy_support_5}; 
(2) Change Talk \cite{affirmation_5_autonomy_support_2_change_talk_1, change_talk_2, change_talk_3, change_talk_4, change_talk_5}; \newline
(3) Strength Identification \cite{strength_identification_1, strength_identification_2, strength_identification_3, strength_identification_4, affirmation_1_strength_identification_5}. \\
    \midrule
    \textbf{Advanced Skills} & 12 & 
(1) Instillation of Hope \cite{instillation_of_hope_1, instillation_of_hope_2, instillation_of_hope_3, instillation_of_hope_4, instillation_of_hope_5}; 
(2) Experiential Learning / Role-Playing \cite{experiential_learning_role_playing_1, experiential_learning_role_playing_2, experiential_learning_role_playing_3, experiential_learning_role_playing_4, experiential_learning_role_playing_5}; \newline
(3) Problem-Solving \cite{problem_solving_1, problem_solving_2, problem_solving_3, problem_solving_4, problem_solving_5}; 
(4) Emotion Regulation Skills Training \cite{emotion_regulation_skills_training_1, emotion_regulation_skills_training_2, emotion_regulation_skills_training_3, emotion_regulation_skills_training_4, emotion_regulation_skills_training_5}; \newline
(5) Relapse Prevention Planning \cite{relapse_prevention_planning_1, relapse_prevention_planning_2, relapse_prevention_planning_3, relapse_prevention_planning_4, relapse_prevention_planning_5}; 
(6) Humor in Therapy \cite{humor_in_therapy_1, humor_in_therapy_2, humor_in_therapy_3, humor_in_therapy_4, humor_in_therapy_5}; \newline
(7) Here-and-Now Processing \cite{hereandnow_processing_1, hereandnow_processing_2, hereandnow_processing_3, hereandnow_processing_4, hereandnow_processing_5};
(8) Narrative Reframing \cite{narrative_reframing_1, narrative_reframing_2, narrative_reframing_3, narrative_reframing_4, narrative_reframing_5}; \newline
(9) Externalization of Problem \cite{externalization_of_problem_1, externalization_of_problem_2, externalization_of_problem_3, externalization_of_problem_4, externalization_of_problem_5}; 
(10) Psychodynamic Interpretation \cite{psychodynamic_interpretation_1, psychodynamic_interpretation_2, psychodynamic_interpretation_3, psychodynamic_interpretation_4, psychodynamic_interpretation_5}; \newline
(11) Countertransference Management \cite{countertransference_management_1_managing_resistance_4, countertransference_management_2, countertransference_management_3, countertransference_management_4, countertransference_management_5}; 
(12) Dialectical Strategies \cite{dialectical_strategies_1, cultural_humility_3_dialectical_strategies_2, dialectical_strategies_3, dialectical_strategies_4, dialectical_strategies_5}. \\
    \midrule
    \textbf{Solution-Focused} & 3 & 
(1) Solution-Building Questions \cite{solutionbuilding_questions_1, solutionbuilding_questions_2, solutionbuilding_questions_3, solutionbuilding_questions_4, solutionbuilding_questions_5}; 
(2) Positive Reinforcement \cite{positive_reinforcement_1, positive_reinforcement_2, brookman-positive_reinforcement_3, positive_reinforcement_4, positive_reinforcement_5}; \newline
(3) Values Clarification \cite{values_clarification_1, clarification_3_values_clarification_2_autonomy_support_1, values_clarification_3, values_clarification_4, values_clarification_5}. \\
    \midrule
    \textbf{Mindfulness \& Body} & 3 & 
(1) Mindfulness Induction \cite{mindfulness_induction_1, mindfulness_induction_2, mindfulness_induction_3, mindfulness_induction_4, mindfulness_induction_5}; 
(2) Body-Oriented Interventions \cite{bodyoriented_interventions_1, bodyoriented_interventions_2, bodyoriented_interventions_3, bodyoriented_interventions_4, bodyoriented_interventions_5}; \newline
(3) Grounding Techniques \cite{grounding_techniques_1, grounding_techniques_2, grounding_techniques_3, grounding_techniques_4, grounding_techniques_5}. \\
    \midrule
    \textbf{Emotion Processing} & 4 & 
(1) Self-Compassion Promotion \cite{selfcompassion_promotion_1, selfcompassion_promotion_2, selfcompassion_promotion_3, selfcompassion_promotion_4, selfcompassion_promotion_5}; 
(2) Emotional Processing Facilitation \cite{emotional_processing_facilitation_1, emotional_processing_facilitation_2, emotional_processing_facilitation_3, emotional_processing_facilitation_4, emotional_processing_facilitation_5}; \newline
(3) Emotion Focused Skills \cite{emotion_focused_skills_1, emotion_focused_skills_2, emotion_focused_skills_3, emotion_focused_skills_4, emotion_focused_skills_5}; 
(4) Grief Processing \cite{grief_processing_1, grief_processing_2, grief_processing_3, grief_processing_4, grief_processing_5}. \\
    \midrule
    \textbf{Crisis \& Trauma} & 2 & 
(1) Trauma Processing \cite{trauma_processing_1, trauma_processing_2, trauma_processing_3, trauma_processing_4, trauma_processing_5}; 
(2) Safety Planning \cite{safety_planning_1, safety_planning_2, safety_planning_3, safety_planning_4, safety_planning_5}. \\
    \bottomrule
    \end{tabular}
    \end{adjustbox}
\end{table*}

%% file: table/tab-participants.tex
\begin{table}[!ht]
\centering
\small
\setlength{\tabcolsep}{4pt}
\caption{Participants' common AI use types and topics for mental health and wellbeing support.}
\label{tab:participant}
\begin{tabular}{L{0.04\textwidth} L{0.3\textwidth} L{0.2\textwidth} L{0.15\textwidth} L{0.15\textwidth}}
\toprule
\textbf{ID} & \textbf{Common AI support topics} & \textbf{Mental health \& other health concerns} & \textbf{Primary AI in use} & \textbf{Length of AI use} \\
\midrule
Pilot~1 & Academic stress, job search, mindfulness & - & ChatGPT & Cannot recall \\
\addlinespace

Pilot~2 & Work stress, reassurance & - & ChatGPT & Infrequent \\
\addlinespace

Pilot~3 & Interpersonal conflicts, emotional support & - & Gemini & Cannot recall \\
\addlinespace

Pilot~4 & Job stress, understanding emotions & - & ChatGPT & Occasional \\
\midrule

P01 & Social anxiety, reporting progress & Agoraphobia, depression, anxiety & Character.ai, Copilot & 3 years \\
\addlinespace

P02 & Job search, self-esteem & Anxiety, depression & Gemini & 6 months -- 1 year \\
\addlinespace

P03 & Managing ADHD symptoms & ADHD, anxiety & Duck.ai, Lumo & 1.5 -- 2 years \\
\addlinespace

P04 & Emotional support, procrastination & Anxiety & ChatGPT & Cannot recall \\
\addlinespace

P05 & Workplace stress, job search & - & Gemini & 1 year \\
\addlinespace

P06 & Relationships, work stress, emotional support & Depression & ChatGPT & 1.5 -- 2 years \\
\addlinespace

P07 & Interpersonal conflicts, health queries & Social anxiety & ChatGPT & 2 years \\
\addlinespace

P08 & Health queries, conflict resolution & - & ChatGPT & Cannot recall \\
\addlinespace

P09 & Decision-making, progress-tracking & - & ChatGPT & 3 years \\
\addlinespace

P10 & Health management, career stress & Panic disorder, pneumonia & Claude & Cannot recall \\
\addlinespace

P11 & Self-reflection, understanding emotions & - & ChatGPT & Cannot recall \\
\addlinespace

P12 & education, professional development, immigration & Trauma & ChatGPT & Cannot recall \\
\addlinespace

P13 & Emotional validation, health queries, relationship advice & - & Gemini & Recent \\
\addlinespace

P14 & Grief, interpersonal conflict & - & Gemini & 3 years \\
\addlinespace

P15 & Informational and emotional support & - & Snapchat AI, ChatGPT & 1.5 years \\
\addlinespace

P16 & Practical and actionable suggestions & - & ChatGPT & 2 years \\
\addlinespace

P17 & Researching, emotional validation & - & ChatGPT & 2 months \\
\addlinespace

P18 & Relationship support, articulation, problem-solving & - & ChatGPT & Cannot recall \\
\addlinespace

P19 & Emotional support, interpersonal conflict, and job stress & - & ChatGPT & - \\
\addlinespace

P20 & Articulating perspectives, understanding emotions & - & ChatGPT & Cannot recall \\
\addlinespace

P21 & Academic stress, relationship advice & - & ChatGPT & For 3 years \\
\bottomrule
\end{tabular}
\end{table}

%% file: table/tab-counselreflect-use.tex
\begin{table*}[h!]
\centering
\small
\caption{Distribution of evaluation metrics selected by participants ($N=21$).
Model-based metrics are counted individually, while literature-derived metrics
are grouped by their highest-level rubric category. Participants could select
multiple metrics or categories; therefore, percentages do not sum to 100\%.}
\label{tab:metric_selection_distribution}

\begin{tabularx}{\textwidth}{
    @{} X r X r @{}
}
\toprule
\multicolumn{2}{c}{\textbf{Model-based Metrics}}
&
\multicolumn{2}{c}{\textbf{Literature-derived / Rubric-based Categories}}
\\
\cmidrule(lr){1-2}
\cmidrule(lr){3-4}

\textbf{Metric} & \textbf{$n$ (\%)}
& \textbf{Category} & \textbf{$n$ (\%)} \\
\midrule

Emotional Triggers (RECCON)      & 12 (57.1\%) &
Communication Skills             & 10 (47.6\%) \\

Emotional Support Strategy       & 12 (57.1\%) &
Solution-Focused                 & 9 (42.9\%) \\

Emotion Classification           & 11 (52.4\%) &
Core Conditions                  & 7 (33.3\%) \\

PAIR Reflection Scorer           & 11 (52.4\%) &
Emotional Processing             & 7 (33.3\%) \\

FactScore                         & 10 (47.6\%) &
Mindfulness \& Body              & 5 (23.8\%) \\

Empathy ER (Emotional Reaction)  & 10 (47.6\%) &
CBT Techniques                   & 4 (19.0\%) \\

Empathy IP (Interpretation)      & 10 (47.6\%) &
Relationship Repair              & 3 (14.3\%) \\

Talk Type (Change/Neutral/Sustain) & 9 (42.9\%) &
Advanced Skills                  & 2 (9.5\%) \\

Empathy EX (Exploration)         & 7 (33.3\%) &
MI Techniques                    & 1 (4.8\%) \\

Toxicity (Detoxify ML)           & 7 (33.3\%) &
Crisis \& Trauma                 & 1 (4.8\%) \\

MedScore                          & 5 (23.8\%) &
                                 & \\

\bottomrule
\end{tabularx}
\end{table*}

%% file: paper_appendix.tex
\section*{Appendix}

\section{Evaluation Metrics and Resources}

\subsection{Model-Based Metrics}
\label{appendix_model_based_metrics}

\paragraph{Empathy} is a foundational component of effective counseling \citep{elliott_bohart_watson_greenberg_2011_empathy, rogers_1957_the_necessary_and_sufficient_conditions_of_therapeutic_personality_change}. We include the computational empathy framework proposed by \citet{sharma-etal-2020-computational-empathy}, which decomposes empathic responses into emotional reactions, interpretations, and explorations. 

\paragraph{Emotion Identification} We use a RoBERTa-large model trained on a combination of six emotion classification datasets from HuggingFace \cite{hartmann2022emotionenglish}. This module helps identify affective states expressed by clients \citep{lazarus_atzil_slonim_bar_kalifa_hasson_ohayon_rafaeli_2019_clients_emotional_instability_and_therapists_inferential_flexibility_predict_therapists_session_by_session_empathetic_accuracy, atzil_slonim_bar_kalifa_fisher_lazarus_hasson_ohayon_lutz_rubel_rafaeli_2019_therapists_empathic_accuracy_accuracy_towards_their_clients_emotions}.

\paragraph{Emotional Trigger} In addition to emotion identification,  identifying emotional triggers provides additional context for understanding client affect \citep{liu_wei_tu_lin_jiang_cambria_2025_knowing_what_and_why, poria2021recognizingemotioncauseconversations}.  We utilize RECCON \cite{poria2021recognizingemotioncauseconversations} to detect emotion-cause relationships in dialogue.

\paragraph{Reflection} 

To assess reflective quality in therapist responses, we adopt PAIR \citep{min-etal-2022-pair}, which follows standard MI categories for reflection: non-, simple, or complex reflection.

\paragraph{Supporter Strategy}
We train a RoBERTa-based classifier on the ESConv dataset \citep{liu2021emotionalsupportdialogsystems} to identify supporter strategies in therapist utterances. ESConv specifies supportive conversational behaviors, such as affirmation and reassurance, based on Hill’s Helping Skills Theory \citep{hill2009helping}.

\paragraph{Talk Type}
We train a separate RoBERTa-based classifier on the AnnoMI dataset \cite{anno_mi_dataset} to predict motivational interviewing talk types at the utterance level.

\paragraph{Toxicity}
We measure potentially harmful language using two independent systems: the Perspective API \citep{lees2022newgenerationperspectiveapi} and Detoxify \citep{Detoxify}. The Perspective API outputs 0-1 scores for toxicity-related attributes (e.g., insult, threat, identity attack), while Detoxify is a Transformer-based classifier trained on Jigsaw Toxic Comment datasets.

\paragraph{Factuality}
We assess factual consistency using FactScore \citep{factscore} and MedScore \citep{huang2025medscoregeneralizablefactualityevaluation}. 
FactScore evaluates general factual alignment using Wikipedia as the reference corpus, while MedScore targets medical accuracy by grounding claims in authoritative clinical sources\footnote{We use evidence from medical textbooks for MedScore considering the retrieval efficiency.}. Both metrics use the model selected on the configuration page. 

\subsection{Licensing for Datasets}
\label{sec:licensing}

\emph{EPITOME \cite{sharma-etal-2020-computational-empathy}} is publicly released via \href{https://github.com/behavioral-data/Empathy-Mental-Health}{Github}, which includes two license files: an academic non-commercial research license and an attribution-based redistribution license. Our use follows the licensing terms in the repository.

\emph{Emotion \cite{hartmann2022emotionenglish}} uses a public Hugging Face model checkpoint (emotion-english-roberta-large). No explicit license information is provided in the  repository at the time of access. 

\emph{Reccon \cite{poria2021recognizingemotioncauseconversations}} is publicly released via \href{https://github.com/declare-lab/RECCON/tree/main}{Github} without an explicit license file. Our experiments use the DailyDialog subset, which is released under the CC BY-NC-SA 4.0 license.

\emph{AnnoMI \cite{anno_mi_dataset}} is publicly released via \href{https://github.com/uccollab/AnnoMI}{GitHub} without an explicit license file. Our use follows the repository usage and citation guidelines for research purposes.

\emph{Pair \cite{min-etal-2022-pair}} is publicly released via \href{https://github.com/MichiganNLP/PAIR}{GitHub} without an explicit license file. Our use follows the repository usage and citation guidelines for research purposes.

\emph{Factscore \cite{factscore}} is publicly released via \href{https://github.com/shmsw25/FActScore}{GitHub} under the MIT License.

\emph{Perspective API \cite{lees2022newgenerationperspectiveapi}} is accessed under Perspective’s API Terms of Service and related platform terms. API keys are treated as sensitive credentials and are not shared publicly; our use follows the provider’s terms.

\emph{Detoxicify \cite{Detoxify}} is released via \href{https://github.com/unitaryai/detoxify}{GitHub} under the Apache-2.0 License.

\emph{Emotion Support Strategy \cite{liu2021emotionalsupportdialogsystems}} is publicly released via \href{https://github.com/thu-coai/Emotional-Support-Conversation}{GitHub} under the CC BY-NC 4.0 license.

\emph{Medscore \cite{huang2025medscoregeneralizablefactualityevaluation}} implementation is publicly available via \href{https://github.com/Heyuan9/MedScore}{GitHub} without an explicit license file. \\
\\
For resources without explicit license files, usage is limited to research evaluation and follows repository usage guidelines.

\section{Study 1: Usability Validation Details}
% =========================
% Table 4: Key stats (single-column)
% =========================
\begin{table}[]
\centering
\caption{ Statistics Across User Groups (Mean $\pm$ SD).}
\label{tab:key_stats_overall}
\tiny
\setlength{\tabcolsep}{3pt}
\renewcommand{\arraystretch}{0.92}
\resizebox{\columnwidth}{!}{%
\begin{tabular}{lccc}
\toprule
\textbf{Metric} & \textbf{LLM Support Users} ($n=10$) & \textbf{Prior Counseling Recipients} ($n=10$) &  \\
\midrule
Authenticity (auth) & 5.50 $\pm$ 1.08 & 5.10 $\pm$ 1.66 \\
Resemblance (resemble) & 5.20 $\pm$ 1.81 & 4.70 $\pm$ 1.70  \\
Satisfaction & 5.80 $\pm$ 0.79 & 6.10 $\pm$ 0.99 \\
\midrule
\textbf{Usability (SUS)} & & & \\
SUS Total & 78.75 $\pm$ 10.43 & 75.25 $\pm$ 11.08\\
\midrule
\textbf{Trust (TIA)} & & & \\
TIA Distrust & 1.84 $\pm$ 0.98 & 2.58 $\pm$ 1.05  \\
TIA Trust & 5.41 $\pm$ 1.00 & 4.87 $\pm$ 1.11\\
TIA Overall-trust & 5.73 $\pm$ 0.84 & 5.10 $\pm$ 1.02\\
\bottomrule
\end{tabular}%
}
% \vspace{-2mm} % tighten gap before Table 5 (adjust: -1mm to -3mm if needed)

\end{table}

% =========================
\begin{table}[]
\centering

\label{tab:role_specific_compact}

\caption{Role-specific items (Mean $\pm$ SD).}
\label{tab:role_specific_scores}
% \small
\setlength{\tabcolsep}{2.2pt}
\renewcommand{\arraystretch}{0.92}

\begin{tabular}{cc|cc}
\toprule
\multicolumn{2}{c}{\textbf{Prior Counseling Recipients}} &
\multicolumn{2}{c}{\textbf{LLM Support Users}} \\
\cmidrule(r){1-2}\cmidrule(l){3-4}
\textbf{ID} & \textbf{Mean$\pm$SD} &
\textbf{ID} & \textbf{Mean$\pm$SD} \\
\midrule
Q26 & 5.4 $\pm$ 1.8 & Q36 & 5.5 $\pm$ 1.2 \\
Q27 & 5.5 $\pm$ 1.5 & Q37 & 5.2 $\pm$ 1.1 \\
Q28 & 5.0 $\pm$ 2.2 & Q38 & 5.1 $\pm$ 1.1 \\
Q29 & 5.9 $\pm$ 0.7 & Q39 & 5.5 $\pm$ 0.7 \\
Q30 & 2.1 $\pm$ 0.7 & Q40 & 3.1 $\pm$ 2.2 \\
Q31 & 4.3 $\pm$ 1.9 & Q41 & 4.4 $\pm$ 1.3 \\
Q32 & 6.0 $\pm$ 0.7 & Q42 & 5.4 $\pm$ 1.5 \\
Q33 & 2.4 $\pm$ 1.5 & Q43 & 2.2 $\pm$ 1.5 \\
Q34 & 2.0 $\pm$ 1.2 & Q44 & 6.2 $\pm$ 0.8 \\
Q35 & 2.7 $\pm$ 1.5 & Q45 & 1.9 $\pm$ 0.7 \\
 &  & Q46 & 5.5 $\pm$ 0.7 \\
 &  & Q47 & 5.9 $\pm$ 0.7 \\
 &  & Q48 & 5.6 $\pm$ 1.5 \\
 &  & Q49 & 6.1 $\pm$ 0.7 \\
 &  & Q50 & 2.5 $\pm$ 1.3 \\
 &  & Q51 & 2.3 $\pm$ 0.9 \\
\bottomrule
\end{tabular}
\end{table}

\subsection{Study Design}
\label{human_study_design}
\paragraph{Synthetic Counseling Transcript}
We use synthetic transcripts to evaluate \textsc{CounselReflect} under controlled conditions while avoiding privacy and confidentiality concerns associated with real counseling data. All transcripts are generated with LLMs using two prompting strategies that produce two transcript formats used in our study. Following the topic selection procedure of CounselBench \citep{li2025counselbench}, we select the 10 most popular topics and use one representative question--answer pair per topic as the seed. In the first strategy, we prompt GPT-4o to generate a complete multi-turn support dialogue conditioned on the seed (30 turns). In the second strategy, we generate an interactive chat transcript by assigning two LLM roles: GPT-4o simulates the user while ChatGPT plays the assistant, iterated for 10 turns by relaying messages between the two models. 

To verify suitability for user evaluation, participants rated authenticity and resemblance to real counseling interactions on a 7-point scale. The multi-turn transcripts received mean scores of 5.10 (authenticity) and 4.70 (resemblance), and the interactive chat transcripts received 5.50 and 5.20, respectively, indicating that the transcripts were perceived as reasonably realistic.

\paragraph{Procedure}
The evaluation began with a standardized orientation and a baseline reading of a randomized counseling transcript. Before using the system, participants identified five priority metrics for auditing the conversation and checked whether these were included among \textsc{CounselReflect}’s default metrics; any omissions were documented and used to test the custom metric generator. To ensure varied exposure, each participant evaluated two randomly sampled pre-trained metrics and three literature-derived metrics. In the main phase, they reviewed session-level summaries and turn-level temporal metric shifts via the Reporting Interface. The study concluded with a qualitative interview on UI friction and perceived insight reliability, followed by a post-study survey (Table \ref{appendix:full_survey}).

% \begin{table}[]
% \centering
% \caption{Full Survey Instrument (Q1 - Q51).}
% \label{appendix:full_survey}
% % \tiny
% \setlength{\tabcolsep}{1.5pt}
% \renewcommand{\arraystretch}{0.72}
% \adjustbox{width=\linewidth}{
% \begin{tabularx}{\linewidth}{p{0.06\linewidth} p{0.16\linewidth} >{\RaggedRight\arraybackslash}X}
% \toprule
% \textbf{ID} & \textbf{Scale} & \textbf{Item} \\
% \midrule
{\small
\begin{longtable}{
    p{0.06\linewidth}
    p{0.16\linewidth}
    >{\RaggedRight\arraybackslash}p{0.72\linewidth}
}
\caption{Full Survey Instrument (Q1--Q51).}
\label{appendix:full_survey} \\
\toprule
\textbf{ID} & \textbf{Scale} & \textbf{Item} \\
\midrule
\endfirsthead

\multicolumn{3}{l}{\textit{Table continued from previous page}} \\[2pt]
\toprule
\textbf{ID} & \textbf{Scale} & \textbf{Item} \\
\midrule
\endhead

\midrule
\multicolumn{3}{r}{\textit{Continued on next page}} \\
\endfoot

\bottomrule
\endlastfoot

\multicolumn{3}{l}{\textbf{Module 0: Sanity Check (1--7 Likert)}} \\
Q1 & Authenticity & Does the conversation look real and authentic? (1 = Not at all authentic; 7 = Highly Authentic) \\
Q2 & Resemblance & How does the conversation in the study compare to a typical conversation you would have in your own counseling/therapy experience (1 = Does not resemble my experience at all; 7 = Resembles my experience a lot) \\
Q3 & Satisfaction & How satisfied are you with the evaluation results provided by \textsc{CounselReflect} for this conversation? (1 = Not satisfied at all; 7 = Very satisfied) \\
\midrule
\multicolumn{3}{l}{\textbf{Module 1: System Usability Scale(SUS) (1 (Strongly Disagree) - 5 (Strongly Agree)
)}} \\
Q4 & SUS & I think that I would like to use this AI tool frequently. \\
Q5 & SUS & I found this AI tool unnecessarily complex. \\
Q6 & SUS & I thought this AI tool was easy to use. \\
Q7 & SUS & I think that I would need the support of a technical person to be able to use this AI tool. \\
Q8 & SUS & I found the various functions in this AI tool were well integrated. \\
Q9 & SUS & I thought there was too much inconsistency in this AI tool. \\
Q10 & SUS & I would imagine that most people would learn to use this AI tool very quickly. \\
Q11 & SUS & I found this AI tool very cumbersome to use. \\
Q12 & SUS & I felt very confident using this AI tool. \\
Q13 & SUS & I needed to learn a lot of things before I could get going with this AI tool. \\
\midrule
\multicolumn{3}{l}{\textbf{Module 2: Trust in Automation (1 (Strongly Disagree) -  7 (Strongly Agree)
)}} \\
Q14 & TiA & This AI tool is deceptive. \\
Q15 & TiA & This AI tool behaves in an underhanded manner. \\
Q16 & TiA & I am suspicious of this AI tool’s intent, action, or outputs. \\
Q17 & TiA & I am wary of this AI tool. \\
Q18 & TiA & This AI tool's actions will have a harmful or injurious outcome. \\
Q19 & TiA & I am confident in this AI tool. \\
Q20 & TiA & This AI tool provides security. \\
Q21 & TiA & This AI tool has integrity. \\
Q22 & TiA & This AI tool is dependable. \\
Q23 & TiA & This AI tool is reliable. \\
Q24 & TiA & I can trust this AI tool. \\
Q25 & TiA & I am familiar with this AI tool. \\
\midrule
\multicolumn{3}{l}{\textbf{Module 3: Role Specific Questions (1 (Strongly Disagree) - 7 (Strongly Agree)
)}} \\
\multicolumn{3}{l}{\textit{Patients}} \\
Q26 & Patient & I feel that this AI tool is ``on my side'' and primarily motivated to help me reflect and improve the quality of counseling/therapy conversations. \\
Q27 & Patient & I felt comfortable and willing to be open and vulnerable when using this AI tool. \\
Q28 & Patient & I am confident that this AI tool will adhere to professional principles of confidentiality and privacy regarding my session data. \\
Q29 & Patient & I perceive this AI tool’s feedback as positive and helpful in understanding the quality of counseling conversations. \\

Q30 & Patient & I felt insecure, discouraged, or stressed because this AI tool was analyzing the conversation. \\
Q31 & Patient & A significant amount of mental activity (such as thinking about wording or deciding what to share) was required because I knew this AI tool was analyzing the conversation. \\
Q32 & Patient & Overall, this AI tool would be successful in helping my therapy session become more productive or insightful. \\
Q33 & Patient & The evaluation time before this AI tool provided feedback made the session review process feel slow and irritable. \\
Q34 & Patient & I found the setup and configuration settings for this AI tool very awkward to use. \\
Q35 & Patient & I found the information provided in this AI tool’s session reports or summaries to be unnecessarily complex. \\

\multicolumn{3}{l}{\textit{LLM Users}} \\
Q36 & LLM & Compared to generic LLMs like ChatGPT, this AI tool has the skills and competencies to offer better therapeutic insights. \\
Q37 & LLM & This AI tool is reliable in avoiding factual errors or ``hallucinations'' compared to standard LLM performance. \\
Q38 & LLM & I have a clear understanding of this AI tool’s underlying logic and why it provides specific dialogue analysis results. \\
Q39 & LLM & In general, I am willing to trust and rely on this AI tool . \\
Q40 & LLM & Compared to the effort of ``prompt engineering'' in a generic chatbot, I had to work hard to get a useful analysis from this AI tool. \\
Q41 & LLM & Significant mental and perceptual activity (e.g., thinking, deciding, searching) was required to verify the accuracy of this AI tool's dialogue analysis. \\
Q42 & LLM & I was successful in accomplishing my goal of obtaining high-quality therapeutic dialogue analysis with this AI tool. \\
Q43 & LLM & I felt insecure, discouraged, or stressed by clinical ``guardrails'' or system limitations when using this AI tool. \\
Q44 & LLM & Given my experience with LLMs, I believe most people with my background would learn to use this AI tool effectively and quickly. \\
Q45 & LLM & I felt there was too much inconsistency in the analysis this AI tool provided for the same or similar inputs. \\
Q46 & LLM & I feel that this AI tool is ``on my side'' and primarily motivated to help me reflect and improve the quality of counseling/therapy conversations. \\
Q47 & LLM & I felt comfortable and willing to be open and vulnerable when using this AI tool. \\
Q48 & LLM & I am confident that this AI tool will adhere to professional principles of confidentiality and privacy regarding my session data. \\
Q49 & LLM & I perceive this AI tool’s feedback as positive and helpful in understanding the quality of counseling conversations. \\
Q50 & LLM & I found the setup and configuration settings for this AI tool very awkward to use. \\
Q51 & LLM & I found the information provided in this AI tool’s session reports or summaries to be unnecessarily complex. \\
% \bottomrule
% \end{tabularx}}
% \end{table}
\end{longtable}

\subsection{Detailed Results}

\begin{figure}
    \centering
    \includegraphics[width=\linewidth]{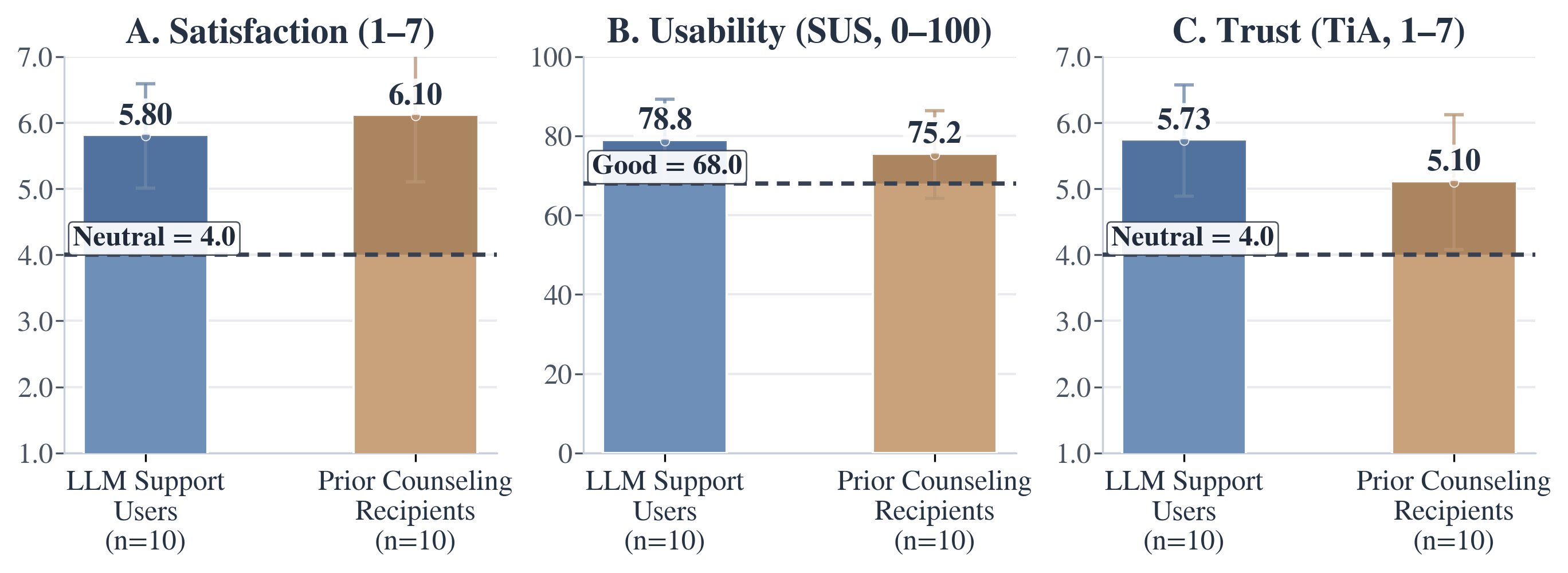}
    \caption{Post-study survey results across two participant groups for \textsc{CounselReflect} usability study. Dashed lines indicate reference thresholds. Bars show group means and error bars denote standard deviations.
    }
    \label{fig:human_eval_results}
\end{figure}

Figure~\ref{fig:human_eval_results} summarizes overall satisfaction (single-item), usability as measured by SUS (10 items) \cite{lewis_2018_sus}, and trust as measured by the TiA (12 items) scale \cite{jian_bisantz_drury_2000_tia}.  Table~\ref{tab:key_stats_overall} reports mean and standard deviation values for key metrics across LLM users ($n=10$) and prior counseling recipients ($n=10$) for survey questions. Table ~\ref{tab:role_specific_scores} presents the scores for rule-specific questions.

\section{Interview Study Details}

\subsection{Participants}
\label{appendix_participants}
Participants were recruited through mailing lists and Slack channels at two large U.S. universities. Eligibility required participants to (1) have experience using AI chatbots for mental health support, (2) be at least 18 years old and fluent in English, and (3) reside in the United States. We did not collect demographic information. Participants received a \$20 Amazon gift card.

\subsection{Participant-Selected Evaluation Metrics}

Table~\ref{tab:participant_metric_selection} reports the model-based and literature-derived/rubric-based metrics selected by each participant during the formal study. This participant-level breakdown complements the aggregate metric-selection results reported in the main text.

{
\fontsize{7.5}{8.5}\selectfont
\setlength{\tabcolsep}{2pt}
\renewcommand{\arraystretch}{0.88}

\begin{longtable}{
    >{\raggedright\arraybackslash}p{0.08\textwidth}
    >{\raggedright\arraybackslash}p{0.34\textwidth}
    >{\raggedright\arraybackslash}p{0.52\textwidth}
}
\caption{Evaluation metrics selected by participants during the interview sessions, grouped by metric category.}
\label{tab:participant_metric_selection} \\

\toprule
\textbf{Participant}
& \textbf{Model-based Metrics}
& \textbf{Literature-derived / Rubric-based Metrics} \\
\midrule
\endfirsthead

\multicolumn{3}{l}{\textit{Table~\ref{tab:participant_metric_selection} continued}}\\
\toprule
\textbf{Participant}
& \textbf{Model-based Metrics}
& \textbf{Literature-derived / Rubric-based Metrics} \\
\midrule
\endhead

\midrule
\multicolumn{3}{r}{\textit{Continued on next page}}\\
\endfoot

\bottomrule
\endlastfoot

P01 (total 7) & Emotion Classification; Emotional Triggers (RECCON); FactScore; MedScore &  Core Conditions (Therapeutic Alliance); Relationship Repair (Alliance Rupture Repair; Therapist Self-Disclosure) \\
P02 (total 34) & Emotion Classification, Emotional Triggers (RECCON), Empathy ER (Emotional Reaction), Empathy EX (Exploration), Empathy IP (Interpretation), Emotional Support Strategy, PAIR Reflection Scorer &
Communication Skills (Affirmation, Clarification, Non-Stigmatizing / Person-First Language, Normalizing, Open-Ended Questions, Patient-Centered/Person-Centered Language, Personalization to Client Context, Reflective Listening, Summarization, Therapeutic Concreteness); Core Condition (Affective Attunement, Cultural Humility, Emotional Support Provision, Empathic Responding, Empathy, Nonjudgmental Stance, Therapeutic Alliance, Therapist Genuineness, Therapist Validation, Unconditional Positive Regard); Emotional Processing (Emotion Focused Skills, Emotional Processing Facilitation, Grief Processing, Self-Compassion Promotion); Solution-Focused (Positive Reinforcement, Solution-Building Questions, Values Clarification) \\
P03 (total 22) & Toxicity (Detoxify ML), FactScore, MedScore, Emotional Support Strategy, PAIR Reflection Scorer  &  Advanced Skills (Dialectical Strategies, Here-and-Now Processing, Humor in Therapy, Narrative Reframing, Problem-Solving, Psychodynamic Interpertation, Relapse Prevention Planning); CBT Techniques (Chain Analysis, Cognitive Reframing, Sleep Hygiene Education, Social Skills Training); Communication Skills (Normalizing, Personalization to Client Context, Therapeutic Concreteness); MI Techniques (Strength Identification), Solution-Focused (Solution-Building Question, Values Clarification)  \\
P04 (total 16) & Emotional Support Strategy; Talk Type (Change/Neutral/Sustain); PAIR Reflection Scorer  & Communication Skills (Affirmation, Clarification, Non-Stigmatizing / Person-First Language, Normalizing, Open-Ended Questions, Patient-Centered/Person-Centered Language, Personalization to Client Context, Reflective Listening, Summarization, Therapeutic Concreteness);  Solution-Focused (Positive Reinforcement, Solution-Building Questions, Values Clarification)\\
P05 (total 9) & Emotion Classification, Emotional Triggers (RECCON), Emotional Support Strategy, PAIR Reflection Scorer, Talk Type (Change/Neutral/Sustain) & Emotion Processing (Emotion Focused Skills, Emotional Processing Facilitation, Grief Processing, Self-Compassion Promotion) \\
P06 (total 28) & Emotional Support Strategy, PAIR Reflection Scorer, FactScore, Toxicity (Detoxify ML)  & CBT Techniques (Assertiveness Training, Behavioral Activation, Chain Analysis, Cognitive Defusion, Cognitive Reframing, Homework Assignment, Identifying Core Beliefs, Psychoeducation, Sleep Hygiene Education, Social Skills Training, Socratic Questioning),  Core Conditions (Affective Attunement, Cultural Humility, Emotional Support Provision, Empathic Responding, Empathy, Nonjudgmental Stance, Therapeutic Alliance, Therapist Genuineness, Therapist Validation, Unconditional Positive Regard); Solution-Focused (Positive Reinforcement, Solution-Building Questions, Values Clarification) \\
P07 (total 4) & Talk Type (Change/Neutral/Sustain), FactScore, Emotional Support Strategy, PAIR Reflection Scorer  & - \\
P08 (total 9) & FactScore, Talk Type (Change/Neutral/Sustain), Empathy ER (Emotional Reaction), Empathy EX (Exploration), Empathy IP (Interpretation), Emotion Classification, Emotional Triggers (RECCON) &  Communication Skills (Clarification); Solution-Focused (Solution-Building Questions) \\
P09 (total 10) & Toxicity (Detoxify ML)  &  Emotion Processing (Emotion Focused Skills, Emotional Processing Facilitation, Grief Processing, Self-Compassion Promotion); Relationship Repair (Alliance Rupture Repair, Therapist Self-Disclosure)\& Mindfulness \& Body (Body-Oriented Interventions, Grounding Techniques, Mindfulness Induction) \\
P10 (15 metrics) & FactScore, MedScore, Emotional Support Strategy, PAIR Reflection Scorer, Toxicity (Detoxify ML) & Communication Skills (Clarification, Open-Ended Questions, Patient-Centered/Person-Centered Langauge, Personalization to Client Context, Reflective Listening, Summarization); Emotion Processing (Self-Compassion Promotion); Mindfulness \& Body (Grounding Techniques, Mindfulness Induction); Solution-Focused (Solution-Building Questiones) \\
P11 (total 6) & Talk Type (Change/Neutral/Sustain), Emotional Support Strategy, PAIR Reflection Scorer & Mindfulness \& Body (Body-Oriented Interventions, Grounding Techniques, Mindfulness Induction) \\
P12 (total 10) &  Talk Type (Change/Neutral/Sustain), FactScore, MedScore,  Emotion Classification; Emotional Triggers (RECCON) & CBT Techniques (Chain Analysis, Identifying Core Beliefs, Psychoeducation, Sleep Hygiene Education); Crisis \& Trauma (Trauma Processing) \\
P13 (total 15) & Emotion Classification, Emotional Triggers (RECCON), Empathy IP (Interpretation), FactScore,  Talk Type (Change/Neutral/Sustain)  & Communication Skills (Affirmation, Clarification, Non-Stigmatizing / Person-First Language, Normalizing, Open-Ended Questions, Patient-Centered/Person-Centered Language, Personalization to Client Context, Reflective Listening, Summarization, Therapeutic Concreteness) \\
P14 (total 37) &  Emotion Classification, Emotional Triggers (RECCON), Empathy ER (Emotional Reaction), Empathy EX (Exploration), Empathy IP (Interpretation), Emotional Support Strategy, PAIR Reflection Scorer, Talk Type (Change/Neutral/Sustain), Toxicity (Detoxify ML) & Communication Skills (Affirmation, Clarification, Non-Stigmatizing / Person-First Language, Normalizing, Open-Ended Questions, Patient-Centered/Person-Centered Language, Personalization to Client Context, Reflective Listening, Summarization, Therapeutic Concreteness); Core Conditions (Affective Attunement, Cultural Humility, Emotional Support Provision, Empathic Responding, Empathy, Nonjudgmental Stance, Therapeutic Alliance, Therapist Genuineness, Therapist Validation, Unconditional Positive Regard); Emotion Processing (Emotion Focused Skills, Emotional Processing Facilitation, Self-Compassion Promotion) ; Mindfulness \& Body (Grounding Techniques, Mindfulness Induction); Solution-Focused (Positive Reinforcement, Solution-Building Questions, Values Clarification)  \\
P15 (total 5) & Emotion Classification, Emotional Triggers (RECCON), Empathy ER (Emotional Reaction), Empathy EX (Exploration), Empathy IP (Interpretation)  &  N/A \\
P16 (total 5) & Emotion Classification, Emotional Triggers (RECCON), Empathy ER (Emotional Reaction), Empathy EX (Exploration), Empathy IP (Interpretation)  & N/A  \\
P17 (total 15) & FactScore, Empathy ER (Emotional Reaction), PAIR Reflection Scorer  &  Communication Skills (Affirmation, Clarification, Non-Stigmatizing / Person-First Language, Normalizing, Open-Ended Questions, Patient-Centered/Person-Centered Language, Personalization to Client Context, Reflective Listening, Summarization, Therapeutic Concreteness) ; Relationship Repair (Alliance Rupture Repair, Therapist Self-Disclosure)\\
P18 (total 14) &  Empathy ER (Emotional Reaction), Empathy IP (Interpretation), Emotional Support Strategy & CBT Techniques (Cognitive Reframing, Chain Analysis), Communication Skills (Clarification, Open-Ended Questions), Core Conditions (Affective Attunement, Emotional Support Provision, Empathic Responding, Empathy, Nonjudgmental Stance), Emotion Processing (Self-Compassion Promotion);  Mindfulness \& Body (Mindfulness Induction) \\
P19 (total 16) & Emotion Classification, Emotional Triggers (RECCON), Empathy ER (Emotional Reaction),  Empathy IP (Interpretation), FactScore, MedScore, Emotional Support Strategy, Talk Type (Change/Neutral/Sustain), Toxicity (Detoxify ML) & Core Conditions (Emotional Support Provision, Empathy, Nonjudgmental Stance, Therapeutic Alliance, Therapist Genuineness, Therapist Validation); Solution-Focused (Solution-Building Questions) \\
P20 (total 15) & Emotional Triggers (RECCON), Empathy ER (Emotional Reaction), Empathy EX (Exploration), Empathy IP (Interpretation) & Advanced Skills (Countertransference Management, Dialectical Strategies, Emotion Regulation Skills Training, Experiential Learning / Role-Playing, Externalization of Problem, Humor in Therapy, Instillation of Hope, Narrative Reframing, Problem-Solving, Psychodynamic Interpretation, Relapse Prevention Planning) \\
P21 (total 28) & Emotion Classification, Emotional Triggers (RECCON), Empathy ER (Emotional Reaction), Empathy EX (Exploration), Empathy IP (Interpretation), Toxicity (Detoxify ML), Emotional Support Strategy, PAIR Reflection Scorer   & Communication Skills (Affirmation, Non-Stigmatizing/Person-First Language, Normalizing, Patient-Centered/Person-Centered Language, Personalization to Client Context, Reflective Listening, Therapeutic Concreteness ), Core Conditions (Affective Attunement, Emotional Support Provision, Empathic Responding, Empathy, Nonjudgmental Stance, Therapeutic Alliance, Therapist Genuineness, Therapist Validation); Emotional Processing (Emotional Processing Facilitation, Emotion Focused Skills); Solution-Focused (Positive Reinforcement, Solution-Building Questions, Values Clarification) \\
% \bottomrule
% \end{tabular}
% \end{table*}
\end{longtable}
}}

\renewcommand{\labelitemi}{\textbullet}
\renewcommand{\labelitemii}{--}
\renewcommand{\labelitemiii}{$\circ$}
\renewcommand{\labelitemiv}{$\cdot$}

\subsection{Interview Protocol}
\label{sec:interview-protocol}

\subsubsection{Preliminary Questions}

\paragraph{Current use of AI for mental health and wellbeing support.}

\begin{itemize}
  \item Tell me about a time when you used AI chatbots for mental health and
        wellbeing support.
  \item What motivates you to seek support from AI?
  \item What kinds of things have you shared with the AI? What has the overall
        experience been like so far?
        \begin{itemize}
          \item Tell me about an experience when the conversation with AI went
                well.
                \begin{itemize}
                  \item What made the experience great?
                \end{itemize}
          \item Tell me about an experience when the conversation with AI went
                off.
                \begin{itemize}
                  \item What made the experience less satisfying?
                \end{itemize}
        \end{itemize}
  \item What are some of the reasons that drive you to continuously use AI for
        mental health and wellbeing support?
  \item Besides talking to an AI chatbot, how else do you seek mental health and
        wellbeing support?
\end{itemize}

\paragraph{Awareness to reflect on conversations with AI.}

\begin{itemize}
  \item Has an AI chatbot surprised you (in either a good or a bad way) in your
        previous conversations?
  \item What does a supportive (or unsupportive) conversation look like for you?
        \begin{itemize}
          \item What qualities does an AI chatbot hold to deliver such a
                conversation? (For instance, whether it seemed to understand how
                you felt or not, whether it felt genuinely warm or scripted.)
        \end{itemize}
  \item After talking to an AI chatbot about your mental health and wellbeing,
        have you ever taken the time to reflect on your conversations?
        \begin{itemize}
          \item If yes, tell us about your current approaches to reflection.
          \item Was there ever a time when you wondered, ``huh, why did the
                chatbot just say that?''
          \item Whether and how do you gauge the quality of the conversations?
          \item If you are unsure how to assess, tell me some questions that you
                have about AI chatbots and your conversations.
        \end{itemize}
\end{itemize}

\subsubsection{Cognitive Walkthrough Questions}

\emph{(Switch to the user testing session.)}

We will test out a tool called \textsc{CounselReflect}. It is designed to
help people reflect on mental health support provided by AI models. In our
session today, you will go through using this tool and share your feedback as a
user. It would be great if you can think aloud as you are using the tool.

\begin{itemize}
  \item Choose the model you would usually select for emotional support (or one
        you are interested in).
  \item Think of a recent difficult experience that you are comfortable sharing
        and see if the chatbot can provide you with emotional support.
\end{itemize}

\paragraph{Users' mental model when talking to AI about mental health and
wellbeing topics.}

\begin{itemize}
 \item \emph{Slowing down:}
        \begin{itemize}
          \item As you engage in conversation with the AI chatbot on this topic, has anything come to mind?
                \begin{itemize}
                  \item Any specific questions?
                  \item Any responses from the chatbot catch your attention?
                \end{itemize}
          \item What do you think about the quality of mental health support the
                chatbot is giving you? Why do you think this?
          \item How else would you engage in the conversation?
          \item Think about having this same conversation with a friend, family
                member, or therapist---would you carry out the conversation in a
                different way?
          \item What forms of support might your friend, family member, or
                therapist provide if you bring up a similar conversation to
                them?
                \begin{itemize}
                  \item How might the support be different from AI?
                  \item If so, do you have a preference among different forms of
                        support?
                \end{itemize}
        \end{itemize}
  \item \emph{Questions to encourage quiet participants:}
        \begin{itemize}
          \item What is going through your mind as you read this? Feel free to
                talk me through it.
          \item What is making you trust or doubt this response or suggestion?
                Is it the wording, the tone, or something else?
        \end{itemize}
\end{itemize}

\paragraph{Reflect on the conversation with AI.}

\begin{itemize}
  \item \emph{Slowing down:} Now, take a minute to reflect on the conversation
        that you just had with AI. Overall, how did the conversation go?
        \begin{itemize}
          \item Did anything stand out to you?
          \item Which part of the conversation was helpful? Less helpful?
                Confusing? Concerning?
          \item Was any part of the conversation unique from how you would talk
                about the same thing with a friend, family member, or therapist?
          \item How else might you engage in the conversation?
        \end{itemize}
\end{itemize}

%\paragraph{[Transition to using CounselReflect]}

\begin{itemize}
  \item Would you be interested in seeing how our AI tool evaluated this
        conversation?
    
  \item If you had to choose five metrics to check the quality of the AI
        support, how would you go about it? Feel free to add your own if you
        think something is missing.
        \begin{itemize}
          \item \emph{Follow up:} Among the metrics you chose, are there specific
                ones that are more important than others?
          \item \emph{Follow up:} Why did you choose those specific ones?
          \item \emph{Follow up:} If you had to design a metric yourself, what
                would it be?
        \end{itemize}
  \item Before reading your results, what questions are you hoping to get
        answered?
  \item Walk through your results, read the summary, and go through the
        infographic or data visualization.
        \begin{itemize}
          \item \emph{Follow up:} How well does the summary address the questions
                you had? (What did it cover, and what did it miss?)
          \item \emph{Follow up:} What format would be most helpful to see this
                information in?
        \end{itemize}
  \item Earlier you gave me your own view on the conversation. Now that you have
        seen the tool's evaluation, has your view changed?
        \begin{itemize}
          \item Is there anything the tool flagged that you had not noticed
                yourself?
          \item Is there anything you noticed that it missed?
        \end{itemize}
  \item After reflecting on the conversation and reviewing \textsc{CounselReflect}'s
        evaluation, would you engage with the chatbot differently?
\end{itemize}

\subsubsection{Debriefing and Final Thoughts}

\begin{itemize}
  \item All the evaluation approaches and metrics from \textsc{CounselReflect} are based
        on psychology literature and counseling best practices.
        \begin{itemize}
          \item Do these evaluations help you reflect on your conversations with
                the chatbot in a different way? Would following this guidance
                help you reflect on and improve the quality of support you
                receive from the chatbot?
          \item Alternatively, if you follow these ``textbook'' approaches to
                engage with the chatbot, would you miss out on some of the unique
                support that you would typically receive from a chatbot?
        \end{itemize}
  \item Any final takeaways or moments that stood out to you most?
\end{itemize}

%% file: ref.bib
@String{Computer = "{IEEE} Computer" }

@String{Academic = "Academic Press" }

@String{Springer = "Springer-Verlag" }

@misc{li2024crowdsourceddatahighqualitybenchmarks,
      title={From Crowdsourced Data to High-Quality Benchmarks: Arena-Hard and BenchBuilder Pipeline}, 
      author={Tianle Li and Wei-Lin Chiang and Evan Frick and Lisa Dunlap and Tianhao Wu and Banghua Zhu and Joseph E. Gonzalez and Ion Stoica},
      year={2024},
      eprint={2406.11939},
      archivePrefix={arXiv},
      primaryClass={cs.LG},
      url={https://arxiv.org/abs/2406.11939}, 
}

@misc{gu2025surveyllmasajudge,
      title={A Survey on LLM-as-a-Judge}, 
      author={Jiawei Gu and Xuhui Jiang and Zhichao Shi and Hexiang Tan and Xuehao Zhai and Chengjin Xu and Wei Li and Yinghan Shen and Shengjie Ma and Honghao Liu and Saizhuo Wang and Kun Zhang and Yuanzhuo Wang and Wen Gao and Lionel Ni and Jian Guo},
      year={2025},
      eprint={2411.15594},
      archivePrefix={arXiv},
      primaryClass={cs.CL},
      url={https://arxiv.org/abs/2411.15594}, 
}

@misc{liu2023gevalnlgevaluationusing,
      title={G-Eval: NLG Evaluation using GPT-4 with Better Human Alignment}, 
      author={Yang Liu and Dan Iter and Yichong Xu and Shuohang Wang and Ruochen Xu and Chenguang Zhu},
      year={2023},
      eprint={2303.16634},
      archivePrefix={arXiv},
      primaryClass={cs.CL},
      url={https://arxiv.org/abs/2303.16634}, 
}

@misc{liu2019robertarobustlyoptimizedbert,
      title={RoBERTa: A Robustly Optimized BERT Pretraining Approach}, 
      author={Yinhan Liu and Myle Ott and Naman Goyal and Jingfei Du and Mandar Joshi and Danqi Chen and Omer Levy and Mike Lewis and Luke Zettlemoyer and Veselin Stoyanov},
      year={2019},
      eprint={1907.11692},
      archivePrefix={arXiv},
      primaryClass={cs.CL},
      url={https://arxiv.org/abs/1907.11692}, 
}

@article{civil_comments_dataset,
  author    = {Daniel Borkan and
               Lucas Dixon and
               Jeffrey Sorensen and
               Nithum Thain and
               Lucy Vasserman},
  title     = {Nuanced Metrics for Measuring Unintended Bias with Real Data for Text
               Classification},
  journal   = {CoRR},
  volume    = {abs/1903.04561},
  year      = {2019},
  url       = {http://arxiv.org/abs/1903.04561},
  archivePrefix = {arXiv},
  eprint    = {1903.04561},
  bibsource = {dblp computer science bibliography, https://dblp.org}
}

@inproceedings{barbieri_camacho_collados_espinosa_anke_neves_2020_tweeteval,
    title = "{T}weet{E}val: Unified Benchmark and Comparative Evaluation for Tweet Classification",
    author = "Barbieri, Francesco  and
      Camacho-Collados, Jose  and
      Espinosa Anke, Luis  and
      Neves, Leonardo",
    editor = "Cohn, Trevor  and
      He, Yulan  and
      Liu, Yang",
    booktitle = "Findings of the Association for Computational Linguistics: EMNLP 2020",
    month = nov,
    year = "2020",
    address = "Online",
    publisher = "Association for Computational Linguistics",
    url = "https://aclanthology.org/2020.findings-emnlp.148/",
    doi = "10.18653/v1/2020.findings-emnlp.148",
    pages = "1644--1650"
}

@INPROCEEDINGS{anno_mi_dataset,
  author={Wu, Zixiu and Balloccu, Simone and Kumar, Vivek and Helaoui, Rim and Reiter, Ehud and Reforgiato Recupero, Diego and Riboni, Daniele},
  booktitle={ICASSP 2022 - 2022 IEEE International Conference on Acoustics, Speech and Signal Processing (ICASSP)}, 
  title={Anno-MI: A Dataset of Expert-Annotated Counselling Dialogues}, 
  year={2022},
  volume={},
  number={},
  pages={6177-6181},
  doi={10.1109/ICASSP43922.2022.9746035}}

@misc{li2017dailydialogmanuallylabelledmultiturn,
      title={DailyDialog: A Manually Labelled Multi-turn Dialogue Dataset}, 
      author={Yanran Li and Hui Su and Xiaoyu Shen and Wenjie Li and Ziqiang Cao and Shuzi Niu},
      year={2017},
      eprint={1710.03957},
      archivePrefix={arXiv},
      primaryClass={cs.CL},
      url={https://arxiv.org/abs/1710.03957}, 
}

@misc{liu2021emotionalsupportdialogsystems,
      title={Towards Emotional Support Dialog Systems}, 
      author={Siyang Liu and Chujie Zheng and Orianna Demasi and Sahand Sabour and Yu Li and Zhou Yu and Yong Jiang and Minlie Huang},
      year={2021},
      eprint={2106.01144},
      archivePrefix={arXiv},
      primaryClass={cs.CL},
      url={https://arxiv.org/abs/2106.01144}, 
}

@article{hyzy_bond_mulvenna_bai_dix_leigh_hunt_2022_sus_for_digital_health_apps, title={System Usability Scale Benchmarking for Digital Health Apps: Meta-analysis}, volume={10}, DOI={https://doi.org/10.2196/37290}, number={8}, journal={JMIR mHealth and uHealth}, author={Hyzy, Maciej and Bond, Raymond and Mulvenna, Maurice and Bai, Lu and Dix, Alan and Leigh, Simon and Hunt, Sophie}, year={2022}, month={Aug} }

@article{empathy_1, title={Evaluating Conversational Agents for Mental Health: Scoping Review of Outcomes and Outcome Measurement Instruments}, volume={25}, DOI={https://doi.org/10.2196/44548}, journal={Journal of Medical Internet Research}, author = {Jabir, Ahmad Ishqi and
          Martinengo, Laura and
          Lin, Xiaowen and
          Torous, John and
          Subramaniam, Mythily and
          Tudor Car, Lorainne}, year={2023}, month={Apr} }

@article{empathy_2, title={Empathy},  url={https://www.ncbi.nlm.nih.gov/pmc/articles/PMC8795635/pdf/aps-2021-02271.pdf}, DOI={https://doi.org/10.5999/aps.2021.02271}, journal={Archives of Plastic Surgery}, author={Zhang, Wuyang and Hallock, Geoffrey G.}, year = {2022},
volume = {49},
number = {1},
pages = {3--4},}

@article{empathy_3, title={Foundation metrics for evaluating effectiveness of healthcare conversations powered by generative AI}, volume={7}, url={https://www.nature.com/articles/s41746-024-01074-z}, DOI={https://doi.org/10.1038/s41746-024-01074-z}, number={1}, journal={npj Digital Medicine}, author={Abbasian, Mahyar and Khatibi, Elahe and Azimi, Iman and Oniani, David and Shakeri Hossein Abad, Zahra and Thieme, Alexander and Sriram, Ram and Yang, Zhongqi and Wang, Yanshan and Lin, Bryant and Gevaert, Olivier and Li, Li-Jia and Jain, Ramesh and Rahmani, Amir M.}, year={2024}, month={Mar}, pages={1–14} }

@article{empathy_4_empathic_responding_4, title={Empathy in the Clinician–Patient Relationship}, volume={4}, url={https://pmc.ncbi.nlm.nih.gov/articles/PMC5513642/}, DOI={https://doi.org/10.1177/2374373517699271}, number={2}, journal={Journal of Patient Experience}, author={Finset, Arnstein and Ørnes, Knut}, year={2017}, pages={64–68} }

@article{empathy_5, title={Empathy in patient care: From “clinical empathy” to “empathic concern”}, volume={24}, DOI={https://doi.org/10.1007/s11019-021-10033-4}, number={4}, journal={Medicine, Health Care and Philosophy}, author={Guidi, Clarissa and Traversa, Chiara}, year={2021}, month={Jul}, pages={573–585} }

@article{therapeutic_alliance_1, title={The Therapy Process Questionnaire ‐ Factor analysis and psychometric properties of a multidimensional self‐rating scale for high‐frequency monitoring of psychotherapeutic processes}, volume={26}, DOI={https://doi.org/10.1002/cpp.2384}, number={5}, journal={Clinical Psychology \& Psychotherapy}, author={Schiepek, Günter and Stöger‐Schmidinger, Barbara and Kronberger, Helmut and Aichhorn, Wolfgang and Kratzer, Leonhard and Heinz, Peter and Viol, Kathrin and Lichtwarck‐Aschoff, Anna and Schöller, Helmut}, year={2019}, month={Jul}, pages={586–602} }

@article{therapeutic_alliance_2_open_ended_questions_1_engagement_participation_facilitation_5, title={The therapeutic alliance: The fundamental element of psychotherapy}, volume={16}, url={https://pmc.ncbi.nlm.nih.gov/articles/PMC6493237/}, DOI={https://doi.org/10.1176/appi.focus.20180022}, number={4}, journal={FOCUS}, author={Stubbe, Dorothy E.}, year={2018}, pages={402–403} }

@article{therapeutic_alliance_3, title={Therapeutic Alliance and Outcome of Psychotherapy: Historical Excursus, Measurements, and Prospects for Research}, volume={2}, DOI={https://doi.org/10.3389/fpsyg.2011.00270}, number={270}, journal={Frontiers in Psychology}, author={Ardito, Rita B. and Rabellino, Daniela}, year={2011}, month={Oct} }

@article{therapeutic_alliance_4, title={Harnessing the potential of the therapeutic alliance}, volume={13}, url={https://www.ncbi.nlm.nih.gov/pmc/articles/PMC4219056/}, DOI={https://doi.org/10.1002/wps.20147}, number={3}, journal={World Psychiatry}, author={Arnow, Bruce A. and Steidtmann, Dana}, year={2014}, month={Oct}, pages={238–240} }

@article{therapeutic_alliance_5, title={Therapeutic Alliance: A Concept for the Childbearing Season}, volume={18}, DOI={https://doi.org/10.1624/105812409x461216}, number={3}, journal={Journal of Perinatal Education}, author={Doherty, Mary Ellen}, year={2009}, month={Jan}, pages={39–47} }

@article{therapist_genuineness_2, title={Do clients train therapists to become eclectic and use the common factors? A qualitative study listening to experienced psychotherapists}, volume={10}, DOI={https://doi.org/10.1186/s40359-022-00886-6}, number={1}, journal={BMC Psychology}, author={Behan, Douglas}, year={2022}, month={Jul} }

@article{therapist_genuineness_3, title={Some personal reflections relating to psychotherapy}, volume={50}, DOI={https://doi.org/10.4103/0019-5545.44756}, number={4}, journal={Indian Journal of Psychiatry}, author={Shamasundar, C}, year={2008}, pages={301-304} }

@article{therapist_genuineness_4, title={The Importance of Genuineness in Public Engagement—An Exploratory Study of Pediatric Communication on Social Media in China}, volume={17}, DOI={https://doi.org/10.3390/ijerph17197078}, number={19}, journal={International Journal of Environmental Research and Public Health}, author={Lu, Wenze and Ngai, Cindy Sing Bik and Yang, Lu}, year={2020}, month={Sep}, pages={7078} }

@article{therapist_genuineness_5, title={Psychotherapy as investigation: cultivating curiosity and insight in the therapeutic process}, volume={16}, DOI={https://doi.org/10.3389/fpsyg.2025.1603719}, journal={Frontiers in Psychology}, publisher={Frontiers Media}, author={Brewer, Judson A and Fabio Giommi}, year={2025}, month={Jul} }

@article{unconditional_positive_regard_1, title={R-E-S-P-E-C-T—What it Means to Patients}, volume={29}, DOI={https://doi.org/10.1007/s11606-013-2710-z}, number={3}, journal={Journal of General Internal Medicine}, author={Frosch, Dominick L. and Tai-Seale, Ming}, year={2014}, month={Jan}, pages={427–428} }

@article{unconditional_positive_regard_2, title={Effects of Mindfulness on Psychological health: a Review of Empirical Studies}, volume={31}, url={https://pmc.ncbi.nlm.nih.gov/articles/PMC3679190/}, DOI={https://doi.org/10.1016/j.cpr.2011.04.006}, number={6}, journal={Clinical Psychology Review}, author={Keng, Shian Ling and Smoski, Moria J. and Robins, Clive J.}, year={2011}, pages={1041–1056} }

@article{unconditional_positive_regard_3, title={Revisiting the organismic valuing process theory of personal growth: A theoretical review of Rogers and its connection to positive psychology}, volume={11}, url={https://pmc.ncbi.nlm.nih.gov/articles/PMC7385226/}, DOI={https://doi.org/10.3389/fpsyg.2020.01706}, number={1706}, journal={Frontiers in Psychology}, author={Maurer, Mia M. and Daukantaitė, Daiva}, year={2020}, month={Jul} }

@article{unconditional_positive_regard_4, title={The Impact of Parental Styles on the Development of Psychological Complaints}, volume={11}, DOI={https://doi.org/10.5964/ejop.v11i1.836}, number={1}, journal={Europe’s Journal of Psychology}, author={Rocha Lopes, Daniela and van Putten, Kees and Moormann, Peter Paul}, year={2015}, month={Feb}, pages={155–168} }

@article{unconditional_positive_regard_5, title={Show Me the Real You: Enhanced Expression of Rogerian Conditions in Therapeutic Relationship Building with Autistic Adults}, volume={4}, DOI={https://doi.org/10.1089/aut.2021.0065}, number={2}, journal={Autism in Adulthood}, author={Hume, Romy}, year={2022}, month={Feb} }

@article{therapist_validation_1_therapeutic_concreteness_5, title={Are there interactional differences between telephone and face-to-face psychological therapy? A systematic review of comparative studies}, volume={265}, DOI={https://doi.org/10.1016/j.jad.2020.01.057},  journal={Journal of Affective Disorders}, author={Irvine, Annie and Drew, Paul and Bower, Peter and Brooks, Helen and Gellatly, Judith and Armitage, Christopher J. and Barkham, Michael and McMillan, Dean and Bee, Penny}, year={2020}, month={Mar}, pages={120–131} }

@article{therapist_validation_2, title={The who and what of validation: an experimental examination of validation and invalidation of specific emotions and the moderating effect of emotion dysregulation}, volume={9}, url={https://pmc.ncbi.nlm.nih.gov/articles/PMC9116024/}, DOI={https://doi.org/10.1186/s40479-022-00185-x}, number={1}, journal={Borderline Personality Disorder and Emotion Dysregulation}, author={Kuo, Janice R. and Fitzpatrick, Skye and Ip, Jennifer and Uliaszek, Amanda}, year={2022}, month={May}, pages={15} }

@article{therapist_validation_3, title={Development of a Framework for the Implementation of Synchronous Digital Mental Health: Realist Synthesis of Systematic Reviews}, volume={9}, DOI={https://doi.org/10.2196/34760}, number={3}, journal={JMIR Mental Health}, author={Villarreal-Zegarra, David and Alarcon-Ruiz, Christoper A and Melendez-Torres, GJ and Torres-Puente, Roberto and Navarro-Flores, Alba and Cavero, Victoria and Ambrosio-Melgarejo, Juan and Rojas-Vargas, Jefferson and Almeida, Guillermo and Albitres-Flores, Leonardo and Romero-Cabrera, Alejandra B and Huarcaya-Victoria, Jeff}, year={2022}, month={Mar}, pages={e34760} }

@article{therapist_validation_4, title={Defining pain-validation: the Importance of Validation in Reducing the Stresses of Chronic Pain}, volume={3}, url={https://www.ncbi.nlm.nih.gov/pmc/articles/PMC9614309/}, DOI={https://doi.org/10.3389/fpain.2022.884335}, journal={Frontiers in Pain Research}, author={Nicola, Melinda and Correia, Helen and Ditchburn, Graeme and Drummond, Peter D.}, year={2022}, month={Oct}, pages={884335} }

@article{therapist_validation_5, title={Continuing the Conversation for a Clinical Conceptualization of Attachment Trauma.}, volume={22}, DOI={https://doi.org/10.36131/cnfioritieditore20250516}, number={5}, journal={Clinical Neuropsychiatry}, publisher={National Institutes of Health}, author={Schimmenti, Adriano and Farina, Benedetto}, year={2025}, month={Oct}, pages={428–436} }

@article{reflective_listening_1, title={Motivational Interviewing Is “Doing” What Matters: Integrating Motivational Interviewing Spirit and Skills into What Matters to You? Conversations}, volume={12}, DOI={https://doi.org/10.1177/23743735251317041}, journal={Journal of Patient Experience}, publisher={SAGE Publishing}, author={Gutnick, Damara and McNeilly, Sarah}, year={2025}, month={Jan} }

@article{reflective_listening_2, title={The Power of Active Listening to Address Medication Non-Adherence During Care Transition: A Case Report of a Polypharmacy Patient with Type 2 Diabetes}, volume={13}, url={https://www.mdpi.com/2226-4787/13/3/64}, DOI={https://doi.org/10.3390/pharmacy13030064}, number={3}, journal={Pharmacy}, publisher={Multidisciplinary Digital Publishing Institute}, author={Dost, Solh and Gastaldi, Giacomo and Schneider, Marie P}, year={2025}, month={Jun}, pages={64} }

@article{reflective_listening_3, title={Listening visits: an evaluation of the effectiveness and acceptability of a home-based depression treatment}, volume={20}, url={https://pubmed.ncbi.nlm.nih.gov/21154029/}, DOI={https://doi.org/10.1080/10503307.2010.518636}, number={6}, journal={Psychotherapy Research: Journal of the Society for Psychotherapy Research}, author={Segre, Lisa S. and Stasik, Sara M. and O’Hara, Michael W. and Arndt, Stephan}, year={2010}, month={Nov}, pages={712–721} }

@article{reflective_listening_4, title={Enhancing listening skills among future public health professionals: a pre-post educational intervention study}, volume={13}, DOI={https://doi.org/10.3389/fpubh.2025.1637788}, journal={Frontiers in Public Health}, publisher={Frontiers Media SA}, author={Kercher, Kyle A. and Heeter, Kathleen N. and Dannelley, Claire and King, Julianna F. and Schaefer, Samantha and Martinez Kercher, Vanessa M.}, year={2025}, month={Nov} }

@article{reflective_listening_5, title={More Than Reflections: Empathy in Motivational Interviewing Includes Language Style Synchrony Between Therapist and Client}, volume={46}, url={https://www.ncbi.nlm.nih.gov/pmc/articles/PMC5018199/}, DOI={https://doi.org/10.1016/j.beth.2014.11.002}, number={3}, journal={Behavior Therapy}, author={Lord, Sarah Peregrine and Sheng, Elisa and Imel, Zac E. and Baer, John and Atkins, David C.}, year={2015}, month={May}, pages={296–303} }

@article{socratic_questioning_1_open_ended_questions_5, title={Therapist Use of Socratic Questioning Predicts session-to-session Symptom Change in Cognitive Therapy for Depression}, volume={70}, url={https://pmc.ncbi.nlm.nih.gov/articles/PMC4449800/}, DOI={https://doi.org/10.1016/j.brat.2015.05.004}, number={1}, journal={Behaviour Research and Therapy}, author={Braun, Justin D. and Strunk, Daniel R. and Sasso, Katherine E. and Cooper, Andrew A.}, year={2015}, month={Jul}, pages={32–37} }

@article{socratic_questioning_2, title={The Use of the Socratic Inquiry to Facilitate Critical Thinking in Nursing Education}, volume={24}, url={https://pmc.ncbi.nlm.nih.gov/articles/PMC6917464/}, DOI={https://doi.org/10.4102/hsag.v24i0.1224}, number={1224}, journal={Health SA Gesondheid}, author={Makhene, Agnes}, year={2019}, month={Apr} }

@article{socratic_questioning_3, title={Using GenAI for Socratic Questioning: An Approach to Higher‐Order Thinking for Nursing Education}, volume={12}, DOI={https://doi.org/10.1002/nop2.70355}, number={10}, journal={Nursing Open}, publisher={Wiley}, author={Chan, Jackie and Ho, Ken}, year={2025}, month={Oct}, pages={e70355–e70355} }

@article{socratic_questioning_4, title={The fact of ignorance revisiting the socratic method as a tool for teaching critical thinking}, volume={78}, DOI={https://doi.org/10.5688/ajpe787144}, number={7}, journal={American Journal of Pharmaceutical Education}, author={Oyler, Douglas R. and Romanelli, Frank}, year={2014}, month={Sep}, pages={144} }

@article{socratic_questioning_5, title={Treating Guilt-Inducing Self-Talk in Ocd with Dramatized Socratic Dialogue: A Step by Step Intervention.}, volume={21}, url={https://www.ncbi.nlm.nih.gov/pmc/articles/PMC10979789}, DOI={https://doi.org/10.36131/cnfioritieditore2023060104}, number={1}, journal={Clinical Neuropsychiatry}, publisher={National Institutes of Health}, author={Angelo Maria Saliani and Perdighe, Claudia and Zaccari, Vittoria and Olga Ines Luppino and Mancini, Alessandra and Tenore, Katia and Mancini, Francesco}, year={2024}, month={Feb}, pages={63–78} }

@article{alliance_rupture_repair_1, title={Interpersonal Problems and Their Impact on Alliance Ruptures in Psychotherapy}, volume={32}, DOI={https://doi.org/10.1002/cpp.70071}, number={2}, journal={Clinical Psychology \& Psychotherapy}, author={Iovoli, Flavio and Baranowski, Marie and Sander, Leonie S. and Rubel, Julian A.}, year={2025}, month={Mar} }

@article{alliance_rupture_repair_2, title={Clinical consensus strategies to repair ruptures in the therapeutic alliance.}, volume={28}, DOI={https://doi.org/10.1037/int0000097}, number={1}, journal={Journal of Psychotherapy Integration}, author={Eubanks, Catherine F. and Burckell, Lisa A. and Goldfried, Marvin R.}, year={2018}, month={Mar}, pages={60–76} }

@article{alliance_rupture_repair_3, title={Detecting alliance ruptures: the effects of the therapist’s experience, attachment, empathy and countertransference management skills}, volume={22}, DOI={https://doi.org/10.4081/ripppo.2019.325}, number={1}, journal={Research in Psychotherapy: Psychopathology, Process and Outcome}, author={Talbot, Corinne and Ostiguy-Pion, Rose and Painchaud, Esther and Lafrance, Claudelle and Descôteaux, Jean}, year={2019}, month={Apr} }

@article{alliance_rupture_repair_4, title={Patterns of therapeutic alliance: Rupture–repair episodes in prolonged exposure for posttraumatic stress disorder.}, volume={82}, DOI={https://doi.org/10.1037/a0034696}, number={1}, journal={Journal of Consulting and Clinical Psychology}, author={McLaughlin, AnnaMaria Aguirre and Keller, Stephanie M. and Feeny, Norah C. and Youngstrom, Eric A. and Zoellner, Lori A.}, year={2014}, pages={112–121} }

@article{alliance_rupture_repair_5, title={Thinking transtheoretically about alliance and rupture: Implications for practice and training}, volume={6}, DOI={https://doi.org/10.32872/cpe.12439}, number={Special Issue}, journal={Clinical psychology in Europe}, author={Sigal Zilcha-Mano and J. Christopher Muran}, year={2024}, month={Apr} }

@article{therapist_self_disclosure_1, title={A Phenomenological Investigation into the Use of Therapist Self-disclosure in Compassion-Focused Therapy With Forensic Clients}, DOI={https://doi.org/10.1177/0306624x241227409}, journal={International Journal of Offender Therapy and Comparative Criminology}, publisher={SAGE Publishing}, author={Rachwal, Francesca and Gredecki, Neil}, year={2024}, month={Feb} }

@article{therapist_self_disclosure_2, title={Differentiating self-disclosure interventions from self-involving interventions based on the assessment of the short-term therapeutic effects: preliminary results}, volume={28}, DOI={https://doi.org/10.4081/ripppo.2025.800}, number={1}, journal={Research in Psychotherapy Psychopathology Process and Outcome}, publisher={PAGEPress (Italy)}, author={Monticelli, Fabio and Chiara Massullo and Carcione, Antonino and Tombolini, Lucia and Guerra, Flaminia and Liotti, Marianna and Monticelli, Cecilia and Gasperini, Elena and Russo, Marianna and Samanta Novaretto and Vista, Letizia La and Mallozzi, Paola and Imperatori, Claudio and Brutto, Chiara Del and Farina, Benedetto}, year={2025}, month={Feb} }

@article{therapist_self_disclosure_3, title={Exploring the Use of the Therapist’s Self in Therapy: a Systematic Review}, volume={47}, DOI={https://doi.org/10.1177/02537176241252363}, number={1}, journal={Indian Journal of Psychological Medicine}, publisher={SAGE Publishing}, author={Alva, Meera H and Antony, Sherin P and Kataria, Kanak}, year={2024}, month={May} }

@article{therapist_self_disclosure_4, title={Let’s Dare to Be Vulnerable: Crossing the Self-Disclosure Rubicon}, volume={23}, url={https://pmc.ncbi.nlm.nih.gov/articles/PMC11936366/}, DOI={https://doi.org/10.1370/afm.240310}, number={2}, journal={The Annals of Family Medicine}, publisher={American Academy of Family Physicians}, author={Ohad Avny and Alon, Aya}, year={2025}, month={Mar}, pages={170–172} }

@article{therapist_self_disclosure_5, title={Projective Identification, Self-Disclosure, and the Patient’s View of the Object: The Need for Flexibility}, volume={8}, url={https://pmc.ncbi.nlm.nih.gov/articles/PMC3330553/}, number={3}, journal={The Journal of Psychotherapy Practice and Research}, author={Waska, Robert T}, year={1999}, pages={225-233} }

@article{goal_consensus_1, title={Goal consensus and task agreement as predictors of attendance and compliance in community-based treatments for adolescents with emotional disorders}, volume={19}, DOI={https://doi.org/10.1186/s13034-025-00970-w}, number={1}, journal={Child and Adolescent Psychiatry and Mental Health}, publisher={BioMed Central}, author={Panek, Adam and Butler, Emilie J and Jensen-Doss, Amanda and Ehrenreich-May, Jill and Ginsburg, Golda S}, year={2025}, month={Nov}, pages={130–130} }

@article{goal_consensus_2, title={Goal planning in mental health service delivery: A systematic integrative review}, volume={13}, url={https://www.ncbi.nlm.nih.gov/pmc/articles/PMC9807176/}, DOI={https://doi.org/10.3389/fpsyt.2022.1057915}, journal={Frontiers in Psychiatry}, author={Stewart, Victoria and McMillan, Sara S. and Hu, Jie and Ng, Ricki and El-Den, Sarira and O’Reilly, Claire and Wheeler, Amanda J.}, year={2022}, month={Dec} }

@article{goal_consensus_3, title={Patients’ perceived lack of goal clarity in psychological treatments: Scale development and negative correlates}, volume={27}, DOI={https://doi.org/10.1002/cpp.2479}, number={6}, journal={Clinical Psychology \& Psychotherapy}, author={Geurtzen, Naline and Keijsers, Ger P.J. and Karremans, Johan C. and Tiemens, Bea G. and Hutschemaekers, Giel J.M.}, year={2020}, month={Jun}, pages={915–924} }

@article{goal_consensus_5, title={Goal‐oriented practices in youth mental health and wellbeing settings: A scoping review and thematic analysis of empirical evidence}, volume={98}, DOI={https://doi.org/10.1111/papt.12564}, number={2}, journal={Psychology and Psychotherapy Theory Research and Practice}, publisher={Wiley}, author={Jacob, Jenna and Wozney, Lori and Hanne Weie Oddli and Duncan, Charlie and Chorney, Jill and Emberly, Debbie and Law, Duncan and Clark, Sharon and Heien, Sofie and Boulos, Leah and Cooper, Mick}, year={2024}, month={Dec} }

@article{open_ended_questions_2, title={Opening Up: Clients’ Inner Struggles in the Initial Phase of Therapy}, volume={11}, url={https://www.ncbi.nlm.nih.gov/pmc/articles/PMC7769763/}, DOI={https://doi.org/10.3389/fpsyg.2020.591146}, number={591146}, journal={Frontiers in Psychology}, author={Kleiven, Gøril Solberg and Hjeltnes, Aslak and Råbu, Marit and Moltu, Christian}, year={2020}, month={Dec} }

@article{open_ended_questions_3, title={Questions and Reflections: The Use of Motivational Interviewing Microskills in a Peer-Led Brief Alcohol Intervention for College Students}, volume={39}, DOI={https://doi.org/10.1016/j.beth.2007.07.001}, number={2}, journal={Behavior Therapy}, author={Tollison, Sean J. and Lee, Christine M. and Neighbors, Clayton and Neil, Teryl A. and Olson, Nichole D. and Larimer, Mary E.}, year={2008}, month={Jun}, pages={183–194} }

@article{open_ended_questions_4, title={Motivational interviewing to improve treatment engagement and outcome in individuals seeking treatment for substance abuse: A multisite effectiveness study}, volume={81}, url={https://www.ncbi.nlm.nih.gov/pmc/articles/PMC2386852/}, DOI={https://doi.org/10.1016/j.drugalcdep.2005.08.002}, number={3}, journal={Drug and Alcohol Dependence}, author={Carroll, Kathleen M. and Ball, Samuel A. and Nich, Charla and Martino, Steve and Frankforter, Tami L. and Farentinos, Christiane and Kunkel, Lynn E. and Mikulich-Gilbertson, Susan K. and Morgenstern, Jon and Obert, Jeanne L. and Polcin, Doug and Snead, Ned and Woody, George E.}, year={2006}, month={Feb}, pages={301–312} }

@article{affirmation_1_strength_identification_5, title={The effects of group counseling and self-affirmation on stigma and group relationship development: A replication and extension.}, volume={69}, DOI={https://doi.org/10.1037/cou0000614}, number={5}, journal={Journal of Counseling Psychology}, author={Seidman, Andrew J. and Wade, Nathaniel G. and Geller, Jason}, year={2022}, month={Mar} }

@article{affirmation_2_normalizing_1, title={“Good job!”: Therapists’ encouragement, affirmation, and personal address in internet-based cognitive behavior therapy for adolescents with depression}, volume={30}, url={https://www.sciencedirect.com/science/article/pii/S2214782922000999}, DOI={https://doi.org/10.1016/j.invent.2022.100592}, number={1}, journal={Internet Interventions}, author={Berg, Ida and Hovne, Vera and Carlbring, Per and Bernhard-Oettel, Claudia and Oscarsson, Martin and Mechler, Jakob and Lindqvist, Karin and Topooco, Naira and Andersson, Gerhard and Philips, Björn}, year={2022}, month={Dec}, pages={100592} }

@article{affirmation_3, title={Enhance psychotherapy outcomes by encouraging patients to regularly self-monitor, reflect on, and share their affective responses toward their therapist: Protocol for a randomized controlled trial}, volume={13}, DOI={https://doi.org/10.2196/55369}, journal={JMIR Research Protocols}, publisher={JMIR Publications}, author={Stefana, Alberto and Vieta, Eduard and Paolo Fusar-Poli and Youngstrom, Eric}, year={2023}, month={Dec} }

@article{affirmation_4, title={Self-affirmation activates brain systems associated with self-related processing and reward and is reinforced by future orientation}, volume={11}, url={https://pmc.ncbi.nlm.nih.gov/articles/PMC4814782/}, DOI={https://doi.org/10.1093/scan/nsv136}, number={4}, journal={Social Cognitive and Affective Neuroscience}, author={Cascio, Christopher N. and O’Donnell, Matthew Brook and Tinney, Francis J. and Lieberman, Matthew D. and Taylor, Shelley E. and Strecher, Victor J. and Falk, Emily B.}, year={2015}, month={Nov}, pages={621–629} }

@article{affirmation_5_autonomy_support_2_change_talk_1, title={Measuring Motivation: Change Talk and Counter-Change Talk in Cognitive Behavioral Therapy for Generalized Anxiety}, volume={43}, url={https://www.ncbi.nlm.nih.gov/pmc/articles/PMC3863762/}, DOI={https://doi.org/10.1080/16506073.2013.846400}, number={1}, journal={Cognitive Behaviour Therapy}, author={Lombardi, Diana R. and Button, Melissa L. and Westra, Henny A.}, year={2014}, month={Oct}, pages={12–21} }

@article{summarization_1, title={A Comparison Between Clinical Guidelines and Real-World Treatment Data in Examining the Use of Session Summaries: Retrospective Study}, volume={6}, url={https://formative.jmir.org/2022/8/e39846/pdf}, DOI={https://doi.org/10.2196/39846}, number={8}, journal={JMIR Formative Research}, author={Sadeh-Sharvit, Shiri and Rego, Simon A and Jefroykin, Samuel and Peretz, Gal and Kupershmidt, Tomer}, year={2022}, month={Aug}, pages={e39846} }

@article{summarization_2, title={Patient Information Summarization in Clinical Settings: Scoping Review}, volume={11}, url={https://medinform.jmir.org/2023/1/e44639}, DOI={https://doi.org/10.2196/44639}, number={1}, journal={JMIR Medical Informatics}, author={Keszthelyi, Daniel and Gaudet-Blavignac, Christophe and Bjelogrlic, Mina and Lovis, Christian}, year={2023}, month={Nov}, pages={e44639} }

@article{summarization_3, title={Teaching counseling microskills to audiology students: recommendations from professional counseling educators}, volume={39}, url={https://www.ncbi.nlm.nih.gov/pmc/articles/PMC5802983/}, DOI={https://doi.org/10.1055/s-0037-1613709}, number={1}, journal={Seminars in Hearing}, author={Kulzer, Jamie and Beck, Kelly}, year={2018}, month={Feb}, pages={91–106} }

@article{summarization_4_patientcentered_personcentered_language_3, title={The Critical Role and Effects of Patient-Centered Communication in Psychotherapy: A Narrative Review}, volume={Volume 18}, url={https://www.dovepress.com/the-critical-role-and-effects-of-patient-centered-communication-in-psy-peer-reviewed-fulltext-article-PRBM}, DOI={https://doi.org/10.2147/prbm.s528343}, journal={Psychology Research and Behavior Management}, publisher={Informa UK Limited}, author={Niu, Yaohong and Sun, Jingbo and Zhu, Kerun and Xu, Bojun and Zhang, Yin-Ping and Peng, Min}, year={2025}, month={Aug}, pages={1657–1671} }

@article{summarization_5, title={Exploring the Efficacy of Large Language Models in Summarizing Mental Health Counseling Sessions: Benchmark Study}, volume={11}, DOI={https://doi.org/10.2196/57306}, journal={JMIR Mental Health}, publisher={JMIR Publications}, author={Prottay Kumar Adhikary and Srivastava, Aseem and Kumar, Shivani and Salam Michael Singh and Puneet Manuja and Gopinath, Jini K and Krishnan, Vijay and Swati Kedia Gupta and Koushik Sinha Deb and Chakraborty, Tanmoy}, year={2024}, month={Jul}, pages={e57306–e57306} }

@article{cognitive_reframing_1, title={Cognitive restructuring and psychotherapy outcome: A meta-analytic review.}, volume={60}, url={https://pmc.ncbi.nlm.nih.gov/articles/PMC10440210/}, DOI={https://doi.org/10.1037/pst0000474}, number={3}, journal={Psychotherapy}, author={Ezawa, Iony D. and Hollon, Steven D.}, year={2023}, month={Mar}, pages={396–406} }

@article{cognitive_reframing_2, title={Cognitive Restructuring during Depressive Symptoms: A Scoping Review}, volume={12}, url={https://www.mdpi.com/2227-9032/12/13/1292}, DOI={https://doi.org/10.3390/healthcare12131292}, number={13}, journal={Healthcare}, author={Santos, Bruno and Pinho, Lara and Nogueira, Maria José and Pires, Regina and Sequeira, Carlos and Montesó-Curto, Pilar}, year={2024}, month={Jan}, pages={1–14} }

@article{cognitive_reframing_3, title={Durable effects of cognitive restructuring on conditioned fear.}, volume={12}, DOI={https://doi.org/10.1037/a0029143}, number={6}, journal={Emotion}, author={Shurick, Ashley A. and Hamilton, Jeffrey R. and Harris, Lasana T. and Roy, Amy K. and Gross, James J. and Phelps, Elizabeth A.}, year={2012}, pages={1393–1397} }

@article{cognitive_reframing_4, title={Understanding Mental Health and Cognitive Restructuring With Ecological Neuroscience}, volume={12}, DOI={https://doi.org/10.3389/fpsyt.2021.697095}, journal={Frontiers in Psychiatry}, author={Crum, James}, year={2021}, month={Jun} }

@article{cognitive_reframing_5, title={Practitioner Cognitive Reframing: Working More Effectively in Addictions}, volume={34}, url={https://pmc.ncbi.nlm.nih.gov/articles/PMC6370426/}, number={8}, journal={Federal Practitioner}, author={Madrigal, Karen Burkart}, year={2017}, month={Aug}, pages={26-27} }

@article{agenda_setting_1, title={Quantifying the Association Between Psychotherapy Content and Clinical Outcomes Using Deep Learning}, volume={77}, DOI={https://doi.org/10.1001/jamapsychiatry.2019.2664}, number={1}, journal={JAMA Psychiatry}, author={Ewbank, Michael P. and Cummins, Ronan and Tablan, Valentin and Bateup, Sarah and Catarino, Ana and Martin, Alan J. and Blackwell, Andrew D.}, year={2020}, month={Jan}, pages={35-43} }

@article{agenda_setting_2, title={Agenda setting in psychiatric consultations: An exploratory study.}, volume={36}, DOI={https://doi.org/10.1037/prj0000004}, number={3}, journal={Psychiatric Rehabilitation Journal}, author={Frankel, Richard M. and Salyers, Michelle P. and Bonfils, Kelsey A. and Oles, Sylwia K. and Matthias, Marianne S.}, year={2013}, pages={195–201} }

@article{agenda_setting_3, title={Agenda-setting in the clinical encounter: A systematic review protocol}, volume={19}, DOI={https://doi.org/10.1371/journal.pone.0312613}, number={10}, journal={PLoS ONE}, publisher={Public Library of Science}, author = {Sierpe, Ailyn and
          Yen, Renata W. and
          Stevens, Gabrielle and
          Van Citters, Aricca D. and
          Elwyn, Glyn and
          Saunders, Catherine H.}, year={2024}, month={Oct}, pages={e0312613–e0312613} }

@article{agenda_setting_4, title={Centering the Voice of the Client: On Becoming a Collaborative Practitioner with Low‐Income Individuals and Families}, volume={60}, DOI={https://doi.org/10.1111/famp.12558}, number={2}, journal={Family Process}, author={Falicov, Celia and Nakash, Ora and Alegría, Margarita}, year={2020}, month={Aug}, pages={670–687} }

@article{agenda_setting_5, title={Agenda setting and visit openings in primary care visits involving patients taking opioids for chronic pain}, volume={22}, DOI={https://doi.org/10.1186/s12875-020-01317-4}, number={1}, journal={BMC Family Practice}, author={Hood-Medland, Eve Angeline and White, Anne E. C. and Kravitz, Richard L. and Henry, Stephen G.}, year={2021}, month={Jan} }

@article{psychoeducation_1, title={Psychoeducation: A Basic Psychotherapeutic Intervention for Patients With Schizophrenia and Their Families}, volume={32}, url={https://www.ncbi.nlm.nih.gov/pmc/articles/PMC2683741/}, DOI={https://doi.org/10.1093/schbul/sbl017}, number={Suppl 1}, journal={Schizophrenia Bulletin}, author = {B{\"a}uml, Josef and
          Frob{\"o}se, Teresa and
          Kraemer, Sibylle and
          Rentrop, Michael and
          Pitschel-Walz, Gabriele}, year={2006}, month={Aug}, pages={S1–S9} }

@article{psychoeducation_2, title={Clinical Practice Guidelines for Psychoeducation in Psychiatric Disorders General Principles of Psychoeducation}, volume={62}, url={https://pmc.ncbi.nlm.nih.gov/articles/PMC7001357/}, DOI={https://doi.org/10.4103/psychiatry.indianjpsychiatry_780_19}, number={Suppl 2}, journal={Indian Journal of Psychiatry}, author={Sarkhel, Sujit and Singh, OP and Arora, Manu}, year={2020}, pages={S319–S323} }

@article{psychoeducation_3, title={Psychoeducation in Bipolar disorder: a Systematic Review}, volume={11}, url={https://pmc.ncbi.nlm.nih.gov/articles/PMC8717031/}, DOI={https://doi.org/10.5498/wjp.v11.i12.1407}, number={12}, journal={World Journal of Psychiatry}, author={Rabelo, Juliana Lemos and Cruz, Breno Fiuza and Ferreira, Jéssica Diniz Rodrigues and Viana, Bernardo de Mattos and Barbosa, Izabela Guimarães}, year={2021}, month={Dec}, pages={1407–1424} }

@article{psychoeducation_4, title={Psychoeducation for depression, anxiety and psychological distress: a meta-analysis}, volume={7}, DOI={https://doi.org/10.1186/1741-7015-7-79}, number={1}, journal={BMC Medicine}, author={Donker, Tara and Griffiths, Kathleen M and Cuijpers, Pim and Christensen, Helen}, year={2009}, month={Dec} }

@article{psychoeducation_5, title={The principles and practices of psychoeducation with alcohol or other drug use disorders: A review and brief guide}, volume={126}, DOI={https://doi.org/10.1016/j.jsat.2021.108442}, number={1}, journal={Journal of Substance Abuse Treatment}, author={Magill, Molly and Martino, Steve and Wampold, Bruce}, year={2021}, month={Jul}, pages={108442} }

@article{autonomy_support_3, title={Autonomy can support affect regulation during illness and in health}, volume={25}, url={https://pmc.ncbi.nlm.nih.gov/articles/PMC6933086/}, DOI={https://doi.org/10.1177/1359105318787013}, number={1}, journal={Journal of Health Psychology}, author={Cosme, Danielle and Berkman, Elliot T}, year={2018}, month={Jul}, pages={31–37} }

@article{autonomy_support_4, title={Impact of Autonomy Support on the Association between Personal Control, Healthy Behaviors, and Psychological Well-Being among Patients with Hypertension and Cardiovascular Comorbidities}, volume={19}, DOI={https://doi.org/10.3390/ijerph19074132}, number={7}, journal={International Journal of Environmental Research and Public Health}, author={Yeom, Hyun-E and Lee, Jungmin}, year={2022}, month={Mar}, pages={4132} }

@article{autonomy_support_5, title={Strengthening autonomy in mental health care through a relational approach}, volume={2}, DOI={https://doi.org/10.1038/s44220-024-00337-8}, number={11}, journal={Nature Mental Health}, author={Buedo, Paola and Daly, Timothy}, year={2024}, month={Nov}, pages={1271–1272} }

@article{change_talk_2, title={Change Talk and Relatedness in Group Motivational Interviewing: A Pilot Study}, volume={51}, url={https://www.ncbi.nlm.nih.gov/pmc/articles/PMC4737553/}, DOI={https://doi.org/10.1016/j.jsat.2014.11.003}, journal={Journal of substance abuse treatment}, author={Shorey, Ryan C. and Martino, Steve and Lamb, Kayla E. and LaRowe, Steven D. and Santa Ana, Elizabeth J.}, year={2015}, month={Apr}, pages={75–81} }

@article{change_talk_3, title={The role of therapist MI skill and client change talk class membership predicting dual alcohol and sex risk outcomes}, volume={75}, DOI={https://doi.org/10.1002/jclp.22798}, number={9}, journal={Journal of Clinical Psychology}, author={Janssen, Tim and Magill, Molly and Mastroleo, Nadine R. and Laws, M. Barton and Howe, Chanelle J. and Walthers, Justin W. and Monti, Peter M. and Kahler, Christopher W.}, year={2019}, month={Apr}, pages={1527–1543} }

@article{change_talk_4, title={Which Individual Therapist Behaviors Elicit Client Change Talk and Sustain Talk in Motivational Interviewing?}, volume={61}, url={https://www.ncbi.nlm.nih.gov/pubmed/26547412}, DOI={https://doi.org/10.1016/j.jsat.2015.09.001}, journal={Journal of substance abuse treatment}, author={Apodaca, Timothy R and Jackson, Kristina M and Borsari, Brian and Magill, Molly and Longabaugh, Richard and Mastroleo, Nadine R and Barnett, Nancy P}, year={2016}, pages={60–5} }

@article{change_talk_5, title={Motivational interviewing: an evidence-based Approach for Use in Medical Practice}, volume={118}, url={https://pmc.ncbi.nlm.nih.gov/articles/PMC8200683/}, DOI={https://doi.org/10.3238/arztebl.m2021.0014}, number={7}, journal={Deutsches Aerzteblatt Online}, author={Bischof, Gallus and Bischof, Anja and Rumpf, Hans-Jürgen}, year={2021}, pages={109–115} }

@article{strength_identification_1, title={Strength-based methods – a narrative review and comparative multilevel meta-analysis of positive interventions in clinical settings}, volume={33}, url={https://www.tandfonline.com/doi/full/10.1080/10503307.2023.2181718}, DOI={https://doi.org/10.1080/10503307.2023.2181718}, number={7}, journal={Psychotherapy Research}, author={Flückiger, Christoph and Munder, Thomas and Del Re, A. C. and Solomonov, Nili}, year={2023}, pages={1–17} }

@article{strength_identification_2, title={Accentuate the Positive: Strengths-Based Therapy for Adolescents}, volume={10}, DOI={https://doi.org/10.2174/2210676610666200225105529}, number={3}, journal={Adolescent Psychiatry}, author={Yuen, Eunice and Sadhu, Julie and Pfeffer, Cynthia and Sarvet, Barry and Daily, R. Susan and Dowben, Jonathan and Jackson, Kamilah and Schowalter, John and Shapiro, Theodore and Stubbe, Dorothy}, year={2020}, pages={166–171} }

@article{strength_identification_3, title={Strengths-Based Approach for Mental Health Recovery}, volume={7}, url={https://pmc.ncbi.nlm.nih.gov/articles/PMC3939995/}, number={2}, journal={Iranian Journal of Psychiatry and Behavioral Sciences}, author={Xie, Huiting}, year={2013}, pages={5-10} }

@article{strength_identification_4, title={The Practice of Character Strengths: Unifying Definitions, Principles, and Exploration of What’s Soaring, Emerging, and Ripe With Potential in Science and in Practice}, volume={11}, DOI={https://doi.org/10.3389/fpsyg.2020.590220}, number={590220}, journal={Frontiers in Psychology}, author={Niemiec, Ryan M. and Pearce, Ruth}, year={2021}, month={Jan} }

@article{homework_assignment_1, title={What is the effect of homework engagement in group cognitive behavioral therapy for anxiety disorders and depression?}, volume={13}, DOI={https://doi.org/10.1186/s40359-025-03167-0}, number={1}, journal={BMC Psychology}, publisher={Springer Science and Business Media LLC}, author={Hovmand, Oliver Rumle and Falkenström, Fredrik and Reinholt, Nina and Bryde, Anne and Eskildsen, Anita and Arendt, Mikkel and Poulsen, Stig and Hvenegaard, Morten and Arnfred, Sidse M. and Bach, Bo Sayyad}, year={2025}, month={Sep} }

@article{homework_assignment_2, title={The Relationship Between Homework Compliance and Therapy Outcomes: An Updated Meta-Analysis}, volume={34}, url={https://www.ncbi.nlm.nih.gov/pmc/articles/PMC2939342/}, DOI={https://doi.org/10.1007/s10608-010-9297-z}, number={5}, journal={Cognitive Therapy and Research}, author={Mausbach, Brent T. and Moore, Raeanne and Roesch, Scott and Cardenas, Veronica and Patterson, Thomas L.}, year={2010}, month={Feb}, pages={429–438} }

@article{homework_assignment_3, title={Supporting Homework Compliance in Cognitive Behavioural Therapy: Essential Features of Mobile Apps}, volume={4}, url={https://www.ncbi.nlm.nih.gov/pmc/articles/PMC5481663/}, DOI={https://doi.org/10.2196/mental.5283}, number={2}, journal={JMIR Mental Health}, author={Tang, Wei and Kreindler, David}, year={2017} }

@article{homework_assignment_4, title={Homework in Cognitive Behavioral Supervision: Theoretical Background and Clinical Application}, volume={15}, DOI={https://doi.org/10.2147/prbm.s382246}, number={1}, journal={Psychology Research and Behavior Management}, author={Prasko, Jan and Krone, Ilona and Burkauskas, Julius and Vanek, Jakub and Abeltina, Marija and Juskiene, Alicja and Sollar, Tomas and Bite, Ieva and Slepecky, Milos and Ociskova, Marie}, year={2022}, month={Dec}, pages={3809–3824} }

@article{homework_assignment_5, title={Enhancing Therapeutic Impact and Therapeutic Alliance Through Electronic Mail Homework Assignments}, volume={9}, url={https://pmc.ncbi.nlm.nih.gov/articles/PMC3330606/}, number={4}, journal={The Journal of Psychotherapy Practice and Research}, author={Murdoch, Janice W and Connor-Greene, Patricia A}, year={2000}, pages={232-237} }

@article{feedback_solicitation_1, title={Client feedback in psychological therapy for children and adolescents with mental health problems}, volume={8}, DOI={https://doi.org/10.1002/14651858.cd011729.pub2}, number={8}, journal={Cochrane Database of Systematic Reviews}, author={Bergman, Hanna and Kornør, Hege and Nikolakopoulou, Adriani and Hanssen-Bauer, Ketil and Soares-Weiser, Karla and Tollefsen, Thomas K and Bjørndal, Arild}, year={2018}, month={Aug} }

@article{feedback_solicitation_2, title={Routine Outcome Monitoring and Clinical Feedback in Psychotherapy: Recent Advances and Future Directions}, volume={51}, DOI={https://doi.org/10.1007/s10488-024-01351-9}, number={3}, journal={Administration and Policy in Mental Health and Mental Health Services Research}, publisher={Springer Science+Business Media}, author={McAleavey, Andrew A and Kim de Jong and Nissen-Lie, Helene A and Boswell, James F and Moltu, Christian and Lutz, Wolfgang}, year={2024}, month={Feb} }

@article{feedback_solicitation_3, title={Effects of Therapist Feedback on the Therapeutic Alliance and Alcohol Use Outcomes in the Outpatient Treatment of Alcohol Use Disorder}, volume={44}, DOI={https://doi.org/10.1111/acer.14297}, number={4}, journal={Alcoholism: Clinical and Experimental Research}, author={Maisto, Stephen A. and Schlauch, Robert C. and Connors, Gerard J. and Dearing, Ronda L. and O’Hern, Kelly A.}, year={2020}, month={Mar}, pages={960–972} }

@article{feedback_solicitation_4, title={Monitoring Treatment Progress and Providing Feedback is Viewed Favorably but Rarely Used in Practice}, volume={45}, url={https://pmc.ncbi.nlm.nih.gov/articles/PMC5495625/}, DOI={https://doi.org/10.1007/s10488-016-0763-0}, number={1}, journal={Administration and Policy in Mental Health and Mental Health Services Research}, author={Jensen-Doss, Amanda and Haimes, Emily M. Becker and Smith, Ashley M. and Lyon, Aaron R. and Lewis, Cara C. and Stanick, Cameo F. and Hawley, Kristin M.}, year={2018}, month={Sep}, pages={48–61} }

@article{feedback_solicitation_5, title={Understanding and Using Patient Experience Feedback to Improve Health Care Quality: Systematic Review and Framework Development}, volume={4}, url={https://pmc.ncbi.nlm.nih.gov/articles/PMC6664367/}, DOI={https://doi.org/10.17294/2330-0698.1416}, number={1}, journal={Journal of Patient-Centered Research and Reviews}, author={Kumah, Emmanuel and Osei-Kesse, Felix and Anaba, Cynthia}, year={2017}, pages={24–31} }

@article{normalizing_2, title={The Effect of an Educating versus Normalizing Approach on Treatment Motivation in Patients Presenting with Delusions: An Experimental Investigation with Analogue Patients}, volume={2013}, DOI={https://doi.org/10.1155/2013/261587}, number={1}, journal={Schizophrenia Research and Treatment}, author={Lüllmann, Eva and Lincoln, Tania M.}, year={2013}, pages={1–8} }

@article{normalizing_3, title={The Importance of Responding to Negative Affect in Psychotherapies}, volume={168}, url={https://www.ncbi.nlm.nih.gov/pmc/articles/PMC3627537/}, DOI={https://doi.org/10.1176/appi.ajp.2010.10040636}, number={2}, journal={American Journal of Psychiatry}, author={Markowitz, John C. and Milrod, Barbara L.}, year={2011}, month={Feb}, pages={124–128} }

@article{normalizing_4, title={In support of supportive psychotherapy}, volume={21}, DOI={https://doi.org/10.1002/wps.20949}, number={1}, journal={World Psychiatry}, author={Markowitz, John C.}, year={2022}, month={Jan}, pages={59–60} }

@article{normalizing_5, title={Mental health literacy for anxiety disorders: how perceptions of symptom severity might relate to recognition of psychological distress}, volume={14}, url={https://www.ncbi.nlm.nih.gov/pmc/articles/PMC4755316/}, DOI={https://doi.org/10.1108/jpmh-09-2013-0064}, number={2}, journal={Journal of Public Mental Health}, author={Paulus, Daniel J. and Wadsworth, Lauren Page and Hayes-Skelton, Sarah A.}, year={2015}, month={Jun}, pages={94–106} }

@article{instillation_of_hope_1, title={Examining Hope as a Transdiagnostic Mechanism of Change Across Anxiety Disorders and CBT Treatment Protocols}, volume={51}, DOI={https://doi.org/10.1016/j.beth.2019.06.001}, number={1}, journal={Behavior Therapy}, author={Gallagher, Matthew W. and Long, Laura J. and Richardson, Angela and D’Souza, Johann and Boswell, James F. and Farchione, Todd J. and Barlow, David H.}, year={2020}, month={Jan}, pages={190–202} }

@article{instillation_of_hope_2, title={Identifying Yalom’s group therapeutic factors in anonymous mental health discussions on Reddit: a mixed-methods analysis using large language models, topic modeling and human supervision}, volume={16}, DOI={https://doi.org/10.3389/fpsyt.2025.1503427}, journal={Frontiers in Psychiatry}, publisher={Frontiers Media SA}, author={Ferizaj, Drin and Lalk, Christopher and Lahmann, Nils and Strube-Lahmann, Sandra and Rubel, Julian}, year={2025}, month={Jun} }

@article{instillation_of_hope_3, title={Mechanisms of Change in Cognitive Processing Therapy and Prolonged Exposure Therapy for PTSD: Preliminary Evidence for the Differential Effects of Hopelessness and Habituation}, volume={36}, DOI={https://doi.org/10.1007/s10608-011-9423-6}, number={6}, journal={Cognitive Therapy and Research}, author={Gallagher, Matthew W. and Resick, Patricia A.}, year={2011}, month={Nov}, pages={750–755} }

@article{instillation_of_hope_4, title={A Brief Hope Intervention to Increase Hope Level and Improve Well-Being in Rehabilitating Cancer Patients: A Feasibility Test}, volume={5}, DOI={https://doi.org/10.1177/2377960819844381}, journal={SAGE Open Nursing}, author={Chan, Kitty and Wong, Frances K. Y. and Lee, Paul H.}, year={2019}, month={Jan}, pages={237796081984438} }

@article{instillation_of_hope_5, title={Interdisciplinary approach of Yalom’s group therapy factors: A theoretical model for including animal presence in social work education and practice}, volume={9}, url={https://www.ncbi.nlm.nih.gov/pmc/articles/PMC9596786/}, DOI={https://doi.org/10.3389/fvets.2022.1024355}, journal={Frontiers in Veterinary Science}, author={Rusu, Alina Simona and Davis, Rebecca}, year={2022}, month={Oct}, pages={1024355} }

@article{experiential_learning_role_playing_1, title={The use of clinical role-play and reflection in learning therapeutic communication skills in mental health education: An integrative review}, volume={10}, url={https://www.ncbi.nlm.nih.gov/pmc/articles/PMC6593356/}, DOI={https://doi.org/10.2147/amep.s202115}, journal={Advances in Medical Education and Practice}, author={Rønning, Solrun Brenk and Bjørkly, Stål}, year={2019}, month={Jun}, pages={415–425} }

@article{experiential_learning_role_playing_2, title={Strategies that promote therapist engagement in active and experiential learning: micro-level sequential analysis}, volume={40}, DOI={https://doi.org/10.1080/07325223.2020.1870023}, number={1}, journal={The Clinical Supervisor}, author={Caron, EB and Lind, Teresa A. and Dozier, Mary}, year={2021}, month={Jan}, pages={112–133} }

@article{experiential_learning_role_playing_3, title={Promoting Personal Growth through Experiential Learning: The Case of Expressive Arts Therapy for Lecturers in Thailand}, volume={8}, DOI={https://doi.org/10.3389/fpsyg.2017.02276}, journal={Frontiers in Psychology}, author={Binson, Bussakorn and Lev-Wiesel, Rachel}, year={2018}, month={Feb} }

@article{experiential_learning_role_playing_4, title={Exploring the Impact of an Innovative Peer Role-Play Simulation to Cultivate Student Pharmacists’ Motivational Interviewing Skills}, volume={11}, url={https://www.mdpi.com/2226-4787/11/4/122}, DOI={https://doi.org/10.3390/pharmacy11040122}, number={4}, journal={Pharmacy}, author={Denvir, Paul and Briceland, Laurie L.}, year={2023}, month={Aug}, pages={122} }

@article{experiential_learning_role_playing_5, title={Role-play Games (RPGs) for Mental Health (Why Not?): Roll for Initiative}, volume={21}, DOI={https://doi.org/10.1007/s11469-022-00832-y},  journal={International Journal of Mental Health and Addiction}, author={Baker, Ian S. and Turner, Ian J. and Kotera, Yasuhiro}, year={2022}, month={May} }

@article{problem_solving_1, title={Psychotherapy of Mood Disorders}, volume={10}, url={https://www.ncbi.nlm.nih.gov/pmc/articles/PMC4258697/}, DOI={https://doi.org/10.2174/1745017901410010140}, number={1}, journal={Clinical Practice \& Epidemiology in Mental Health}, author={Picardi, Angelo and Gaetano, Paola}, year={2014}, month={Nov}, pages={140–158} }

@article{problem_solving_2, title={Problem Solving Therapy Improves Effortful Cognition in Major Depression}, volume={12}, DOI={https://doi.org/10.3389/fpsyt.2021.607718}, journal={Frontiers in Psychiatry}, author={Jiang, Chenguang and Zhou, Hongliang and Chen, Limin and Zhou, Zhenhe}, year={2021}, month={Apr} }

@article{problem_solving_3, title={Problem-solving training as an active ingredient of treatment for youth depression: a scoping review and exploratory meta-analysis}, volume={21}, DOI={https://doi.org/10.1186/s12888-021-03260-9}, number={1}, journal={BMC Psychiatry}, author={Krause, Karolin R. and Courtney, Darren B. and Chan, Benjamin W. C. and Bonato, Sarah and Aitken, Madison and Relihan, Jacqueline and Prebeg, Matthew and Darnay, Karleigh and Hawke, Lisa D. and Watson, Priya and Szatmari, Peter}, year={2021}, month={Aug} }

@article{problem_solving_4, title={Problem-Solving Therapy and Supportive Therapy in Older Adults With Major Depression and Executive Dysfunction}, volume={167}, DOI={https://doi.org/10.1176/appi.ajp.2010.09091327}, number={11}, journal={American Journal of Psychiatry}, author={Areán, Patricia A. and Raue, Patrick and Mackin, R. Scott and Kanellopoulos, Dora and McCulloch, Charles and Alexopoulos, George S.}, year={2010}, month={Nov}, pages={1391–1398} }

@article{problem_solving_5, title={Problem-solving interventions and depression among adolescents and young adults: A systematic review of the effectiveness of problem-solving interventions in preventing or treating depression}, volume={18}, url={https://www.ncbi.nlm.nih.gov/pmc/articles/PMC10464969/}, DOI={https://doi.org/10.1371/journal.pone.0285949}, number={8}, journal={PLOS ONE}, publisher={Public Library of Science}, author={Metz, Kristina and Lewis, Jane A and Mitchell, Jade and Chakraborty, Sangita and McLeod, Bryce D and Ludvig Daae Bjørndal and Mildon, Robyn and Shlonsky, Aron}, year={2023}, month={Aug}, pages={e0285949–e0285949} }

@article{emotion_regulation_skills_training_1, title={An emotion regulation skills training for adolescents and parents: perceptions and acceptability of methodological aspects}, volume={15}, DOI={https://doi.org/10.3389/fpsyt.2024.1448529}, journal={Frontiers in Psychiatry}, publisher={Frontiers Media}, author={Larsson, Kristina Holmqvist and Zetterqvist, Maria}, year={2024}, month={Dec} }

@article{emotion_regulation_skills_training_2, title={From surviving to thriving in the face of threats: the emerging science of emotion regulation training}, volume={24}, url={https://www.sciencedirect.com/science/article/pii/S2352154618300196}, DOI={https://doi.org/10.1016/j.cobeha.2018.08.007}, journal={Current Opinion in Behavioral Sciences}, author={Cohen, Noga and Ochsner, Kevin N}, year={2018}, month={Dec}, pages={143–155} }

@article{emotion_regulation_skills_training_3, title={Emotion Regulation Therapy for Generalized Anxiety Disorder}, volume={20}, url={https://www.ncbi.nlm.nih.gov/pmc/articles/PMC4973631/}, DOI={https://doi.org/10.1016/j.cbpra.2013.02.001}, number={3}, journal={Cognitive and Behavioral Practice}, author={Fresco, David M. and Mennin, Douglas S. and Heimberg, Richard G. and Ritter, Michael}, year={2013}, month={Aug}, pages={282–300} }

@article{emotion_regulation_skills_training_4, title={The Effectiveness of Emotion Regulation Training and Cognitive Therapy on the Emotional and Addictional Problems of Substance Abusers}, volume={5}, url={https://pmc.ncbi.nlm.nih.gov/articles/PMC3430496/}, number={2}, journal={Iranian Journal of Psychiatry}, author={Azizi, Alireza and Borjali, Ahmad and Mahmoud Golzari}, year={2010}, pages={60-65} }

@article{emotion_regulation_skills_training_5, title={Emotion Regulation Enhancement of Cognitive Behavior Therapy for College Student Problem Drinkers: A Pilot Randomized Controlled Trial}, volume={27}, DOI={https://doi.org/10.1080/1067828x.2017.1400484}, number={1}, journal={Journal of Child \& Adolescent Substance Abuse}, author={Ford, Julian D. and Grasso, Damion J. and Levine, Joan and Tennen, Howard}, year={2017}, month={Dec}, pages={47–58} }

@article{relapse_prevention_planning_1, title={Developing and Implementing a Web-Based Relapse Prevention Psychotherapy Program for Patients With Alcohol Use Disorder: Protocol for a Randomized Controlled Trial}, volume={12}, DOI={https://doi.org/10.2196/44694}, journal={JMIR Research Protocols}, author={Eadie, Jazmin and Gutierrez, Gilmar and Moghimi, Elnaz and Stephenson, Callum and Khalafi, Payam and Nikjoo, Niloofar and Jagayat, Jasleen and Gizzarelli, Tessa and Reshetukha, Taras and Omrani, Mohsen and Yang, Megan and Alavi, Nazanin}, year={2023}, month={Jan}, pages={e44694} }

@article{relapse_prevention_planning_2, title={Preventing Relapse Following Smoking Cessation}, volume={4}, url={https://www.ncbi.nlm.nih.gov/pmc/articles/PMC4636196/}, DOI={https://doi.org/10.1007/s12170-010-0124-6}, number={6}, journal={Current cardiovascular risk reports}, author={Collins, Susan E. and Witkiewitz, Katie and Kirouac, Megan and Marlatt, G. Alan}, year={2010}, month={Nov}, pages={421–428} }

@article{relapse_prevention_planning_3, title={Integrated treatment of substance use and psychiatric disorders}, volume={28}, DOI={https://doi.org/10.1080/19371918.2013.774673}, number={3-4}, journal={Social Work in Public Health}, author={Kelly, Thomas M. and Daley, Dennis C.}, year={2013}, month={May}, pages={388–406} }

@article{relapse_prevention_planning_4, title={Relapse Prevention}, volume={60}, url={https://www.ncbi.nlm.nih.gov/pmc/articles/PMC5844157/}, DOI={https://doi.org/10.4103/psychiatry.IndianJPsychiatry_36_18}, number={Suppl 4}, journal={Indian Journal of Psychiatry}, author={Menon, Jayakrishnan and Kandasamy, Arun}, year={2018}, pages={S473–S478} }

@article{relapse_prevention_planning_5, title={Relapse Prevention and the Five Rules of Recovery}, volume={88}, url={https://pmc.ncbi.nlm.nih.gov/articles/PMC4553654/}, number={3}, journal={The Yale Journal of Biology and Medicine}, author={Melemis, Steven M}, year={2015}, month={Sep}, pages={325-332} }

@article{client_monitoring_progress_tracking_1, title={Routine provision of feedback from patient-reported outcome measurements to healthcare providers and patients in clinical practice}, volume={2021}, DOI={https://doi.org/10.1002/14651858.cd011589.pub2}, number={10}, journal={Cochrane Database of Systematic Reviews}, author={Gibbons, Chris and Porter, Ian and Gonçalves-Bradley, Daniela C and Stoilov, Stanimir and Ricci-Cabello, Ignacio and Tsangaris, Elena and Gangannagaripalli, Jaheeda and Davey, Antoinette and Gibbons, Elizabeth J and Kotzeva, Anna and Evans, Jonathan and van der Wees, Philip J and Kontopantelis, Evangelos and Greenhalgh, Joanne and Bower, Peter and Alonso, Jordi and Valderas, Jose M}, year={2021}, month={Oct} }

@article{client_monitoring_progress_tracking_2, title={Attitudes of Mental Health Professionals Towards the Use of Routine Outcome Monitoring in Psychotherapeutic Inpatient Settings: A Thematic Analysis}, volume={52}, DOI={https://doi.org/10.1007/s10488-025-01455-w}, number={4}, journal={Administration and Policy in Mental Health and Mental Health Services Research}, author={Krakowczyk, Julia Barbara and Teufel, Martin and Skoda, Eva-Maria and Jansen, Christoph and Lalgi, Tania and Martens, Lennart and Dinger, Ulrike and Lutz, Wolfgang and B\"auerle, Alexander}, year={2025}, month={Jul} }

@article{client_monitoring_progress_tracking_3, title={Psychotherapists’ Experience with In-Session Use of Routine Outcome Monitoring: A Qualitative Meta-analysis}, volume={52}, DOI={https://doi.org/10.1007/s10488-024-01348-4}, number={1}, journal={Administration and policy in mental health}, publisher={Springer Science+Business Media}, author={Klára Jonášová and Čevelíček, Michal and Petr Doležal and Tomáš Řiháček}, year={2024}, month={Mar} }

@article{client_monitoring_progress_tracking_4, title={Motives of Therapists for Using Routine Outcome Monitoring (ROM) and How it is Used by Them in Clinical Practice: Two Qualitative Studies}, volume={52}, DOI={https://doi.org/10.1007/s10488-024-01374-2}, number={1}, journal={Administration and policy in mental health}, publisher={Springer Science+Business Media}, author={Shaghayegh Azizian Kia and Wittkampf, Lisette and Jacobine van Lankeren and Janse, Pauline}, year={2024}, month={Apr} }

@article{client_monitoring_progress_tracking_5, title={Using routine outcome measures as clinical process tools: Maximising the therapeutic yield in the IAPT programme when working remotely}, volume={95}, DOI={https://doi.org/10.1111/papt.12400}, number={3}, journal={Psychology and Psychotherapy: Theory, Research and Practice}, author={Faija, Cintia L. and Bee, Penny and Lovell, Karina and Lidbetter, Nicky and Gellatly, Judith and Ardern, Kerry and Rushton, Kelly and Brooks, Helen and McMillan, Dean and Armitage, Christopher J. and Woodhouse, Rebecca and Barkham, Michael}, year={2022}, month={May} }

@article{humor_in_therapy_1, title={Humor interventions in psychotherapy and their effect on levels of depression and anxiety in adult clients, a systematic review}, volume={13}, url={https://www.ncbi.nlm.nih.gov/pmc/articles/PMC9845902/}, DOI={https://doi.org/10.3389/fpsyt.2022.1049476}, journal={Frontiers in Psychiatry}, author={Sarink, Federico S. M. and García-Montes, José M.}, year={2023}, month={Jan} }

@article{humor_in_therapy_2, title={Humor in Psychiatry: Lessons From Neuroscience, Psychopathology, and Treatment Research}, volume={12}, DOI={https://doi.org/10.3389/fpsyt.2021.681903}, journal={Frontiers in Psychiatry}, author={Berger, Philipp and Bitsch, Florian and Falkenberg, Irina}, year={2021}, month={May} }

@article{humor_in_therapy_3, title={The impact of humor therapy on people suffering from depression or anxiety: An integrative literature review}, volume={13}, DOI={https://doi.org/10.1002/brb3.3108}, number={9}, journal={Brain and behavior}, publisher={Wiley-Blackwell}, author={Sun, Xinwei and Zhang, Jindan and Wang, Yidan and Zhang, X and Li, Sixuan and Qu, Zhigang and Zhang, Hongshi}, year={2023}, month={Jun} }

@article{humor_in_therapy_4, title={The Use of Humor in Serious Mental Illness: A Review}, volume={2011}, url={https://www.hindawi.com/journals/ecam/2011/342837/}, DOI={https://doi.org/10.1093/ecam/nep106}, journal={Evidence-Based Complementary and Alternative Medicine}, author={Gelkopf, Marc}, year={2011}, pages={1–8} }

@article{humor_in_therapy_5, title={Humor Therapy: Relieving Chronic Pain and Enhancing Happiness for Older Adults}, volume={2010}, url={https://www.hindawi.com/journals/jar/2010/343574/}, DOI={https://doi.org/10.4061/2010/343574}, journal={Journal of Aging Research}, author={Tse, Mimi M. Y. and Lo, Anna P. K. and Cheng, Tracy L. Y. and Chan, Eva K. K. and Chan, Annie H. Y. and Chung, Helena S. W.}, year={2010}, pages={1–9} }

@article{affective_attunement_1, title={Understanding the role of emotion and expertise in psychotherapy: An application of dynamical systems mathematical modeling to an entire course of therapy}, volume={14}, DOI={https://doi.org/10.3389/fpsyt.2023.980739}, journal={Frontiers in Psychiatry}, publisher={Frontiers Media}, author={Diaz, Patricia and Peluso, Paul R and Freund, Robert R and Baker, Andrew Z and Pena, Gabriel S}, year={2023}, month={Apr} }

@article{affective_attunement_2, title={Development of the Therapist Empathy Scale}, volume={42}, url={https://www.cambridge.org/core/journals/behavioural-and-cognitive-psychotherapy/article/development-of-the-therapist-empathy-scale/E0070ABDC925D8535753224506D302BA}, DOI={https://doi.org/10.1017/s1352465813000039}, number={3}, journal={Behavioural and Cognitive Psychotherapy}, author={Decker, Suzanne E. and Nich, Charla and Carroll, Kathleen M. and Martino, Steve}, year={2013}, month={Mar}, pages={339–354} }

@article{affective_attunement_3, title={Changes in Affect Integration and Internalizing Symptoms After Time-Limited Intersubjective Child Psychotherapy—A Pilot Study}, volume={13}, DOI={https://doi.org/10.3389/fpsyg.2022.906416}, journal={Frontiers in Psychology}, publisher={Frontiers Media}, author={Fiskum, Charlotte and Andersen, Tonje G and Johns, Unni T and Jacobsen, Karl}, year={2022}, month={Jul}, pages={906416–906416} }

@article{affective_attunement_4, title={Movement-based patient-therapist attunement in psychological therapy and its association with early change}, volume={8}, url={https://pmc.ncbi.nlm.nih.gov/articles/PMC9520162/}, DOI={https://doi.org/10.1177/20552076221129098}, journal={Digital Health}, publisher={SAGE Publishing}, author={Schwartz, Brian and Rubel, Julian A and Anne-Katharina Deisenhofer and Lutz, Wolfgang}, year={2022}, month={Jan}, pages={205520762211290-205520762211290} }

@article{affective_attunement_5, title={Empathic Conjectures in Emotionally Focused Couple Therapy (EFCT): A Process Microanalytic Study}, volume={51}, url={https://pmc.ncbi.nlm.nih.gov/articles/PMC12445396/}, DOI={https://doi.org/10.1111/jmft.70075}, number={4}, journal={Journal of Marital and Family Therapy}, publisher={Wiley}, author={Fatahian‐Tehran, Hamed M and Chatha, Simran and Elliott, Robert}, year={2025}, month={Sep}, pages={e70075–e70075} }

@article{bodyoriented_interventions_1, title={BODY-ORIENTED THERAPY IN RECOVERY FROM CHILD SEXUAL ABUSE: AN EFFICACY STUDY}, volume={11}, url={https://pmc.ncbi.nlm.nih.gov/articles/PMC1933482/}, number={5}, journal={Alternative therapies in health and medicine}, author={Price, Cynthia}, year={2005}, pages={46-57} }

@article{bodyoriented_interventions_2, title={Body-Centered Interventions for Psychopathological Conditions: A Review}, volume={10}, url={https://pmc.ncbi.nlm.nih.gov/articles/PMC6993757/}, DOI={https://doi.org/10.3389/fpsyg.2019.02907}, journal={Frontiers in Psychology}, publisher={Frontiers Media}, author={Tarsha, Mary S and Park, Sohee and Tortora, Suzi}, year={2020}, month={Jan}, pages={2907–2907} }

@article{bodyoriented_interventions_3, title={Effectiveness of Body Psychotherapy. A Systematic Review and Meta-Analysis}, volume={12}, url={https://pmc.ncbi.nlm.nih.gov/articles/PMC8458738/}, DOI={https://doi.org/10.3389/fpsyt.2021.709798}, journal={Frontiers in Psychiatry}, publisher={Frontiers Media}, author={Rosendahl, Sophie and Heribert Sattel and Lahmann, Claas}, year={2021}, month={Sep}, pages={709798–709798} }

@article{bodyoriented_interventions_4, title={MINDFUL AWARENESS IN BODY-ORIENTED THERAPY FOR FEMALE VETERANS WITH POST-TRAUMATIC STRESS DISORDER TAKING PRESCRIPTION ANALGESICS FOR CHRONIC PAIN: A FEASIBILITY STUDY}, volume={13}, url={https://pmc.ncbi.nlm.nih.gov/articles/PMC3037268/}, number={6}, journal={Alternative therapies in health and medicine}, author={Price, Cynthia J and McBride, Brittney and Hyerle, Lynne and Kivlahan, Daniel R}, year={2007}, pages={32-40} }

@article{bodyoriented_interventions_5, title={Body‐ and Movement‐Oriented Interventions for Posttraumatic Stress Disorder: A Systematic Review and Meta‐Analysis}, volume={32}, url={https://pmc.ncbi.nlm.nih.gov/articles/PMC6973294/}, DOI={https://doi.org/10.1002/jts.22465}, number={6}, journal={Journal of Traumatic Stress}, publisher={Wiley},
author = {van de Kamp, Minke M. and
          Scheffers, Mia and
          Hatzmann, Janneke and
          Emck, Claudia and
          Cuijpers, Pim and
          Beek, Peter J.},
year={2019}, month={Oct}, pages={967–976} }

@article{termination_preparation_1, title={Termination of psychotherapy: a systematic review}, volume={4}, url={https://pmc.ncbi.nlm.nih.gov/articles/PMC12442594/}, DOI={https://doi.org/10.1080/28324765.2025.2535626}, number={1}, journal={Cogent Mental Health}, publisher={Informa UK Limited}, author={Rabinowitz, Yocheved L. and Yim, Brian and Muran, J. Christopher}, year={2025}, month={Jul} }

@article{termination_preparation_2, title={Client and therapist views of contextual factors related to termination from psychotherapy: A comparison between unilateral and mutual terminators}, volume={20}, url={https://pmc.ncbi.nlm.nih.gov/articles/PMC2924572/}, DOI={https://doi.org/10.1080/10503301003645796}, number={4}, journal={Psychotherapy Research}, publisher={Routledge}, author={Westmacott, Robin and Hunsley, John and Best, Marlene and Orly Rumstein-McKean and Schindler, Dwayne}, year={2010}, month={Jun}, pages={423–435} }

@article{termination_preparation_3, title={Termination of a Long-Term Harm Reduction Psychotherapy Group for Veterans With Substance Use Disorders}, volume={44}, url={https://pmc.ncbi.nlm.nih.gov/articles/PMC12396149/}, DOI={https://doi.org/10.1353/grp.2020.0008}, number={1}, journal={Group}, author={Ruan, Hang and Nelson, Lindsey Parrish}, year={2020}, month={Mar}, pages={27–39} }

@article{termination_preparation_4, title={Decision making on (dis)continuation of long-term treatment in mental health services is an interpersonal negotiation rather than an objective process: qualitative study}, volume={19}, url={https://pmc.ncbi.nlm.nih.gov/articles/PMC6421659/}, DOI={https://doi.org/10.1186/s12888-019-2072-0}, number={1}, journal={BMC Psychiatry}, publisher={BioMed Central}, author={B. Koekkoek and B. van Meijel and A. Perquin and G. Hutschemaekers}, year={2019}, month={Mar}, pages={92–92} }

@article{termination_preparation_5, title={Premature dropout from psychotherapy: Prevalence, perceived reasons and consequences as rated by clinicians}, volume={4}, url={https://pmc.ncbi.nlm.nih.gov/articles/PMC9667417/}, DOI={https://doi.org/10.32872/cpe.6695}, number={2}, journal={Clinical Psychology in Europe}, author={Niclas Kullgard and Holmqvist, Rolf and Andersson, Gerhard}, year={2022}, month={Jun}, pages={e6695–e6695} }

@article{managing_resistance_1, title={Examining the Therapeutic Relationship and Confronting Resistances in Psychodynamic Psychotherapy: A Certified Public Accountant Case}, volume={8}, url={https://pmc.ncbi.nlm.nih.gov/articles/PMC3115766/}, number={5}, journal={Innovations in Clinical Neuroscience}, author={Manetta, Christopher T and Gentile, Julie P and Gillig, Paulette Marie}, year={2011}, month={May}, pages={35-40} }

@article{managing_resistance_2, title={The interpersonal context of client motivational language in cognitive–behavioral therapy.}, volume={53}, url={https://pmc.ncbi.nlm.nih.gov/articles/PMC4659770/}, DOI={https://doi.org/10.1037/pst0000017}, number={1}, journal={Psychotherapy}, publisher={American Psychological Association (APA)}, author={Sijercic, Iris and Button, Melissa L. and Westra, Henny A. and Hara, Kimberley M.}, year={2016}, month={Mar}, pages={13–21} }

@article{managing_resistance_3, title={Getting to “Yes”: Overcoming Client Reluctance to Engage in Chair Work}, volume={11}, url={https://pmc.ncbi.nlm.nih.gov/articles/PMC7573290/}, DOI={https://doi.org/10.3389/fpsyg.2020.582856}, journal={Frontiers in Psychology}, publisher={Frontiers Media SA}, author={Muntigl, Peter and Horvath, Adam O. and Chubak, Lynda and Angus, Lynne}, year={2020}, month={Oct} }

@article{managing_resistance_5, title={Patient, therapist, and relational antecedents of hostile resistance in cognitive–behavioral therapy for panic disorder: A qualitative investigation.}, volume={58}, url={https://pmc.ncbi.nlm.nih.gov/articles/PMC8333228/}, DOI={https://doi.org/10.1037/pst0000308}, number={2}, journal={Psychotherapy}, publisher={American Psychological Association}, author={Schwartz, Rachel A and Chambless, Dianne L and Milrod, Barbara and Barber, Jacques P}, year={2021}, month={Feb}, pages={230–241} }

@article{positive_reinforcement_1, title={Treatment of escape-maintained challenging behavior using chained schedules: an evaluation of the effects of thinning positive plus negative reinforcement during functional communication training}, volume={62}, url={https://pmc.ncbi.nlm.nih.gov/articles/PMC5473629/}, DOI={https://doi.org/10.1080/20473869.2016.1176308}, number={3}, journal={International Journal of Developmental Disabilities}, publisher={Taylor & Francis}, author={Zangrillo, Amanda N and Fisher, Wayne W and Greer, Brian D and Owen, Todd M and DeSouza, Andresa A}, year={2016}, month={Apr}, pages={147–156} }

@article{positive_reinforcement_2, title={Improving rehabilitation motivation and motor learning ability of stroke patients using different reward strategies: study protocol for a single-center, randomized controlled trial}, volume={15}, url={https://pmc.ncbi.nlm.nih.gov/articles/PMC11178101/}, DOI={https://doi.org/10.3389/fneur.2024.1418247}, journal={Frontiers in Neurology}, publisher={Frontiers Media}, author={Zhao, Jingwang and Guo, Jiangling and Chen, Yeping and Li, Wenxi and Zhou, Ping and Zhu, Guangyue and Han, Peipei and Xu, Dongsheng}, year={2024}, month={May}, pages={1418247–1418247} }

@article{brookman-positive_reinforcement_3, title={Therapists’ Attitudes Towards Psychotherapeutic Strategies in Community-Based Psychotherapy with Children with Disruptive Behavior Problems}, volume={36}, url={https://pmc.ncbi.nlm.nih.gov/articles/PMC2657660/}, DOI={https://doi.org/10.1007/s10488-008-0195-6}, number={1}, journal={Administration and Policy in Mental Health and Mental Health Services Research}, publisher={Springer Science and Business Media LLC}, author={Brookman-Frazee, Lauren and Garland, Ann F. and Taylor, Robin and Zoffness, Rachel}, year={2008}, month={Nov}, pages={1–12} }

@article{positive_reinforcement_4, title={What builds the bond? Child and therapist behavior in a group intervention for aggression}, url={https://pmc.ncbi.nlm.nih.gov/articles/PMC12888082/}, DOI={https://doi.org/10.1017/s0954579425101119}, journal={Development and Psychopathology}, publisher={Cambridge University Press}, author={Laird, Robert D and Lochman, John E and McDonald, Kristina L and Boxmeyer, Caroline L and Powell, Nicole P and Saavedra, Lissette M and Qu, Lixin}, year={2026}, month={Jan}, pages={1–13} }

@article{positive_reinforcement_5, title={Training therapists in evidence-based practice: A critical review of studies from a systems-contextual perspective.}, volume={17}, url={https://pmc.ncbi.nlm.nih.gov/articles/PMC2945375/}, DOI={https://doi.org/10.1111/j.1468-2850.2009.01187.x}, number={1}, journal={Clinical Psychology: Science and Practice}, publisher={American Psychological Association (APA)}, author={Beidas, Rinad S. and Kendall, Philip C.}, year={2010}, month={Mar}, pages={1–30} }

@article{selfcompassion_promotion_1, title={The Effectiveness of Self-Compassion-Focused Therapy on Cognitive Vulnerability to Depression}, url={https://pmc.ncbi.nlm.nih.gov/articles/PMC10293688/}, DOI={https://doi.org/10.18502/ijps.v18i2.12364}, journal={Iranian Journal of Psychiatry}, publisher={Knowledge E DMCC}, author={Farhadi, Mehran and Rahimi, Haniyeh and Zoghi Paydar, Mohammad Reza and Yarmohamadi Vasel, Mosayeb}, year={2023}, month={Apr} }

@article{selfcompassion_promotion_2, title={Effectiveness of mindful self-compassion therapy on psychopathology symptoms, psychological distress and life expectancy in infertile women treated with in vitro fertilization: a two-arm double-blind parallel randomized controlled trial}, volume={24}, url={https://pmc.ncbi.nlm.nih.gov/articles/PMC10908010/}, DOI={https://doi.org/10.1186/s12888-023-05411-6}, number={1}, journal={BMC Psychiatry}, publisher={BioMed Central}, author={Kimia Sahraian and Ranjbar, Hamed Abdollahpour and Jahromi, Bahia Namavar and Cheung, Ho Nam and Ciarrochi, Joseph and Mojtaba Habibi Asgarabad}, year={2024}, month={Mar}, pages={174–174} }

@article{selfcompassion_promotion_3, title={The mediating role of self-compassion in the relationship between childhood trauma and the symptoms of body dysmorphic disorder}, volume={13}, url={https://pmc.ncbi.nlm.nih.gov/articles/PMC12306069/}, DOI={https://doi.org/10.1186/s40359-025-03204-y}, number={1}, journal={BMC Psychology}, publisher={Springer Science and Business Media LLC}, author={Shooroki, Motahhareh Kargar and Choobforoushzadeh, Azadeh and Ardakan, Azra Mohammadpanah}, year={2025}, month={Jul} }

@article{selfcompassion_promotion_4, title={The Benefits of Self-Compassion in Mental Health Professionals: A Systematic Review of Empirical Research}, volume={Volume 15}, url={https://pmc.ncbi.nlm.nih.gov/articles/PMC9482966/}, DOI={https://doi.org/10.2147/prbm.s359382}, journal={Psychology Research and Behavior Management}, publisher={Informa UK Limited}, author={Crego, Antonio and Yela, José Ramón and Riesco-Matías, Pablo and Gómez-Martínez, María-Ángeles and Vicente-Arruebarrena, Aitor}, year={2022}, month={Sep}, pages={2599–2620} }

@article{selfcompassion_promotion_5, title={The role of self‐compassion in psychotherapy}, volume={21}, url={https://pmc.ncbi.nlm.nih.gov/articles/PMC8751548/}, DOI={https://doi.org/10.1002/wps.20925}, number={1}, journal={World Psychiatry}, publisher={Wiley}, author={Neff, Kristin and Germer, Christopher}, year={2022}, month={Jan}, pages={58–59} }

@article{trauma_processing_1, title={Constructive and Unproductive Processing of Traumatic Experiences in Trauma-Focused Cognitive-Behavioral Therapy for Youth}, volume={48}, url={https://pmc.ncbi.nlm.nih.gov/articles/PMC5344908/}, DOI={https://doi.org/10.1016/j.beth.2016.06.004}, number={2}, journal={Behavior Therapy}, publisher={Elsevier BV}, author={Hayes, Adele M. and Yasinski, Carly and Grasso, Damion and Ready, C. Beth and Alpert, Elizabeth and McCauley, Thomas and Webb, Charles and Deblinger, Esther}, year={2017}, month={Mar}, pages={166–181} }

@article{trauma_processing_2, title={The effectiveness of trauma-focused psychotherapy for complex post-traumatic stress disorder: A retrospective study}, volume={66}, url={https://pmc.ncbi.nlm.nih.gov/articles/PMC9879871/}, DOI={https://doi.org/10.1192/j.eurpsy.2022.2346}, number={1}, journal={European Psychiatry}, publisher={Cambridge University Press}, author={Eirini Melegkovits and Blumberg, Jocelyn and Dixon, Emily and Ehntholt, Kimberley and Gillard, Julia and Kayal, Hamodi and Kember, Tim and Ottisova, Livia and Walsh, Eileen and Wood, Maximillian and Gafoor, Rafael and Brewin, Chris and Billings, Jo and Robertson, Mary and Bloomfield, Michael}, year={2022}, month={Nov}, pages={e4–e4} }

@article{trauma_processing_3, title={Teaching trauma-focused exposure therapy for PTSD: Critical clinical lessons for novice exposure therapists.}, volume={3}, url={https://pmc.ncbi.nlm.nih.gov/articles/PMC3188445/}, DOI={https://doi.org/10.1037/a0024642}, number={3}, journal={Psychological Trauma Theory Research Practice and Policy}, publisher={American Psychological Association}, author={Zoellner, Lori A and Feeny, Norah C and Bittinger, Joyce N and Bedard-Gilligan, Michele A and Slagle, David M and Post, Loren M and Chen, Jessica A}, year={2011}, month={Jan}, pages={300–308} }

@article{trauma_processing_4, title={Processes of Change in Trauma-Focused Cognitive Behavioral Therapy for Youths: An Approach Informed by Emotional Processing Theory}, volume={9}, url={https://pmc.ncbi.nlm.nih.gov/articles/PMC7984413/}, DOI={https://doi.org/10.1177/2167702620957315}, number={2}, journal={Clinical Psychological Science}, publisher={SAGE Publishing}, author={Alpert, Elizabeth and Hayes, Adele M and Yasinski, Carly and Webb, Charles and Deblinger, Esther}, year={2021}, month={Feb}, pages={270–283} }

@article{trauma_processing_5, title={Using the Trauma Reintegration Process to Treat Posttraumatic Stress Disorder with Dissociation and Somatic Features: A Case Series}, volume={13}, url={https://pmc.ncbi.nlm.nih.gov/articles/PMC12111324/}, DOI={https://doi.org/10.3390/healthcare13101092}, number={10}, journal={Healthcare}, publisher={Multidisciplinary Digital Publishing Institute}, author={Sise, Mary T}, year={2025}, month={May}, pages={1092–1092} }

@article{psychodynamic_interpretation_1, title={Operationalized psychodynamic diagnosis as an instrument to transfer psychodynamic constructs into neuroscience}, volume={7}, url={https://pmc.ncbi.nlm.nih.gov/articles/PMC3829565/}, DOI={https://doi.org/10.3389/fnhum.2013.00718}, journal={Frontiers in Human Neuroscience}, publisher={Frontiers Media SA}, author={Kessler, Henrik and Stasch, Michael and Cierpka, Manfred}, year={2013} }

@article{psychodynamic_interpretation_2, title={The effectiveness of psychodynamic psychotherapies: An update}, volume={14}, url={https://pmc.ncbi.nlm.nih.gov/articles/PMC4471961/}, DOI={https://doi.org/10.1002/wps.20235}, number={2}, journal={World Psychiatry}, publisher={Wiley}, author={Fonagy, Peter}, year={2015}, month={Jun}, pages={137–150} }

@article{psychodynamic_interpretation_3, title={The status of psychodynamic psychotherapy as an empirically supported treatment for common mental disorders – an umbrella review based on updated criteria}, volume={22}, url={https://pmc.ncbi.nlm.nih.gov/articles/PMC10168167/}, DOI={https://doi.org/10.1002/wps.21104}, number={2}, journal={World Psychiatry}, publisher={Wiley}, author={Leichsenring, Falk and Abbass, Allan and Heim, Nikolas and Keefe, John R and Kisely, Steve and Luyten, Patrick and Sven Rabung and Steinert, Christiane}, year={2023}, month={May}, pages={286–304} }

@article{psychodynamic_interpretation_4, title={Psychodynamic psychotherapy for complex trauma: targets, focus, applications, and outcomes}, volume={3}, url={https://pmc.ncbi.nlm.nih.gov/articles/PMC3218759/}, DOI={https://doi.org/10.2147/prbm.s10215}, journal={Psychology Research and Behavior Management}, publisher={Dove Medical Press}, author={Spermon, Deborah and Darlington, Yvonne and Gibney, Paul}, year={2010}, month={Dec}, pages={119–127} }

@article{psychodynamic_interpretation_5, title={The four basic components of psychoanalytic technique and derived psychoanalytic psychotherapies}, volume={15}, url={https://pmc.ncbi.nlm.nih.gov/articles/PMC5032492/}, DOI={https://doi.org/10.1002/wps.20368}, number={3}, journal={World Psychiatry}, publisher={Wiley}, author={Kernberg, Otto F}, year={2016}, month={Sep}, pages={287–288} }

@article{emotional_processing_facilitation_1, title={A theoretical model of emotional processing in visual artmaking and art therapy}, volume={90}, url={https://pmc.ncbi.nlm.nih.gov/articles/PMC11391909/}, DOI={https://doi.org/10.1016/j.aip.2024.102196}, journal={The Arts in Psychotherapy}, publisher={Elsevier BV}, author={Asnat Weinfeld-Yehoudayan and Czamanski-Cohen, Johanna and Cohen, Miri and Weihs, Karen L}, year={2024}, month={Aug}, pages={102196–102196} }

@article{emotional_processing_facilitation_2, title={In-session emotional expression predicts symptomatic and panic-specific reflective functioning improvements in panic-focused psychodynamic psychotherapy.}, volume={56}, url={https://pmc.ncbi.nlm.nih.gov/articles/PMC6745012/}, DOI={https://doi.org/10.1037/pst0000215}, number={4}, journal={Psychotherapy}, publisher={American Psychological Association}, author={Keefe, John R and Huque, Zeeshan M and DeRubeis, Robert J and Barber, Jacques P and Milrod, Barbara L and Chambless, Dianne L}, year={2019}, month={Mar}, pages={514–525} }

@article{emotional_processing_facilitation_3, title={New horizons in group psychotherapy research and practice from third wave positive psychology: a practice-friendly review}, volume={25}, url={https://pmc.ncbi.nlm.nih.gov/articles/PMC9893048/}, DOI={https://doi.org/10.4081/ripppo.2022.643}, number={3}, journal={Research in Psychotherapy Psychopathology Process and Outcome}, publisher={PAGEPress (Italy)}, author={Marmarosh, Cheri L and Sandage, Steven and Wade, Nathaniel and Captari, Laura E and Crabtree, Sarah}, year={2022}, month={Nov} }

@article{emotional_processing_facilitation_4, title={Eliciting emotional expressions in psychodynamic psychotherapies using telehealth: a clinical review and single case study using emotional awareness and expression therapy}, volume={36}, url={https://pmc.ncbi.nlm.nih.gov/articles/PMC9881109/}, DOI={https://doi.org/10.1080/02668734.2022.2037691}, number={2}, journal={Psychoanalytic Psychotherapy}, publisher={Informa UK Limited}, author={Ahlquist, Lauren R. and Yarns, Brandon C.}, year={2022}, month={Mar}, pages={124–140} }

@article{emotional_processing_facilitation_5, title={Facilitating emotional processing in depression: the application of exposure principles}, volume={4}, url={https://pmc.ncbi.nlm.nih.gov/articles/PMC4410732/}, DOI={https://doi.org/10.1016/j.copsyc.2015.03.032}, journal={Current Opinion in Psychology}, publisher={Elsevier BV}, author={Hayes, Adele M}, year={2015}, month={Aug}, pages={61–66} }

@article{countertransference_management_1_managing_resistance_4, title={Managing Transference and Countertransference in Cognitive Behavioral Supervision: Theoretical Framework and Clinical Application}, volume={Volume 15}, url={https://pmc.ncbi.nlm.nih.gov/articles/PMC9384966/}, DOI={https://doi.org/10.2147/prbm.s369294}, journal={Psychology Research and Behavior Management}, publisher={Dove Medical Press}, author={Prasko, Jan and Ociskova, Marie and Vanek, Jakub and Burkauskas, Julius and Milos Slepecky and Ieva Bite and Krone, Ilona and Sollar, Tomas and Alicja Juskiene}, year={2022}, month={Aug}, pages={2129–2155} }

@article{countertransference_management_2, title={Balancing clinical risk with countertransference management protects alliance}, volume={20}, url={https://pmc.ncbi.nlm.nih.gov/articles/PMC7451367/}, DOI={https://doi.org/10.4081/ripppo.2017.279}, number={3}, journal={Research in Psychotherapy Psychopathology Process and Outcome}, publisher={PAGEPress (Italy)}, author={Barreto, Jo{\~a}o F. and Mena Matos, Paula}, year={2017}, month={Oct}, pages={279–279} }

@article{countertransference_management_3, title={Experience of Managing Countertransference Through Self-Guided Imagery in Meditation Among Healthcare Professionals}, volume={13}, url={https://pmc.ncbi.nlm.nih.gov/articles/PMC8891567/}, DOI={https://doi.org/10.3389/fpsyt.2022.793784}, journal={Frontiers in Psychiatry}, publisher={Frontiers Media}, author={Olaug Julie Aasan and Hildfrid Vikkelsmo Brataas and Bente Nordtug}, year={2022}, month={Feb}, pages={793784–793784} }

@article{countertransference_management_4, title={Associations between countertransference reactions towards patients with borderline personality disorder and therapist experience levels and mentalization ability}, volume={43}, url={https://pmc.ncbi.nlm.nih.gov/articles/PMC8317550/}, DOI={https://doi.org/10.47626/2237-6089-2020-0025}, number={2}, journal={Trends in Psychiatry and Psychotherapy}, publisher={Associação de Psiquiatria do Rio Grande do Sul}, author={Bhola, Poornima and Mehrotra, Kanika}, year={2021}, month={Jan}, pages={116–125} }

@article{countertransference_management_5, title={The role of countertransference in contemporary psychiatric treatment}, volume={19}, url={https://pmc.ncbi.nlm.nih.gov/articles/PMC7214951/}, DOI={https://doi.org/10.1002/wps.20746}, number={2}, journal={World Psychiatry}, publisher={Wiley}, author={Gabbard, Glen O.}, year={2020}, month={May}, pages={243–244} }

@article{cultural_humility_1,
  author = {Lekas, Helen-Maria and Pahl, Kerstin and Fuller Lewis, Crystal},
  month = {12},
  title = {Rethinking cultural competence: Shifting to cultural humility},
  doi = {10.1177/1178632920970580},
  url = {https://www.ncbi.nlm.nih.gov/pmc/articles/PMC7756036/},
  volume = {13},
  year = {2020},
  journal = {Health Services Insights}
}

@article{cultural_humility_2,
  author = {Asnaani, Anu and Hofmann, Stefan G.},
  month = {01},
  pages = {187-197},
  title = {Collaboration in multicultural therapy: Establishing a strong therapeutic alliance across cultural lines},
  doi = {10.1002/jclp.21829},
  url = {https://www.ncbi.nlm.nih.gov/pmc/articles/PMC3641707/},
  volume = {68},
  year = {2012},
  journal = {Journal of Clinical Psychology}
}

@article{cultural_humility_3_dialectical_strategies_2,
  author = {Oshin, Linda A and Rizvi, Shireen L},
  month = {08},
  publisher = {American Psychological Association},
  title = {Considerations for the use of dialectical behavior therapy for individuals experiencing oppression.},
  doi = {10.1037/pst0000541},
  volume = {61},
  year = {2024},
  journal = {Psychotherapy}
}

@article{cultural_humility_4,
  author = {Konidaris, Mayio and Petrakis, Melissa},
  month = {06},
  pages = {1342-1342},
  publisher = {Multidisciplinary Digital Publishing Institute},
  title = {Cultural Humility Training in Mental Health Service Provision: A Scoping Review of the Foundational and Conceptual Literature},
  doi = {10.3390/healthcare13111342},
  url = {https://www.mdpi.com/2227-9032/13/11/1342},
  volume = {13},
  year = {2025},
  journal = {Healthcare}
}

@article{cultural_humility_5,
  author = {Bauer, Alexandria G. and Balaraman, Amudha and Gilmore, Ayanna},
  month = {06},
  publisher = {Springer Science and Business Media LLC},
  title = {Mental Health Providers’ Attitudes, Norms, and Beliefs About Cultural Humility in Service Delivery},
  doi = {10.1007/s11414-025-09953-3},
  volume = {52},
  year = {2025},
  journal = {The Journal of Behavioral Health Services \& Research}
}

@article{behavioral_activation_1,
  author = {Medina-Jiménez, Erick A and Acosta-Quiroz, Christian O and García-Flores, Raquel and Aguilar-Navarro, Sara G and Sotelo-Ojeda, Jesús E},
  month = {01},
  publisher = {SAGE Publishing},
  title = {Behavioral Activation Therapy for Depression Led by Health Personnel in Older People: A Scoping Review},
  doi = {10.1177/23337214241300652},
  volume = {10},
  year = {2024},
  journal = {Gerontology and Geriatric Medicine}
}

@article{behavioral_activation_2,
  author = {Dichter, Gabriel S. and Felder, Jennifer N. and Smoski, Moria J.},
  month = {10},
  pages = {236-244},
  title = {The effects of Brief Behavioral Activation Therapy for Depression on cognitive control in affective contexts: An fMRI investigation},
  doi = {10.1016/j.jad.2010.03.022},
  volume = {126},
  year = {2010},
  journal = {Journal of Affective Disorders}
}

@article{behavioral_activation_3,
  author = {Mazzucchelli, Trevor G. and Kane, Robert T. and Rees, Clare S.},
  month = {03},
  pages = {105-121},
  title = {Behavioral activation interventions for well-being: A meta-analysis},
  doi = {10.1080/17439760903569154},
  volume = {5},
  year = {2010},
  journal = {The Journal of Positive Psychology}
}

@article{behavioral_activation_4,
  author = {Uphoff, Eleonora and Ekers, David and Dawson, Sarah and Richards, David and Churchill, Rachel},
  month = {04},
  title = {Behavioural activation therapies for depression in adults},
  doi = {10.1002/14651858.cd013305},
  volume = {2019},
  year = {2019},
  journal = {Cochrane Database of Systematic Reviews}
}

@article{behavioral_activation_5,
  author = {Farrand, Paul and Pentecost, Claire and Greaves, Colin and Taylor, Rod S and Warren, Fiona and Green, Colin and Hillsdon, Melvyn and Evans, Phil and Welsman, Jo and Taylor, Adrian H},
  month = {05},
  title = {A written self-help intervention for depressed adults comparing behavioural activation combined with physical activity promotion with a self-help intervention based upon behavioural activation alone: study protocol for a parallel group pilot randomised controlled trial (BAcPAc)},
  doi = {10.1186/1745-6215-15-196},
  urldate = {2019-09-17},
  volume = {15},
  year = {2014},
  journal = {Trials}
}

@article{values_clarification_1, title={Working With Values: An Overview of Approaches and Considerations in Implementation}, volume={15}, url={https://pmc.ncbi.nlm.nih.gov/articles/PMC8854463/}, DOI={https://doi.org/10.1007/s40617-021-00589-1}, number={1}, journal={Behavior Analysis in Practice}, publisher={Springer Science+Business Media}, author={Berkout, Olga V}, year={2021}, month={Jul}, pages={104–114} }

@article{values_clarification_3, title={PREDICTORS AND IMPLICATIONS OF VALUES CLARITY IN FIRST-YEAR COLLEGE STUDENTS}, volume={53}, url={https://pmc.ncbi.nlm.nih.gov/articles/PMC8278292/}, number={4}, journal={College student journal}, author={BAYLY, BENJAMIN and BUMPUS, MATTHEW F}, year={2020}, month={Mar}, pages={397-404} }

@article{values_clarification_4, title={Clarifying Values: An Updated and Expanded Systematic Review and Meta-Analysis}, volume={41}, url={https://pmc.ncbi.nlm.nih.gov/articles/PMC8482297/}, DOI={https://doi.org/10.1177/0272989x211037946}, number={7}, journal={Medical Decision Making}, publisher={SAGE Publications}, author={Witteman, Holly O. and Ndjaboue, Ruth and Vaisson, Gratianne and Dansokho, Selma Chipenda and Arnold, Bob and Bridges, John F. P. and Comeau, Sandrine and Fagerlin, Angela and Gavaruzzi, Teresa and Marcoux, Melina and Pieterse, Arwen and Pignone, Michael and Provencher, Thierry and Racine, Charles and Regier, Dean and Rochefort-Brihay, Charlotte and Thokala, Praveen and Weernink, Marieke and White, Douglas B. and Wills, Celia E.}, year={2021}, month={Sep}, pages={801–820} }

@article{values_clarification_5, title={Exploring Values Clarification and Health-Literate Design in Patient Decision Aids: A Qualitative Interview Study}, volume={45}, url={https://pmc.ncbi.nlm.nih.gov/articles/PMC12166136/}, DOI={https://doi.org/10.1177/0272989x251334356}, number={5}, journal={Medical Decision Making}, publisher={SAGE Publishing}, author={Ayre, Julie and Jenkins, Hazel and Kumarage, Richie and McCaffery, Kirsten J and Maher, Christopher G and Hancock, Mark J}, year={2025}, month={May}, pages={510–521} }

@article{chain_analysis_1, title={Emotion Regulation, Physical Diseases, and Borderline Personality Disorders: Conceptual and Clinical Considerations}, volume={12}, url={https://pmc.ncbi.nlm.nih.gov/articles/PMC7882545/}, DOI={https://doi.org/10.3389/fpsyg.2021.567671}, journal={Frontiers in Psychology}, publisher={Frontiers Media SA}, author={Cavicchioli, Marco and Barone, Lavinia and Fiore, Donatella and Marchini, Monica and Pazzano, Paola and Ramella, Pietro and Riccardi, Ilaria and Sanza, Michele and Maffei, Cesare}, year={2021}, month={Feb} }

@article{chain_analysis_2, title={Emotion Regulation in Schema Therapy and Dialectical Behavior Therapy}, volume={7}, url={https://pmc.ncbi.nlm.nih.gov/articles/PMC5021701/}, DOI={https://doi.org/10.3389/fpsyg.2016.01373}, journal={Frontiers in Psychology}, publisher={Frontiers Media}, author={Fassbinder, Eva and Schweiger, Ulrich and Martius, Desiree and Wilde, Odette Brand-de and Arntz, Arnoud}, year={2016}, month={Sep}, pages={1373–1373} }

@article{chain_analysis_3, title={A dialectical behavior therapy skills intervention for women with suicidal behaviors in rural Nepal: A single‐case experimental design series}, volume={74}, url={https://pmc.ncbi.nlm.nih.gov/articles/PMC6002746/}, DOI={https://doi.org/10.1002/jclp.22588}, number={7}, journal={Journal of Clinical Psychology}, publisher={Wiley}, author={Ramaiya, Megan K and McLean, Caitlin and Upasana Regmi and Fiorillo, Devika and Robins, Clive J and Kohrt, Brandon A}, year={2018}, month={Feb}, pages={1071–1091} }

@article{chain_analysis_4, title={Strategies to Deal With Suicide and Non-suicidal Self-Injury in Borderline Personality Disorder, the Case of DBT}, volume={9}, url={https://pmc.ncbi.nlm.nih.gov/articles/PMC6304419/}, DOI={https://doi.org/10.3389/fpsyg.2018.02595}, journal={Frontiers in Psychology}, publisher={Frontiers Media SA}, author={Prada, Paco and Perroud, Nader and Rüfenacht, Eva and Nicastro, Rosetta}, year={2018}, month={Dec} }

@article{chain_analysis_5, title={Treating Depressed and Suicidal Adolescents}, volume={22}, url={https://pmc.ncbi.nlm.nih.gov/articles/PMC3565721/}, number={1}, journal={Journal of the Canadian Academy of Child and Adolescent Psychiatry}, author={Grintuch, Benjamin and Waheed, Waqar}, year={2013}, month={Feb}, pages={71-72} }

@article{safety_planning_1, title={Suicide Safety Planning: Clinician Training, Comfort, and Safety Plan Utilization}, volume={17}, url={https://pmc.ncbi.nlm.nih.gov/articles/PMC7559434/}, DOI={https://doi.org/10.3390/ijerph17186444}, number={18}, journal={International Journal of Environmental Research and Public Health}, publisher={Multidisciplinary Digital Publishing Institute}, author={Moscardini, Emma H and Hill, Ryan M and Dodd, Cody G and Do, Calvin and Kaplow, Julie B and Tucker, Raymond P}, year={2020}, month={Sep}, pages={6444–6444} }

@article{safety_planning_2, title={Effectiveness of Suicide Safety Planning Interventions: A Systematic Review Informing Occupational Therapy}, volume={90}, url={https://pmc.ncbi.nlm.nih.gov/articles/PMC10189833/}, DOI={https://doi.org/10.1177/00084174221132097}, number={2}, journal={Canadian Journal of Occupational Therapy}, publisher={SAGE Publishing}, author={Marshall, Carrie Anne and Crowley, Pavlina and Carmichael, Dave and Goldszmidt, Rebecca and Suliman Aryobi and Holmes, Julia and Easton, Corinna and Isard, Roxanne and Murphy, Susanne}, year={2022}, month={Nov}, pages={208–236} }

@article{safety_planning_3, title={Enhancing motivation and self-efficacy for safety plan use: Incorporating motivational interviewing strategies in a brief safety planning intervention for adolescents at risk for suicide.}, volume={59}, url={https://pmc.ncbi.nlm.nih.gov/articles/PMC8799764/}, DOI={https://doi.org/10.1037/pst0000374}, number={2}, journal={Psychotherapy}, publisher={American Psychological Association}, author={Micol, Valerie J and Prouty, David and Czyz, Ewa K}, year={2021}, month={Jul}, pages={174–180} }

@article{safety_planning_4, title={How can we work together to keep you safe? A simple and effective intervention that can save lives}, volume={71}, url={https://pmc.ncbi.nlm.nih.gov/articles/PMC8378553/}, DOI={https://doi.org/10.3399/bjgp21x716909}, number={710}, journal={British Journal of General Practice}, publisher={Royal College of General Practitioners}, author={Michail, Maria}, year={2021}, month={Aug}, pages={409–409} }

@article{safety_planning_5, title={Assessing Variability and Implementation Fidelity of Suicide Prevention Safety Planning in a Regional VA Healthcare System}, volume={36}, url={https://pmc.ncbi.nlm.nih.gov/articles/PMC4675034/}, DOI={https://doi.org/10.1027/0227-5910/a000345}, number={6}, journal={Crisis}, publisher={Hogrefe Verlag}, author={Gamarra, Jennifer M and Luciano, Matthew T and Gradus, Jaimie L and Stirman, Shannon Wiltsey}, year={2015}, month={Nov}, pages={433–439} }

@article{grounding_techniques_1, title={Psychological Interventions for Dissociative disorders}, volume={62}, url={https://pmc.ncbi.nlm.nih.gov/articles/PMC7001344/}, DOI={https://doi.org/10.4103/psychiatry.indianjpsychiatry_777_19}, journal={Indian Journal of Psychiatry}, publisher={Medknow}, author={Subramanyam, Alka A. and Somaiya, Mansi and Shankar, Sunitha and Nasirabadi, Minhaj and Shah, Henal R. and Paul, Imon and Ghildiyal, Rakesh}, year={2020}, volume = {62},
number = {Suppl 2},
pages = {S280--S289}}

@article{grounding_techniques_2, title={Practical applications of grounding to support health}, volume={46}, url={https://pmc.ncbi.nlm.nih.gov/articles/PMC10105020/}, DOI={https://doi.org/10.1016/j.bj.2022.12.001}, number={1}, journal={Biomedical Journal}, publisher={Elsevier BV}, author={Koniver, Laura}, year={2022}, month={Dec}, pages={41–47} }

@article{grounding_techniques_3, title={Mindfulness Exercises Reduce Acute Physiologic Stress Among Female Clinicians}, volume={6}, url={https://pmc.ncbi.nlm.nih.gov/articles/PMC11519409/}, DOI={https://doi.org/10.1097/cce.0000000000001171}, number={11}, journal={Critical Care Explorations}, publisher={Wolters Kluwer}, author = {Wolfe, Amy H. J. and
          Hinds, Pamela S. and
          du Plessis, Adre J. and
          Gordish-Dressman, Heather and
          Freedenberg, Vicki and
          Soghier, Lamia}, year={2024}, month={Oct}, pages={e1171–e1171} }

@article{grounding_techniques_4, title={Mind–Body Practices for Mental Health in Higher Education: Breathing, Grounding, and Consistency Are Essential for Stress and Anxiety Management}, volume={13}, url={https://pmc.ncbi.nlm.nih.gov/articles/PMC12385911/}, DOI={https://doi.org/10.3390/healthcare13162049}, number={16}, journal={Healthcare}, publisher={Multidisciplinary Digital Publishing Institute}, author={Frausing, Kristian Park and Manja Harsted Flammild and Jesper Dahlgaard}, year={2025}, month={Aug}, pages={2049–2049} }

@article{grounding_techniques_5, title={A brief transdiagnostic pandemic mental health maintenance intervention}, volume={34}, url={https://pmc.ncbi.nlm.nih.gov/articles/PMC8664004/}, DOI={https://doi.org/10.1080/09515070.2020.1769026}, number={3-4}, journal={Counselling Psychology Quarterly}, publisher={Taylor & Francis}, author={Arnold, Trisha and Rogers, Brooke G and Norris, Alyssa L and Scherr, Anna Schierberl and Haubrick, Kayla and Renna, Megan E and Sun, Shufang and Danforth, Margaret M and Chu, Christina T and Silva, Elizabeth S and Whiteley, Laura B and Pinkston, Megan}, year={2020}, month={May}, pages={331–351} }

@article{clarification_1, title={Two Techniques of Supportive Psychotherapy}, volume={35}, url={https://pmc.ncbi.nlm.nih.gov/articles/PMC2280366/}, journal={Canadian Family Physician}, author={Steinberg, Paul Ian}, year={1989}, month={May}, pages={1139-1143} }

@article{clarification_2, title={Clinical Practice Guidelines for Practice of Supportive Psychotherapy},  url={https://pmc.ncbi.nlm.nih.gov/articles/PMC7001359/}, DOI={https://doi.org/10.4103/psychiatry.indianjpsychiatry_768_19},  journal={Indian Journal of Psychiatry}, publisher={Medknow}, author={Grover, Sandeep and Ajit Avasthi and Mukesh Jagiwala}, year={2020}, month={Jan}, volume = {62},
number = {Suppl 2}, pages = {S173--S182}, }

@article{clarification_3_values_clarification_2_autonomy_support_1, title={Beyond values clarification: Addressing client values in clinical behavior analysis}, volume={32}, url={https://pmc.ncbi.nlm.nih.gov/articles/PMC2686993/}, DOI={https://doi.org/10.1007/bf03392176}, number={1}, journal={The Behavior Analyst}, publisher={Springer Science+Business Media}, author={Bonow, Jordan T and Follette, William C}, year={2009}, month={Apr}, pages={69–84} }

@article{clarification_4, title={Supportive Techniques: Are They Found in Different Therapies?}, volume={10}, url={https://pmc.ncbi.nlm.nih.gov/articles/PMC3330649/}, number={3}, journal={The Journal of Psychotherapy Practice and Research}, author={Barber, Jacques P and Stratt, Rachael and Halperin, Gregory and Connolly, Mary Beth}, year={2001}, pages={165-172} }

@article{clarification_5, title={How therapists in cognitive behavioral and psychodynamic therapy reflect upon the use of metaphors in therapy: a qualitative study}, volume={22}, url={https://pmc.ncbi.nlm.nih.gov/articles/PMC9235099/}, DOI={https://doi.org/10.1186/s12888-022-04083-y}, number={1}, journal={BMC Psychiatry}, publisher={Springer Science and Business Media LLC}, author={Malkomsen, A and Røssberg, JI and Dammen, T and Wilberg, T and Løvgren, A and Ulberg, R and Evensen, J}, year={2022}, month={Jun} }

@article{dialectical_strategies_1, title={Application of Dialectical Behaviour Therapy in treating common psychiatric disorders: study protocol for a scoping review}, volume={12}, url={https://pmc.ncbi.nlm.nih.gov/articles/PMC9516170/}, DOI={https://doi.org/10.1136/bmjopen-2021-058565}, number={9}, journal={BMJ Open}, publisher={BMJ}, author={Yan, Michelle and McConnell, Bridget and Barlas, Joanna}, year={2022}, month={Sep}, pages={e058565–e058565} }

@article{dialectical_strategies_3, title={Applying Dialectical Behavior Therapy as a Transdiagnostic Treatment in a Case of Borderline Personality Disorder and Eating Disorder}, volume={81}, url={https://pmc.ncbi.nlm.nih.gov/articles/PMC11724320/}, DOI={https://doi.org/10.1002/jclp.23754}, number={2}, journal={Journal of Clinical Psychology}, publisher={Wiley}, author={María Vicenta Navarro‐Haro and Morillo, Alba Abanades and García‐Palacios, Azucena}, year={2024}, month={Nov}, pages={102–112} }

@article{dialectical_strategies_4, title={Teaching Dialectical Behavior Therapy to Psychiatry Residents: The Columbia Psychiatry Residency DBT Curriculum}, volume={41}, url={https://pmc.ncbi.nlm.nih.gov/articles/PMC5247344/}, DOI={https://doi.org/10.1007/s40596-016-0593-0}, number={1}, journal={Academic Psychiatry}, publisher={Springer Science and Business Media LLC}, author={Brodsky, Beth S. and Cabaniss, Deborah L. and Arbuckle, Melissa and Oquendo, Maria A. and Stanley, Barbara}, year={2016}, month={Aug}, pages={10–15} }

@article{dialectical_strategies_5, title={The temporal relationships between therapist adherence and patient outcomes in dialectical behavior therapy.}, volume={90}, url={https://pmc.ncbi.nlm.nih.gov/articles/PMC12020511/}, DOI={https://doi.org/10.1037/ccp0000714}, number={3}, journal={Journal of Consulting and Clinical Psychology}, publisher={American Psychological Association}, author={Harned, Melanie S and Gallop, Robert J and Schmidt, Sara C and Korslund, Kathryn E}, year={2022}, month={Jan}, pages={272–281} }

@article{emotion_focused_skills_1, title={Emotion-Focused Therapy–Therapist Fidelity Scale (EFT-TFS): Conceptual Development and Content Validity}, volume={8}, url={https://pmc.ncbi.nlm.nih.gov/articles/PMC2801569/}, DOI={https://doi.org/10.1080/15332690903048820}, number={3}, journal={Journal of Couple \& Relationship Therapy}, publisher={Taylor & Francis}, author={Denton, Wayne H and Johnson, Susan M and Burleson, Brant R}, year={2009}, month={Aug}, pages={226–246} }

@article{emotion_focused_skills_2, title={New Developments in Emotion-Focused Therapy for Social Anxiety Disorder}, volume={9}, url={https://pmc.ncbi.nlm.nih.gov/articles/PMC7565910/}, DOI={https://doi.org/10.3390/jcm9092918}, number={9}, journal={Journal of Clinical Medicine}, publisher={MDPI AG}, author={Shahar, Ben}, year={2020}, month={Sep}, pages={2918} }

@article{emotion_focused_skills_3, title={Effectiveness of emotion focused skills training for parents: study protocol for a randomized controlled trial in specialist mental health care}, volume={22}, url={https://pmc.ncbi.nlm.nih.gov/articles/PMC9261057/}, DOI={https://doi.org/10.1186/s12888-022-04084-x}, number={1}, journal={BMC Psychiatry}, publisher={BioMed Central}, author={Severinsen, Linda and Stiegler, Jan Reidar and Nissen-Lie, Helene Amundsen and Shahar, Ben and Rune Zahl-Olsen}, year={2022}, month={Jul}, pages={453–453} }

@article{emotion_focused_skills_4, title={Cognitive-behavioral and emotion-focused couple therapy: Similarities and differences}, volume={2}, url={https://pmc.ncbi.nlm.nih.gov/articles/PMC9645475/}, DOI={https://doi.org/10.32872/cpe.v2i3.2741}, number={3}, journal={Clinical Psychology in Europe}, publisher={Leibniz Institute for Psychology (ZPID)}, author={Bodenmann, Guy and Kessler, Mirjam and Kuhn, Rebekka and Hocker, Lauren and Randall, Ashley K.}, year={2020}, month={Sep} }

@article{emotion_focused_skills_5, title={The role of emotion in psychological therapy.}, volume={14}, url={https://pmc.ncbi.nlm.nih.gov/articles/PMC2562704/}, DOI={https://doi.org/10.1111/j.1468-2850.2007.00102.x}, number={4}, journal={Clinical Psychology: Science and Practice}, publisher={American Psychological Association (APA)}, author={Ehrenreich, Jill T. and Fairholme, Christopher P. and Buzzella, Brian A. and Ellard, Kristen K. and Barlow, David H.}, year={2007}, month={Dec}, pages={422–428} }

@article{elliott_bohart_watson_greenberg_2011_empathy, title={Empathy}, volume={48}, DOI={https://doi.org/10.1037/a0022187}, number={1}, journal={Psychotherapy}, author={Elliott, Robert and Bohart, Arthur C. and Watson, Jeanne C. and Greenberg, Leslie S.}, year={2011}, pages={43–49} }

@article{lazarus_atzil_slonim_bar_kalifa_hasson_ohayon_rafaeli_2019_clients_emotional_instability_and_therapists_inferential_flexibility_predict_therapists_session_by_session_empathetic_accuracy, title={Clients’ emotional instability and therapists’ inferential flexibility predict therapists’ session-by-session empathic accuracy.}, volume={66}, DOI={https://doi.org/10.1037/cou0000310}, number={1}, journal={Journal of Counseling Psychology}, author={Lazarus, Gal and Atzil-Slonim, Dana and Bar-Kalifa, Eran and Hasson-Ohayon, Ilanit and Rafaeli, Eshkol}, year={2019}, month={Jan}, pages={56–69} }

@misc{huang2025medscoregeneralizablefactualityevaluation,
      title={MedScore: Generalizable Factuality Evaluation of Free-Form Medical Answers by Domain-adapted Claim Decomposition and Verification}, 
      author={Heyuan Huang and Alexandra DeLucia and Vijay Murari Tiyyala and Mark Dredze},
      year={2025},
      eprint={2505.18452},
      archivePrefix={arXiv},
      primaryClass={cs.CL},
      url={https://arxiv.org/abs/2505.18452}, 
}

@article{atzil_slonim_bar_kalifa_fisher_lazarus_hasson_ohayon_lutz_rubel_rafaeli_2019_therapists_empathic_accuracy_accuracy_towards_their_clients_emotions, title={Therapists’ empathic accuracy toward their clients’ emotions.}, volume={87}, DOI={https://doi.org/10.1037/ccp0000354}, number={1}, journal={Journal of Consulting and Clinical Psychology}, author={Atzil-Slonim, Dana and Bar-Kalifa, Eran and Fisher, Hadar and Lazarus, Gal and Hasson-Ohayon, Ilanit and Lutz, Wolfgang and Rubel, Julian and Rafaeli, Eshkol}, year={2019}, month={Jan}, pages={33–45} }

@article{liu_wei_tu_lin_jiang_cambria_2025_knowing_what_and_why, title={Knowing What and Why: Causal emotion entailment for emotion recognition in conversations}, volume={274}, DOI={https://doi.org/10.1016/j.eswa.2025.126924}, journal={Expert Systems with Applications}, author={Liu, Hao and Wei, Runguo and Tu, Geng and Lin, Jiali and Jiang, Dazhi and Cambria, Erik}, year={2025}, month={May}, pages={126924} }

@article{rogers_1957_the_necessary_and_sufficient_conditions_of_therapeutic_personality_change, title={The necessary and sufficient conditions of therapeutic personality change}, volume={21}, url={https://doi.org/10.1037/h0045357}, DOI={https://doi.org/10.1037/h0045357}, number={2}, journal={Journal of Consulting Psychology}, author={Rogers, Carl }, year={1957}, pages={95–103} }

@article{wampold_2015_how_important_are_the_common_factors_in_psychotherapy, title={How important are the common factors in psychotherapy? An update}, volume={14}, url={https://pmc.ncbi.nlm.nih.gov/articles/PMC4592639/}, DOI={https://doi.org/10.1002/wps.20238}, number={3}, journal={World Psychiatry}, author={Wampold, Bruce E.}, year={2015}, pages={270–277} }

@article{dejonckheere_vaughn_2019_semistructured_interview, title={Semistructured interviewing in primary care research: A balance of relationship and rigour}, volume={7}, DOI={https://doi.org/10.1136/fmch-2018-000057}, number={2}, journal={Family Medicine and Community Health}, publisher={BMJ Specialist Journals}, author={DeJonckheere, Melissa and Vaughn, Lisa M}, year={2019}, month={Mar}, pages={1–8} }

@misc{lees2022newgenerationperspectiveapi,
      title={A New Generation of Perspective API: Efficient Multilingual Character-level Transformers}, 
      author={Alyssa Lees and Vinh Q. Tran and Yi Tay and Jeffrey Sorensen and Jai Gupta and Donald Metzler and Lucy Vasserman},
      year={2022},
      eprint={2202.11176},
      archivePrefix={arXiv},
      primaryClass={cs.CL},
      url={https://arxiv.org/abs/2202.11176}, 
}

@misc{Detoxify,
  title={Detoxify},
  author={Hanu, Laura and {Unitary team}},
  howpublished={Github. https://github.com/unitaryai/detoxify},
  year={2020}
}

@book{hill2009helping,
  author    = {Hill, Clara E.},
  title     = {Helping Skills: Facilitating, Exploration, Insight, and Action},
  year      = {2009},
  publisher = {American Psychological Association}
}

@misc{bai2026irulerintelligiblerubricbaseduserdefined,
      title={iRULER: Intelligible Rubric-Based User-Defined LLM Evaluation for Revision}, 
      author={Jingwen Bai and Wei Soon Cheong and Philippe Muller and Brian Y Lim},
      year={2026},
      eprint={2602.12779},
      archivePrefix={arXiv},
      primaryClass={cs.HC},
      url={https://arxiv.org/abs/2602.12779}, 
}

@misc{shi2025humanintheloopframeworkreliablepatch,
      title={Towards a Human-in-the-Loop Framework for Reliable Patch Evaluation Using an LLM-as-a-Judge}, 
      author={Sherry Shi and Renyao Wei and Michele Tufano and José Cambronero and Runxiang Cheng and Franjo Ivančić and Pat Rondon},
      year={2025},
      eprint={2511.10865},
      archivePrefix={arXiv},
      primaryClass={cs.SE},
      url={https://arxiv.org/abs/2511.10865}, 
}

@misc{hartmann2022emotionenglish,
  author={Hartmann, Jochen},
  title={Emotion English DistilRoBERTa-base},
  year={2022},
  howpublished = {\url{https://huggingface.co/j-hartmann/emotion-english-distilroberta-base/}},
}
